%% file: main.tex
\documentclass[runningheads]{llncs}

\usepackage{eccv}

\usepackage{eccvabbrv}

\usepackage{graphicx}
\usepackage{booktabs}

\usepackage[accsupp]{axessibility}  

\usepackage{booktabs}       
\usepackage{amsfonts}       
\usepackage{nicefrac}       
\usepackage{microtype}      

\usepackage{epsfig}
\usepackage{graphicx}
\usepackage{amsmath}
\usepackage{amssymb}

\usepackage{enumerate}
\usepackage{multirow}
\usepackage{tabularx}
\usepackage{nicefrac}
\usepackage{booktabs}
\usepackage{enumitem}
\usepackage{makecell}
\usepackage{xspace}
\usepackage{graphbox}
\usepackage{pifont}
\usepackage[dvipsnames]{xcolor}
\usepackage{colortbl}
\usepackage{array}
\usepackage[linesnumbered,ruled]{algorithm2e}
\usepackage{wrapfig}
\usepackage{glossaries}
\glsdisablehyper
\usepackage{tikz}
\usepackage{bbm}
\usepackage{soul}

\newcommand{\mh}[1]{{\color{black}{#1}}}

\newcommand{\nds}[1]{{\color{black}{#1}}}

\graphicspath{{figures}}

\usepackage[pagebackref,breaklinks,colorlinks,citecolor=eccvblue]{hyperref}

\usepackage{orcidlink}

\input{commands}

\input{math}

\usepackage{url}

\newif\ifshowappendix
\showappendixtrue

\ifshowappendix
\else
\fi

\begin{document}

\title{Towards Reliable Evaluation and Fast Training of Robust Semantic Segmentation Models}

\titlerunning{Towards Robust Semantic Segmentation}

\author{Francesco Croce$^*$\inst{1} 
\and
Naman D. Singh$^*$\inst{2,3}
\and
Matthias Hein\inst{2,3}
}

\authorrunning{F.~Croce et al.}

\institute{
     EPFL \and University of T{\"u}bingen \and T{\"u}bingen AI Center \\
     }

\maketitle
\makeatletter\def\Hy@Warning#1{}\makeatother
\blfootnote{$^*$ Equal Contribution. Correspondence to: \texttt{naman-deep.singh@uni-tuebingen.de}}

\begin{abstract}
Adversarial robustness has been studied  extensively in image classification, especially for the $\ell_\infty$-threat model, but significantly less so for related tasks such as object detection and semantic segmentation, where attacks turn out to be a much harder optimization problem than for image classification. We propose several problem-specific novel attacks minimizing different metrics in 
accuracy and mIoU. The ensemble of our attacks, \multiloss, shows that existing attacks severely overestimate the robustness of semantic segmentation models.
Surprisingly, existing attempts of adversarial training for semantic segmentation models turn out to be weak or even completely non-robust. We investigate why previous adaptations of adversarial training to semantic segmentation failed and  show how recently proposed robust \imagenet backbones can be used to obtain adversarially robust semantic segmentation models with up to six times less training time for \voc and the more challenging \ade. The associated code and robust models are available at \href{https://github.com/nmndeep/robust-segmentation}{\texttt{https://github.com/nmndeep/robust-segmentation}}.
\end{abstract}

\section{Introduction}
\label{sec:intro}

The vulnerability of neural networks to adversarial perturbations, that is small changes of the input that can drastically modify the output of the models, has been extensively studied \cite{BigEtAl13, SzeEtAl2014, grosse2016adversarial, jin2019bert}, in particular for image classification. 
Adversarial attacks have been developed for various threat models, including $\ell_p$-bounded perturbations \cite{CarWag2016,CheEtAl2018, rony2019decoupling}, sparse attacks \cite{BroEtAl2017, croce2020sparsers}, and those defined by perceptual metrics \cite{wong2019wasserstein, laidlaw2021perceptual}.
At the same time, evaluating adversarial robustness in semantic segmentation, undoubtedly an important vision task, has received much less attention.
While a few early works \cite{xie2017adversarial, hendrik2017universal, arnab2018robustness} have proposed 
adversarial attacks in different threat models, \cite{gu2022segpgd, agnihotri2023cospgd} have recently shown that, even for 
$\ell_\infty$-bounded pertubations, significant improvements are possible over the standard PGD attack \cite{MadEtAl2018} based on the sum of pixel-wise cross-entropy losses.
In fact, the key difference of semantic segmentation to image classification is that for the former one has to flip the predictions of all pixels, not just the prediction for the image, which is a much harder optimization problem.
This might also explain why only few works \cite{xu2021dynamic, gu2022segpgd}  
have produced robust semantic segmentation models via variants of adversarial training \cite{MadEtAl2018}.
In this work, we address 
both the evaluation and training of adversarially robust semantic segmentation models.

\textbf{Challenges of adversarial attacks in semantic segmentation.} 
We 
analyze why the cross-entropy loss,  in contrast to image classification, is not a suitable objective 
for generating strong adversarial attacks against semantic segmentation models. To address this issue, we propose novel loss functions that \hlrev{are specifically designed for semantic segmentation (\cref{sec:attacks}), as they aim at flipping} the decisions of \emph{all} pixels simultaneously. 
Moreover, in contrast to prior work we do not only attack accuracy but also \miou (mean intersection over union), the most common performance metric in semantic segmentation. As the direct optimization of \miou is intractable since it depends on the outcome on all test images together, we derive an upper bound on \miou in terms of an imagewise loss which 
\hlrev{can be used} as attack objective (\cref{sec:mIoU}).
Finally, we study several improvements for the PGD-attack  (\cref{sec:opt}) which boost performance.

\begin{figure}[t] \centering
\scriptsize
\tabcolsep=0.5pt
\newl=.135\columnwidth
\begin{tabular}{c | c c c | cc c} 
&\multicolumn{3}{c}{{\textbf{clean model}}} & \multicolumn{3}{c}{\textbf{Our robust model}} \\
\makecell{ground  truth}&original & \makecell{target: grass} & \makecell{target: sky} & original & \makecell{target: grass} & \makecell{target: sky} \\

\includegraphics[width=\newl]{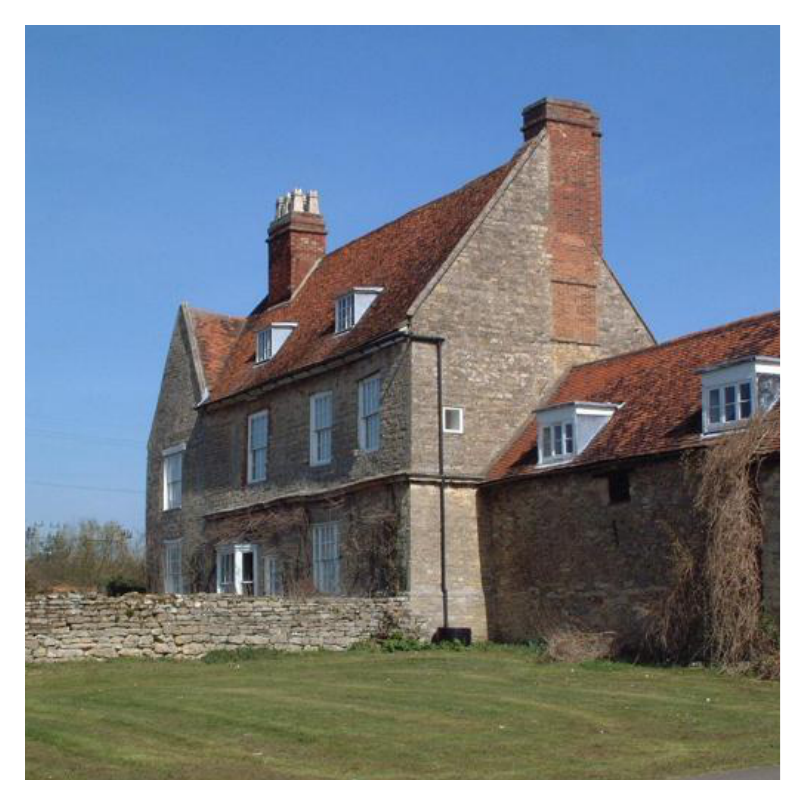} &
\includegraphics[width=\newl]{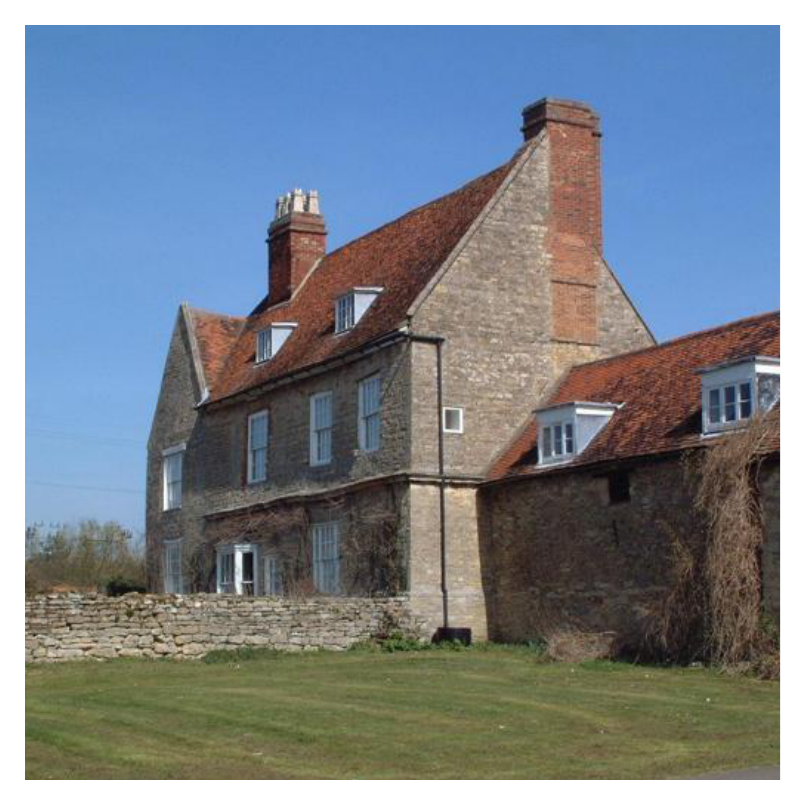} & \includegraphics[width=\newl]{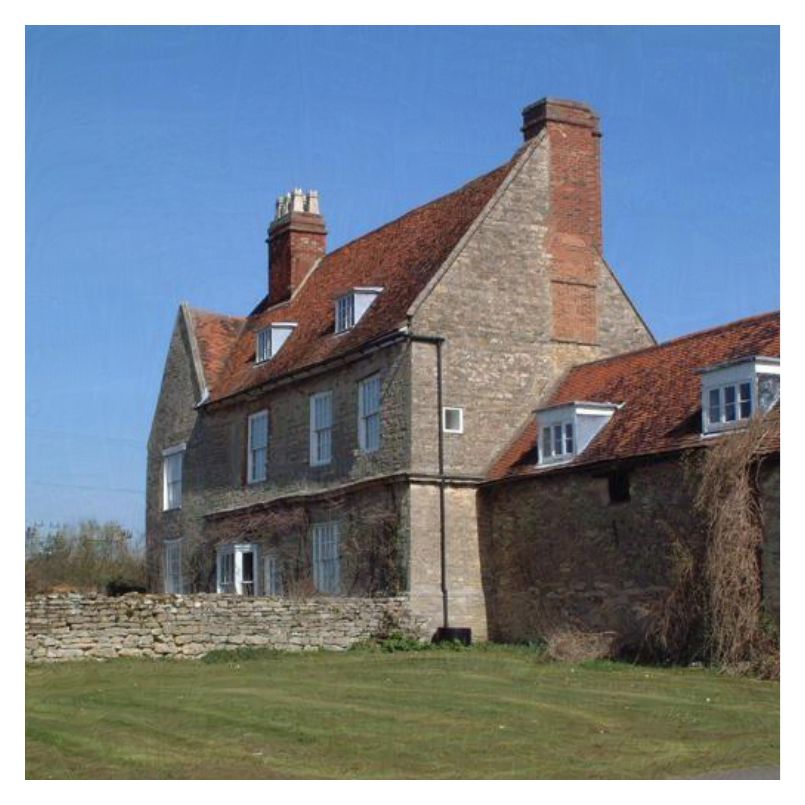} & \includegraphics[width=\newl]{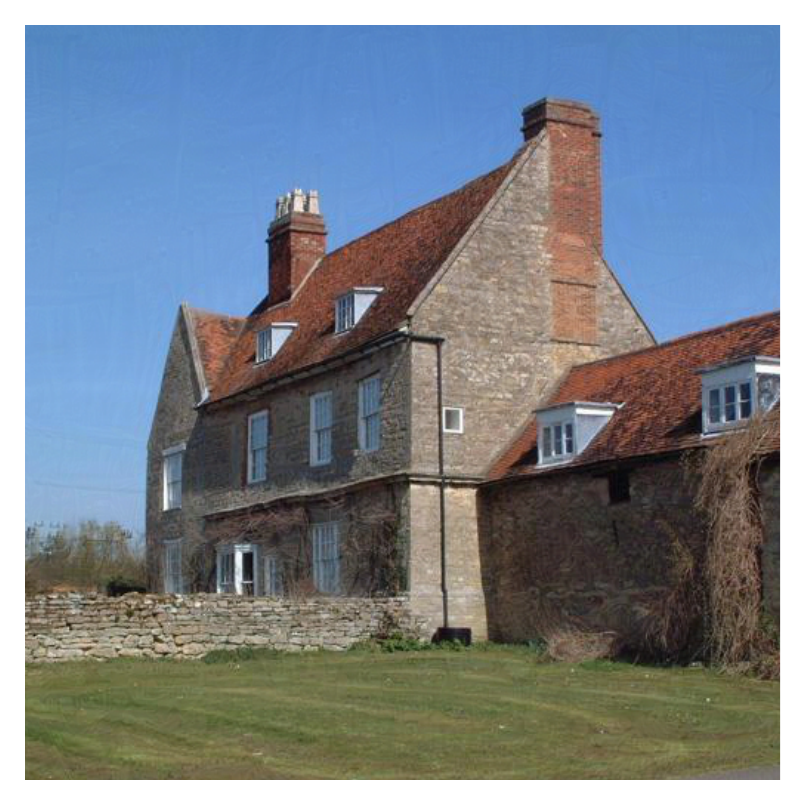} & \includegraphics[width=\newl]{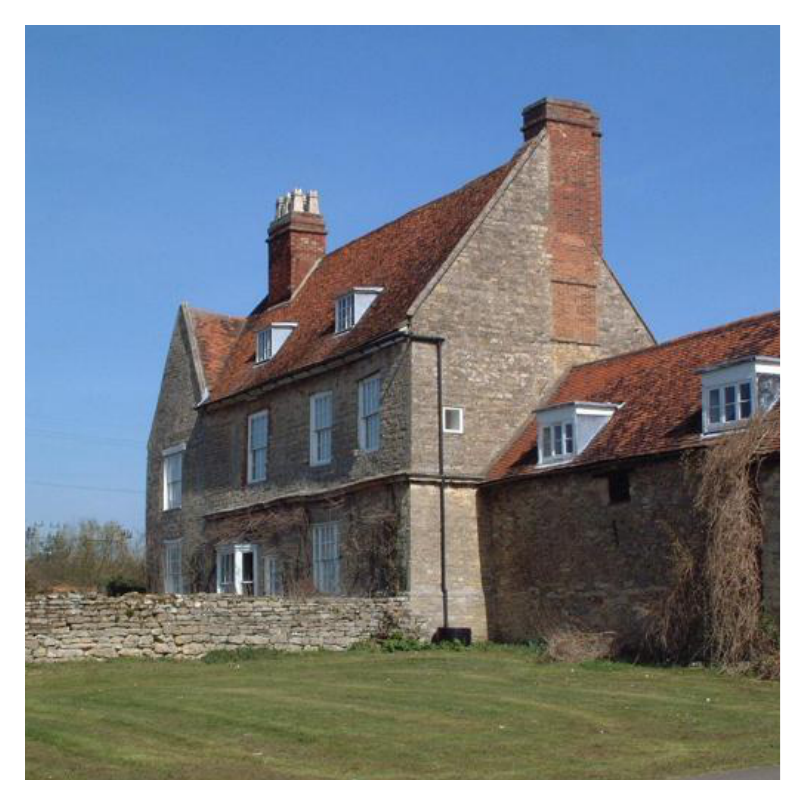} & \includegraphics[width=\newl]{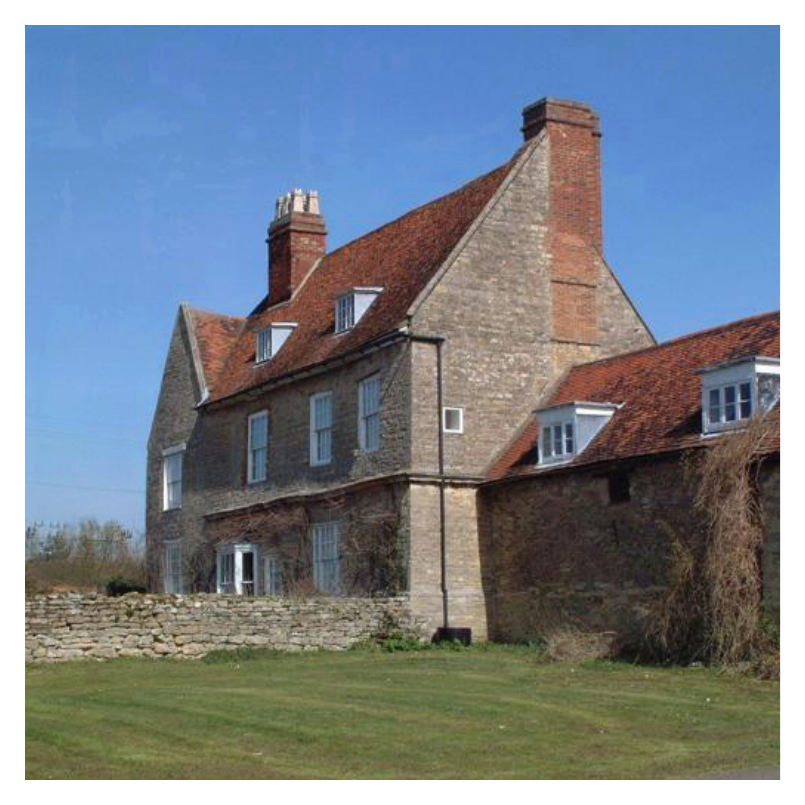} & \includegraphics[width=\newl]{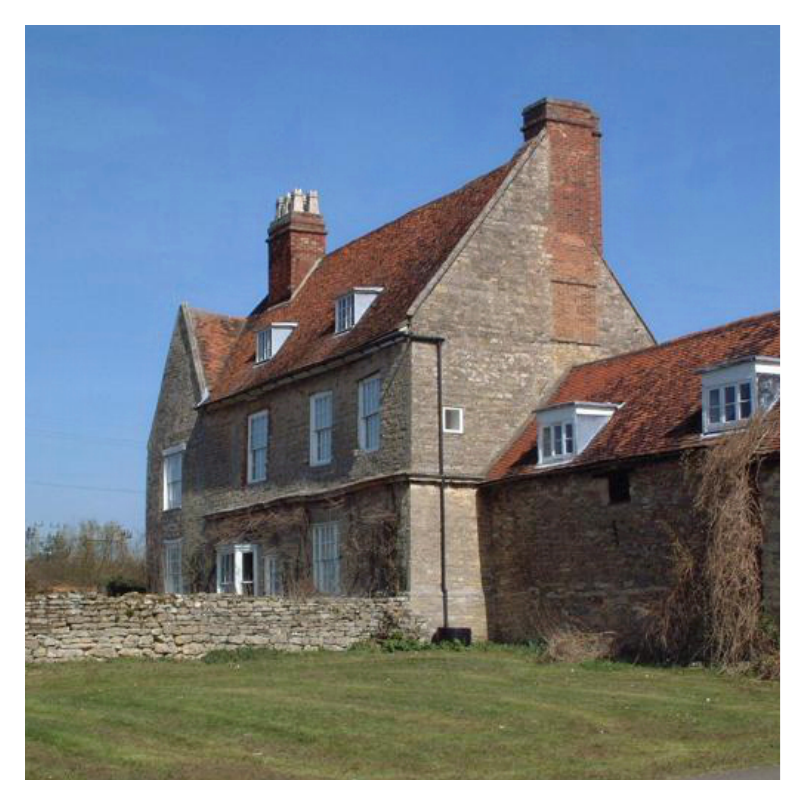} \\
\includegraphics[width=\newl]{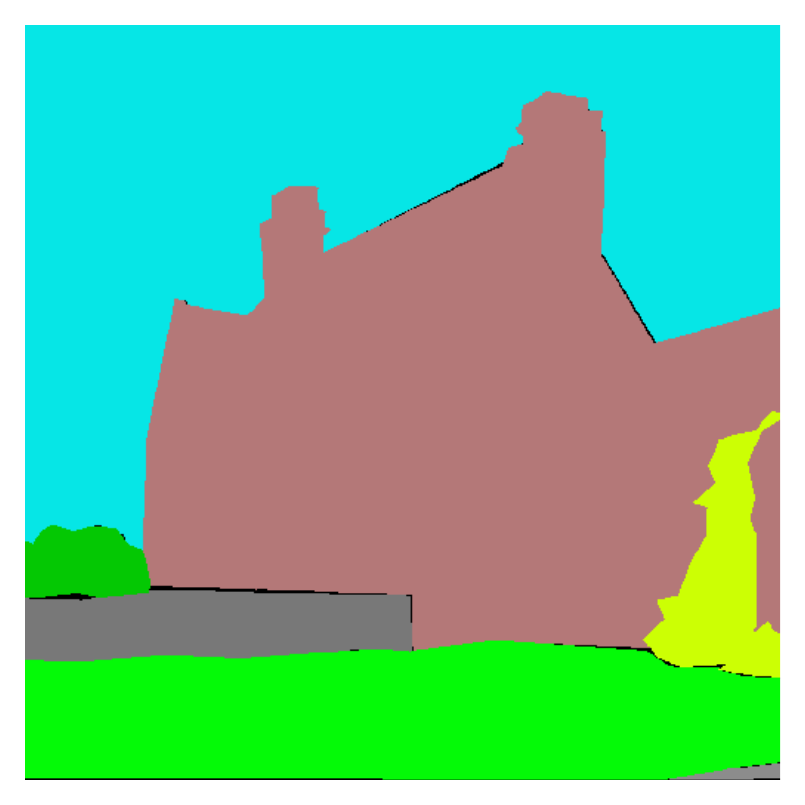} &
\includegraphics[width=\newl]{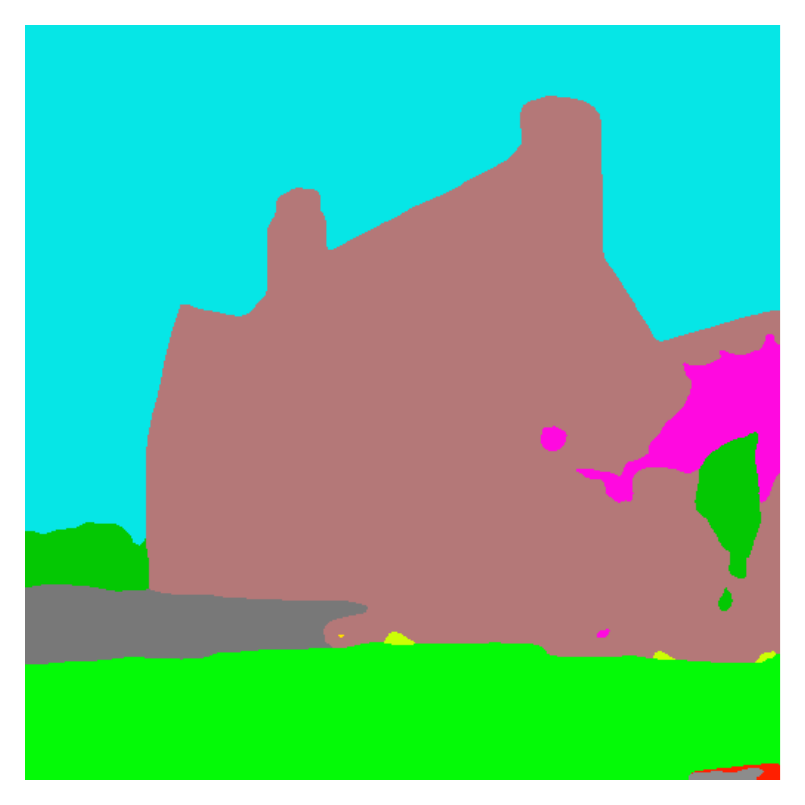} & \includegraphics[width=\newl]{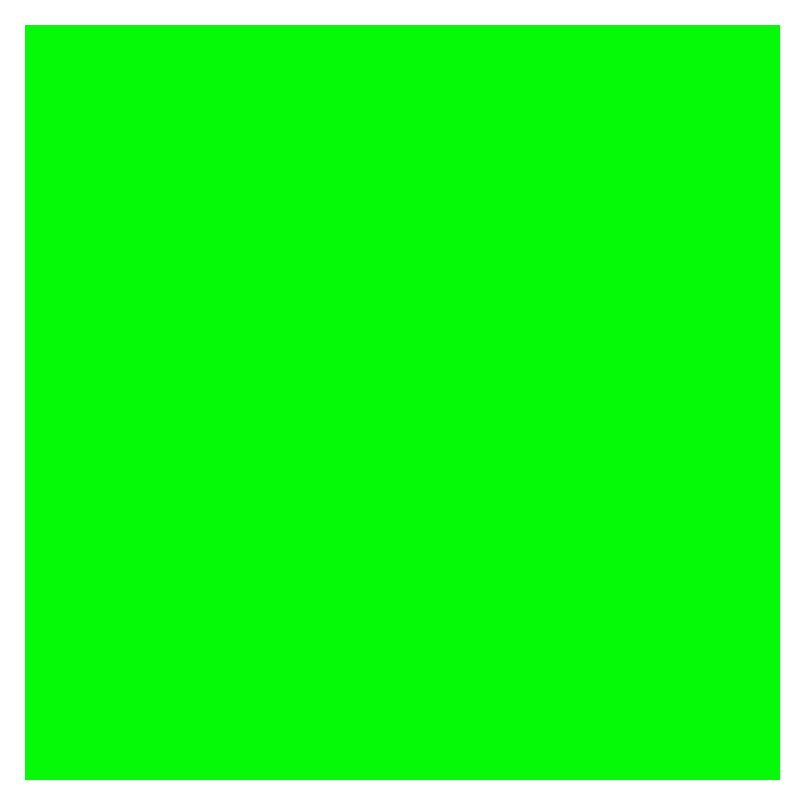} & \includegraphics[width=\newl]{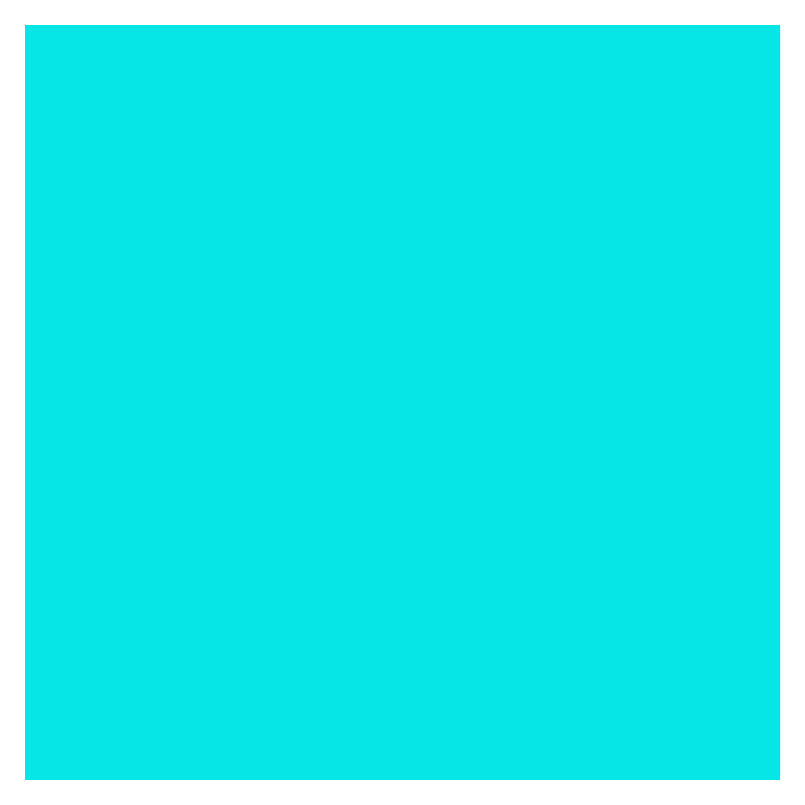} & \includegraphics[width=\newl]{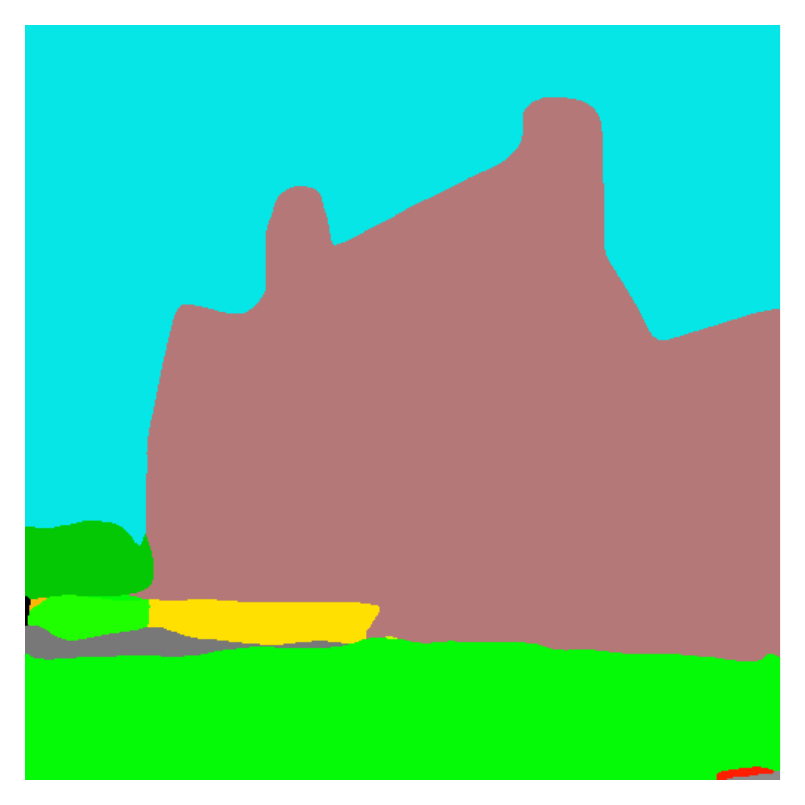} & \includegraphics[width=\newl]{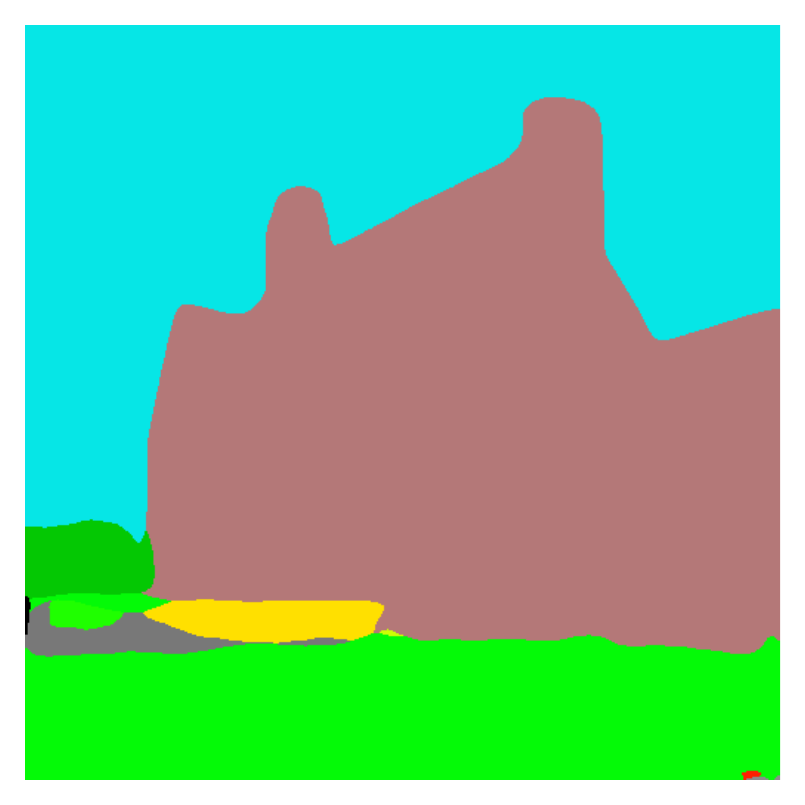} & \includegraphics[width=\newl]{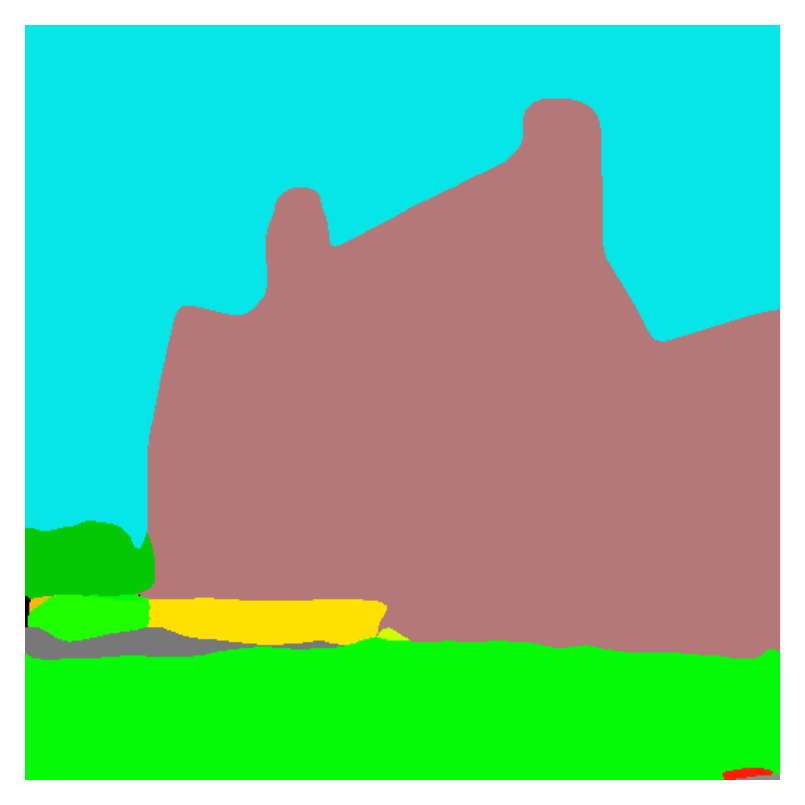}
\end{tabular}
\caption{\label{fig:teaser}\textbf{Effect of adversarial attacks on semantic segmentation models.} For a validation image of \ade (first column, with ground truth mask), we show the image perturbed by targeted $\ell_\infty$-attacks ($\epsilon_\infty=2/255$, target class ``grass'' or ``sky''), and the predicted segmentation. For a clean model the attack completely alters the segmentation map, while our robust model (\upernet+ ConvNeXt-T trained with 5-step \RobAT for 128 epochs) is minimally affected. For illustration, we use targeted attacks, and not untargeted ones as in the rest of the paper. More illustrations in~\cref{app:additional-fig}.}
\end{figure}

\textbf{Strong evaluation of robustness with \multiloss.}
In Sec.~\ref{sec:multiloss_scheme}, we introduce \multiloss (Semantic Ensemble Attack), a reliable robustness evaluation for semantic segmentation for the $\ell_\infty$-threat model 
via an ensemble of our three complementary attacks,
two designed to minimize average pixel accuracy and one 
\hlrev{targeting} \miou.
In the experiments, we show that recent SOTA attacks, SegPGD \cite{gu2022segpgd} and CosPGD \cite{agnihotri2023cospgd}, may significantly overestimate the robustness of semantic segmentation models, in particular regarding \miou. \cref{tab:comp-losses} illustrates, for multiple robust and non-robust models, 
that our individual novel attacks and in particular our ensemble \multiloss consistently outperform  SOTA attacks, with improvements of more than $17.2\%$ in accuracy and $10.3\%$ in \miou.

\textbf{Robust semantic segmentation models with \RobAT.}
It is interesting to note that no adversarially robust semantic segmentation models with respect to $\ell_\infty$-threat model are publicly available. The models of DDC-AT \cite{xu2021dynamic} turned out to be non-robust, see~\cref{tab:comp-losses}. This implies 
that adversarial training (AT) is much harder for semantic segmentation than for image classification, likely due to the much more difficult attack problem.
In fact, to obtain satisfactory robustness with AT we had to increase the number of epochs compared to clean training and use many \nds{(up to 5)} attack steps. However, this yields a high computational cost, that is prohibitive for scaling up to large architectures.
We drastically reduce this cost by leveraging recent progress in robust \imagenet classifiers \cite{debenedetti2022adversarially, singh2023revisiting,liu2023comprehensive}.
In fact, we introduce Pre-trained \imagenet Robust AT (\RobAT), where we initialize
the backbone of the segmentation model with an $\ell_\infty$-robust \imagenet classifier.
This allows us to \textit{i)} reduce the cost of AT \hlrev{(robust \imagenet classifier are widely available e.g. in RobustBench \cite{croce2020robustbench})}, and \textit{ii)} achieve SOTA robustness on 
\hlrev{segmentation datasets.
On \voc \cite{everingham2010pascal},}
\cref{tab:comp-losses} shows that our approach \RobAT (see~\cref{sec:adversarial_training} for details) attains 71.7\% robust average pixel accuracy at attack radius $\epsilon_\infty=8/255$ 
compared to 0.0\% of DDC-AT, with negligible drop in clean performance compared to clean training (92.7\% vs. 93.4\% accuracy, 75.9\% vs 77.2\% mIoU).
\hlrev{For the more challenging \ade \cite{zhou2019semantic} we obtain the first, to our knowledge, robust models, with up to $55.5\%$ robust accuracy at $\epsilon_\infty=4/255$.
Finally, we show that \RobAT consistently outperforms AT across segementation networks (PSPNet \cite{zhao2017pspnet}, \upernet \cite{xiao2018unified}, Segmenter \cite{strudel2021segmenter}).}

\section{Related Work}

\textbf{Adversarial attacks for semantic segmentation.} $\ell_\infty$-bounded attacks on segmentation models have been first proposed by \cite{hendrik2017universal}, which focuses on targeted (universal) attacks, and \cite{arnab2018robustness}, using FGSM \cite{GooShlSze2015} or PGD on the cross-entropy loss. 
Recently, \cite{gu2022segpgd, agnihotri2023cospgd} revisited the loss used in the attack to improve the effectiveness of $\ell_\infty$-bounded attacks, and are closest in spirit to our work (see extended discussion in \cref{sec:discussion_pgd-based_attacks}).
Additionally, there exist attacks for other threat models, including unconstrained, universal and patch attacks \cite{xie2017adversarial, cisse2017houdini, mopuri2018generalizable, shen2019advspade, kang2020adversarial, nesti2022evaluating, rony2022proximal}.
In particular, \cite{rony2022proximal} introduces an algorithm to minimize the $\ell_\infty$-norm of the perturbations such that a fixed percentage of pixels is successfully attacked.
While their threat model is quite different, we adapt our attacks to it and provide a comparison in \cref{sec:comp_almaprox}: especially for robust models, our attacks outperform the method of \cite{rony2022proximal} by large margin. 

\textbf{Robust segmentation models.} Only a few works have developed defenses for semantic segmentation models. \cite{xiao2018characterizing} proposes a method to detect attacks, while stating that adversarial training is hard to adapt for semantic segmentation. Later, DDC-AT~\cite{xu2021dynamic} attempts to integrate adversarial points during training exploiting additional branches of the networks. The seemingly robust DDC-AT  has been shown to be non-robust using SegPGD in \cite{gu2022segpgd} at $\epsilon_\infty=8/255$, whereas we show with SEA (\cref{tab:comp-losses}) that it is non-robust even for $\epsilon_\infty=2/255$ where SegPGD still flags robustness. 
Finally, \cite{cho2020dapas, kapoor2021fourier} present defenses based on denoising the input, with Autoencoders or Wiener filters, to remove perturbations before feeding it to clean models. 
These methods are only tested via attacks with limited budgets, and similar techniques to protect image classifiers have been rendered ineffective with stronger evaluation~\cite{AthEtAl2018, tramer2020adaptive}.

\section{Adversarial Attacks for Semantic Segmentation} \label{sec:background}

\hlrev{First, we discuss the general setup and how to design loss functions for attacks which target either \miou or accuracy of semantic segmentation models.
Then, we introduce our three novel objectives and discuss how to improve the optimization scheme of PGD in this context.
Finally, we present our attack ensemble \multiloss (\cref{sec:multiloss_scheme}) for reliable evaluation of adversarial robustness including our three proposed diverse attacks, which significantly improves over existing SOTA attacks.}

\subsection{Setup}
The goal of semantic segmentation consists in classifying each pixel of a given image into the $K$ available classes (corresponding to different objects or background). We denote by $f:\R^{w \times h \times c} \longrightarrow \R^{w \times h \times K}$ a segmentation model which for an image $\vx$ of size $w \times h$ (and $c$ color channels) returns $\vu = f(\vx)$, where $\vu_{ij}\in \R^K$ contains the logit of each of the $K$ classes for the pixel $\vx_{ij}$. The class predicted by $f$ for $\vx_{ij}$ is given by $\vm_{ij} = \argmax\nolimits_{k=1, \ldots, K} \vu_{ijk}$, and $\vm \in \R^{w \times h}$ is the segmentation map of $\vx$. Given the ground truth map $\vy \in \R^{w \times h}$, the average pixel accuracy of $f$ for $\vx$ is $\frac{1}{w\cdot h}\sum_{i,j} \mathbb{I}(\vm_{ij} = \vy_{ij})$. In the following, we use index $a$ for the pixel $(i,j)$ to simplify the notation.
The goal of an adversarial attack on $f$ is to change the input $x$ such that as many pixels as possible are mis-classified. This can be formalized as the optimization problem

\begin{align}
\maxop_\vdelta \, \sum_{a} \L(f(\vx + \vdelta)_{a}, \vy_{a})
\quad \textrm{s. th.} \quad \norm{\vdelta}_\infty \leq \epsilon, \quad \vx + \vdelta \in [0, 1]^{wh \times c}, \label{eq:adv_opt_diff}
\end{align}
where we use the $\ell_\infty$-threat model for the perturbations $\delta$, $x+\delta$ is restricted to be an image, 
and $\L :\R^K \times \R \longrightarrow \R$ is a differentiable loss whose maximization induces mis-classification. This can then be (approximately) solved by techniques for constrained optimization such as projected gradient descent (PGD) as suggested for image classification in \cite{MadEtAl2018}.

\textbf{Background pixels.}
In semantic segmentation, it is common to exclude pixels that belong to the background class when training or computing the test performance. 
However, it is unrealistic for an attacker to modify only non-background pixels. Thus semantic segmentation models must be robust for all pixels regardless of how they are classified or what their ground-truth label is.
Therefore, we 
train all our models with an additional background class. This has little impact on segmentation performance, see~\cref{app:remove_bakground-test}, and yields a realistic and practically relevant definition of adversarial robustness.

\subsection{How to efficiently attack \miou}\label{sec:mIoU}
In semantic segmentation it is common to use the Intersection over Union (IoU) as performance metric, averaged over classes (\miou). The \miou is typically computed across all the images in the test set and not as average over the images. As this makes image-wise optimization infeasible, a direct optimization of \miou is intractable.
We derive in the following an upper bound on \miou which can be efficiently optimized and which we use as an objective for an attack on \miou.

Let $TP_s$ and $FP_s$ be true and false positive pixels of class $s$ (and accordingly true and false negatives) for all (test) images. For each class $s$ the IoU is given as
\begin{align}
    \textrm{IoU}_s = \frac{TP_s}{TP_s + FP_s + FN_s} \leq \frac{TP_s}{TP_s + FN_s} = \frac{TP_s}{N_s}= \textrm{Acc}_s, 
\end{align}
where 
 Acc$_s$ is the accuracy of the $s$-th class and $N_s=TP_s+FN_s$ the total number of pixels of class $s$ (in the test set). Thus we get 
\begin{align}\label{eq:mIoU-upperBound} \textrm{mIoU}=\frac{1}{K}\sum_{s=1}^K \mathrm{IoU}_s \leq \frac{1}{K}\sum_{s=1}^K \mathrm{Acc}_s=\frac{1}{K}\sum_{s=1}^K \frac{TP_s}{N_s} 
=\frac{1}{K}\sum_{i=1}^I \sum_{s=1}^K \frac{1}{N_s} TP_{si},
\end{align}
where $I$ is the total number of images in the test set and $TP_{si}$ denotes the number of true positives of class $s$ in image $i$ (we use $TP_s=\sum_{i=1}^I TP_{si}$). Thus mIoU is upper-bounded by class-balanced accuracy. We can interpret the last expression as an image-wise weighted accuracy, where correct pixels of class $s$ have an image-independent weight of $\frac{1}{N_s}$. This is an image-wise loss which we can optimize as in Eq.~\eqref{eq:adv_opt_diff}. In practice, to not use any information from the test set, we obtain the number of pixels per class $N_s$ from the training set. We use this upper bound as objective for our \miou-specific attack in~\cref{sec:attacks}.

\input{tableLosses.tex}

\subsection{Why do attacks on semantic segmentation require new loss functions compared to image segmentation?}

The main challenge of an attack on semantic segmentation is to flip the decisions of many pixels \hlrev{($\geq 10^5$)} of an image \hlrev{at the same time}, ideally with only a few iterations. To tackle this problem it is therefore illustrative to consider the gradient of the total loss with respect to the input. We denote by $\vu \in \R^{wh \times K}$ the logit output 
for the full image and by $u_{ak}$ the logit of class $k \in \{1,\ldots,K\}$ for pixel $a$, which yields 
\begin{align}\label{eq:grad}
\nabla_\vx \sum_a \L(u_a,y_a) = \sum_{a} \sum_{k=1}^K \frac{\partial \L}{\partial u_{ak}}(u_a,y_a) \, \nabla_\vx u_{ak}. 
\end{align}
The term $\sum_{k=1}^K \frac{\partial \L}{\partial u_{ak}}(u_a,y_a)$ can be interpreted as the influence of pixel $a$.
The main problem is that successfully attacked pixels (i.e. pixels wrongly classified) typically have non-zero gradients, 
\hlrev{and in some cases, e.g. for the cross-entropy loss, these gradients have the largest magnitude for mis-classified pixels (see next paragraph).}
\hlrev{Thus already misclassified pixels have strong influence on the iterate updates,} and in the worst case prevent that the decisions of still correctly classified pixels are flipped.

The losses we introduce in the following have their own strategy to cope with this problem and yield either implicitly or explicitly a different weighting of the contributions of each pixel.
Some losses are easier described in terms of the predicted probability distribution $\vp \in \R^K$ via the softmax function:  $\vp_r=e^{\vu_r}/ \sum_{k=1}^K e^{\vu_k}$, $k=1,\ldots,K$ ($\vp$ is a function of $\vu$). Moreover, we omit the pixel-index $a$ in the following for easier presentation.

\subsubsection{Why does the cross-entropy loss not work for semantic segmentation?}
The most common choice as objective function in PGD based attacks for image classification is the cross-entropy loss, i.e.
$
\L_\textrm{CE}(\vu, y) =-\log \vp_y = -\vu_{y} + \log\left(\sum_{j=1}^K e^{\vu_j}\right)$.  Maximizing CE-loss is equivalent to minimizing the predicted probability of the correct class. The cross-entropy loss is unbounded, which is problematic for semantic segmentation as misclassified pixels are still optimized instead of focusing on correctly classified pixels. It holds
\[ \frac{\partial \L_\textrm{CE}(\vp,\ve_y)}{\partial \vu_k} =  - \delta_{yk} + \vp_k \; \textrm{ and } \; \frac{K}{K-1} (1-\vp_y)^2 \leq \norm{\nabla_{\vu} \L_\textrm{CE}}^2_2 \leq (1-\vp_y)^2+ 1-\vp_y. \]
The bounds on the gradient norm are monotonically increasing as $\vp_y \rightarrow 0$ (see~\cref{app:JS}).
Therefore maximally misclassified pixels have the strongest influence in Eq.~\eqref{eq:grad} for the CE-loss which explains why it does not work well as an objective for attacks on semantic segmentation, as shown in~\cite{agnihotri2023cospgd, gu2022segpgd}.

\subsection{Novel attacks on semantic segmentation}
\label{sec:attacks}

In the following we introduce our novel attacks, which address the shortcomings of the cross-entropy loss. All of them are specifically designed for semantic segmentation and have complementary properties. 

\begin{description}[leftmargin=0pt,itemsep=2pt,topsep=0pt,parsep=0pt]
\item[\textbf{Jensen-Shannon (JS) divergence:}] the main problem of the cross-entropy loss is that the norm of the gradient is increasing as $\vp_y \rightarrow 0$, that is the more the pixel is mis-classified. We propose to use instead the Jensen-Shannon divergence, a loss which has not been used for adversarial attacks and has properties which make it particularly useful for attacks on semantic segmentation.
Given two distributions $\vq_1$ and $\vq_2$, the Jensen-Shannon divergence is defined as
\[D_\textrm{JS}(\vq_1\, \|\, \vq_2) = \left(D_\textrm{KL}(\vq_1\, \|\, \vm) + D_\textrm{KL}(\vq_2\,\|\, \vm)\right)/ 2,\quad \textrm{with}\quad \vm = (\vq_1 + \vq_2)/2, \]
where $D_\textrm{KL}$ indicates the Kullback–Leibler divergence. Let 
$\ve_y$ be the one-hot encoding of the target $y$. We set 
$\L_\textrm{JS}(\vu, y) = D_\textrm{JS}(\vp \,\|\, \ve_y)$.
As $D_\textrm{JS}$ measures the similarity between two distributions, maximizing $\L_\textrm{JS}$ drives the prediction of the model away from the ground truth $\ve_y$. Unlike the CE loss, the JS divergence is bounded, and thus the influence of 
every pixel is limited. 
The most relevant property is that the gradient of $\L_\textrm{JS}$ vanishes as $\vp_y \rightarrow 0$ (proof in~\cref{app:JS}
)
\[ \lim_{\vp_y \rightarrow 0} \frac{\partial\L_\textrm{JS}(\vu, y)}{\partial \vu_k}=0, \quad \textrm{for} \quad k=1, \ldots, K.\]
Thus 
$\L_\textrm{JS}$ 
automatically down-weights contributions from mis-classified pixels and in turn pixels which are still correctly classified get a higher weight in the gradient in Eq~\eqref{eq:grad}. Note that this property 
is completely undesired for a loss in image classification as this implies that maximally wrong predictions do not lead to gradients, making it very difficult to change very confident wrong predictions. However, for attacks on semantic segmentation 
\hlrev{this conveniently allows us to simultaneously optimize all pixels, without requiring any masking of the misclassified ones as for other attacks (see next paragraph).}
Our $\L_\textrm{JS}$-based attack is in 7 out of the 18 cases considered in Tab.~\ref{tab:comp-losses} the best single attack. In particular, in average pixel accuracy it always outperforms CosPGD \cite{agnihotri2023cospgd}, and SegPGD \cite{gu2022segpgd} in 15 out of 18 cases and is otherwise equal.

\item[\textbf{Masked cross-entropy (unbalanced and balanced):}] we use the masked cross-entropy (MCE) for two attacks, targeting either standard accuracy or \miou. The main idea in order to avoid over-optimizing mis-classified pixels is to apply a mask to the CE loss which excludes misclassified pixels from the loss: 
this gives
\[
\L_\textrm{MCE}(\vu, y) = \mathbb{I}(\argmax_{j=1, \ldots, K} \vu_j = y) \cdot \L_\textrm{CE}(\vu, y), \]
the unbalanced loss addressing average pixel accuracy. 
To minimize the upper bound on \miou in Eq.~\eqref{eq:mIoU-upperBound} we instead use the balanced masked cross-entropy loss
\[ \L_\textrm{MCE-Bal}(\vu, y) = \frac{1}{N_y} \mathbb{I}(\argmax_{j=1, \ldots, K} \vu_j = y) \cdot \L_\textrm{CE}(\vu, y), \]
where we weight the loss of each pixel according to the number of pixels $N_y$ of the ground-truth class in the training images.
The downside of a mask is that the loss becomes discontinuous and ignoring mis-classified pixels can lead to changes which revert back mis-classified pixels into correctly classified ones with the danger of oscillations.
We note that \cite{xie2017adversarial} used masking for a margin based loss and \cite{hendrik2017universal} for targeted attacks to not optimize pixels already classified into the target class with confidence higher than a fixed threshold. 
However, the MCE-loss has not been 
explored for $\ell_\infty$-bounded untargeted attacks, and turns out to be the best among the five single attacks for average pixel accuracy in 11 out of 18 cases, 
with larger margins for the larger radii, 
see~\cref{tab:comp-losses}. Our MCE-Bal loss targeting \miou achieves in 15 out of 18 cases the best \miou of all five single attacks. 

\end{description}

\subsubsection{Performance of our single attacks.}
In \cref{tab:comp-losses} we compare the performance of our three novel attacks, i.e. the JS, MCE, and MCE-Bal losses optimized using APGD \cite{croce2020reliable} with the scheme detailed in \cref{sec:opt}, to CosPGD \cite{agnihotri2023cospgd} and SegPGD \cite{gu2022segpgd} on various robust and non-robust models trained on \voc and \ade.
All attacks have the same budget of 300 iterations.
Regarding average pixel accuracy our attacks are better than the SOTA attacks in 15 out of 18 cases and equal otherwise. Regarding \miou our attacks are better in 17 out of 18 cases and are only 0.1\% worse in the remaining one. 
This highlights that our novel attacks outperform CosPGD and SegPGD, and at the same time are complementary in where they perform best. This motivates our ensemble attack \multiloss in \cref{sec:multiloss_scheme}.

\subsection{Optimization techniques for adversarial attacks on semantic segmentation}\label{sec:opt}

We discuss next how to improve the optimization scheme of the attacks in semantic segmentation, in particular to obtain more efficient algorithms.
\mh{For optimizing the problem in Eq.~\eqref{eq:adv_opt_diff} we use APGD \cite{croce2020reliable}, since it has been shown to outperform the original PGD \cite{MadEtAl2018}.
While it was designed for image classification, it can be applied to general objectives and constraint sets.}

\hlrev{
\textbf{Progressive radius reduction.}
We noticed in early experiments that, when used in the standard formulation, the optimization may at times get stuck, regardless of the objective function. 
At the same time, increasing the radius, i.e. larger $\epsilon_\infty$, reduces robust accuracy eventually to zero meaning the gradient information is valid (there is no gradient masking).
In order to mitigate this problem and profit from the result of larger $\epsilon_\infty$, we run the attack starting with a larger radius and then use its projection onto the feasible set as starting point for the attack with a smaller radius, similar to \cite{croce2021mind} for $\ell_1$-attacks.
We split the budget of iterations into three slots (with ratio $3:3:4$ as in \cite{croce2021mind}) with attack radii $2 \cdot \epsilon_\infty$, $1.5 \cdot \epsilon_\infty$ and $\epsilon_\infty$ respectively.
}

\begin{figure}[t]
    \centering
        \includegraphics[width=.9\columnwidth]{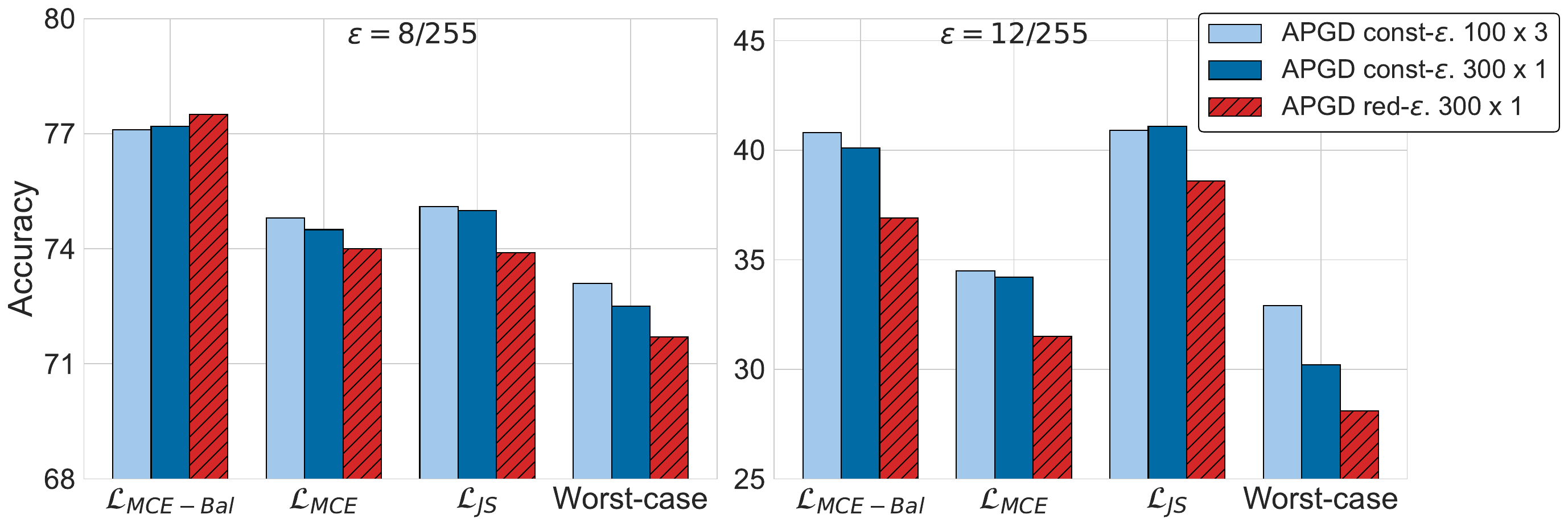}
     
\caption{\textbf{Comparison of const-$\epsilon$ and red-$\epsilon$ optimization schemes.} Attack accuracy for the robust \RobAT UPerNet+\convnext-T model from Table~\ref{tab:comp-losses} on \voc, across different losses for the same iteration budget. The radius reduction (red-$\epsilon$) scheme performs best across all attacks, and it even improves the worst-case over all attacks.}
    \label{fig:loss-wise-teaser}
    
\end{figure}

\textbf{Radius reduction vs more iterations vs restarts.} 
To assess the effectiveness of the scheme with progressive reduction of the radius $\epsilon$ (red-$\epsilon$) described above, we compare it to the standard scheme (const-$\epsilon$) for a fixed budget. For const-$\epsilon$, to match the budget we use either 300 iterations or 3 random restarts with 100 iterations each, and 300 iterations for red-$\epsilon$.
In Fig.~\ref{fig:loss-wise-teaser} we show the robust accuracy achieved by the three attacks with different losses, for $\epsilon_\infty\in\{8/255, 12/255\}$, on our robust \RobAT~\upernet+ \convnext-T model from Table \ref{tab:comp-losses}. Our  progressive reduction scheme red-$\epsilon$ APGD yields the best results (lowest accuracy) for almost every case, with large improvements especially at $\epsilon_\infty=12/255$.
This shows that this scheme is better suited for generating stronger attacks on semantic segmentation models in comparison to more iterations or restarts used in image classification. The same trend holds for \miou, see Fig.~\ref{fig:loss-wise-miou} in~\cref{sec:app-loss-ablation}.

\subsection{Segmentation Ensemble Attack (\multiloss)} \label{sec:multiloss_scheme}
Based on the findings about the complementary properties of the JS and MCE based attacks for different radii $\epsilon_\infty$ targeting average pixel accuracy as well as the MCE-Bal attack targeting \miou, we use all our three attacks in our Segmentation Ensemble Attack (\multiloss) for reliable evaluation of the adversarial robustness of semantic segmentation models. 
\multiloss consists of one run of 300 iterations with red-$\epsilon$ APGD for each of the three losses proposed above, namely $\L_\textrm{MCE}$, $\L_\textrm{MCE-Bal}$, and $\L_\textrm{JS}$, \hlrev{and then taking the worst-case over them (to minimize either accuracy or \miou).}

\textbf{Worst-case attack in semantic segmentation.} While for average pixel accuracy one can simply take for each image the attack which yields the worst image-wise accuracy to get the worst-case accuracy of the ensemble, this is not straightforward for \miou. As the \miou is computed using the results on all test images simultaneously, see Eq. \eqref{eq:mIoU-upperBound}, the computation of the worst-case \miou for a given set of attacks is a combinatorial optimization problem as the mIoU cannot be optimized image-wise (the selection of a different attack changes the numerator \emph{and} denominator).
Since solving this combinatorial optimization problem is out of reach, we use a greedy method to find an approximate solution. We select for all images the attack with the smallest \miou on the test set. Then we iterate the following scheme until there is no improvement in \miou: in each round we shuffle the images and for each image we check if selecting one of the other attacks would result in a lower \miou and, if true, update the selected attack for this image. Typically, only 5-10 rounds are sufficient until such a local optimum is found. We highlight that the computational overhead of this scheme is negligible once one has computed true/false positives and false negatives for each image and attack.

\textbf{Comparison of \multiloss to prior work.} We compare our and SOTA attacks in~\cref{tab:comp-losses}, 
at several radii $\epsilon_\infty$ and on various robust and non-robust models on \voc and \ade, which in turn are based on different architectures: PSPNet, \upernet and Segmenter. 
As our novel single attacks already outperform CosPGD and SegPGD,  the ensemble \multiloss of our attacks performs even better: with a single exception regarding mIoU, \multiloss always achieves the lowest mIoU and accuracy over all six models and different radii.
\multiloss reduces the accuracy compared to CosPGD by up to 28.5\% and mIoU by up to 20.9\% for the models in~\cref{tab:comp-losses}. A similar picture holds for SegPGD where \multiloss always improves in terms of accuracy with maximal gain of $17.2\%$. For mIoU, \multiloss improves in all except one case where SegPGD is $0.1\%$ better, whereas the maximal gain is $10.3\%$.
In summary this shows that \multiloss is a significant improvement in robustness evaluation over previous SOTA attacks and enables a much more reliable robustness evaluation for semantic segmentation.
We analyze \multiloss \hlrev{with several ablation studies} in~\cref{sec:app-loss-ablation}, e.g. a higher budget of 500 iterations \hlrev{does not} yield enough gains to justify the additional computational overhead.

\textbf{Scope of \multiloss.} The popular AutoAttack \cite{croce2020reliable} for robustness evaluation in image classification has, additionally to its three white-box gradient-based attacks, a black-box attack, in particular to be able to attack defenses which are based on gradient masking. As the creation of an efficient and effective black-box attack for semantic segmentation is a non-trivial research question on its own, we leave this to future work. Due to the lack of a black-box algorithm, \multiloss can potentially overestimate robustness for defenses exploiting gradient masking as a defense mechanism. Gradient masking is typically not an issue for models obtained with adversarial training. For this class of models, \multiloss works well as shown by our extensive experiments.

\begin{table}[t]
\centering
\caption{\textbf{Evaluation of our robust models with SEA on \voc and \ade.} For each model and choice of
$\epsilon_\infty$ we report the training details (clean/robust initialization for backbone, number of attack steps and
training epochs) and robust average pixel accuracy (white background) and \miou (grey background)
evaluated with SEA. * indicates results reported in \cite{gu2022segpgd} evaluated with 100 iterations of SegPGD (models are not available) which is much weaker than SEA.}
\label{tab:comp_robust_models_new}
\tabcolsep=1.3pt
\extrarowheight=1.5pt
\newl=7mm
\begin{tabular}{L{24mm} C{8mm} C{7mm} C{14mm}  
*{4}{|
C{\newl}
>{\columncolor{BlueGray}}C{\newl}
}}
Training & Init. & Steps & Backbone & \multicolumn{2}{c}{0} & \multicolumn{2}{c}{4/255} & \multicolumn{2}{c}{8/255} & \multicolumn{2}{c}{12/255} \\
\toprule
\addlinespace[0.5mm]
\multicolumn{12}{l}{\textbf{\voc: PSPNet 50 epochs}}\\
\toprule

DDC-AT~\cite{xu2021dynamic} & clean & 3 &  RN-50 & \textbf{95.1}& \textcolor{NewGray}{\textbf{75.9}} & 0.0& \textcolor{NewGray}{0.0} & 0.0 & \textcolor{NewGray}{0.0} & 0.0 & \textcolor{NewGray}{0.0} 
\\
AT~\cite{xu2021dynamic} & clean & 3 &  RN-50 & 94.0& \textcolor{NewGray}{74.1} & 0.0& \textcolor{NewGray}{0.0} & 0.0 & \textcolor{NewGray}{0.0} & 0.0 & \textcolor{NewGray}{0.0} 
\\
SegPGD-AT~\cite{gu2022segpgd}
& clean & 7 &  RN-50
&-- & \textcolor{NewGray}{74.5}& -- & \textcolor{NewGray}{--}& -- & \textcolor{NewGray}{17.0}\SP{*}& -- & \textcolor{NewGray}{--} 
\\

\RobAT (ours) & robust & 5 &  RN-50 & 90.6 
& \textcolor{NewGray}{68.9}& 81.5  & \textcolor{NewGray}{47.7} & 50.6  & \textcolor{NewGray}{11.2} & 12.9 & \textcolor{NewGray}{1.4} 
\\
\midrule
\addlinespace[0.5mm]
\multicolumn{12}{l}{\textbf{\voc: UPerNet 50 epochs}}\\
\toprule

AT (ours)  & clean & 5 & CN-T  &  91.9 & \textcolor{NewGray}{73.1} & 86.7 & \textcolor{NewGray}{59.0} & 65.3 & \textcolor{NewGray}{28.2} & 22.0 & \textcolor{NewGray}{4.7} 
\\

\RobAT (ours)  & robust & 5 & CN-T  & \textbf{92.7}  & \textcolor{NewGray}{\textbf{75.2}}  & \textbf{88.6}  &\textcolor{NewGray}{\textbf{64.9}}  & \textbf{71.7}  & \textcolor{NewGray}{\textbf{34.6}}  & \textbf{28.1} & \textcolor{NewGray}{\textbf{5.5}} 
\\

\midrule

AT (ours)  & clean & 5 & CN-S & 92.4 & \textcolor{NewGray}{74.6} & 88.1 & \textcolor{NewGray}{61.6} & 68.5 & \textcolor{NewGray}{30.5} & 23.9 & \textcolor{NewGray}{4.3}\\

\RobAT (ours)  & robust & 5 & CN-S & \textbf{93.1} & \textcolor{NewGray}{\textbf{76.6}} & \textbf{89.1} &\textcolor{NewGray}{\textbf{66.0}} & \textbf{71.0} & \textcolor{NewGray}{\textbf{36.4}} & \textbf{27.6} & \textcolor{NewGray}{\textbf{6.2}} 
\\
\midrule
\addlinespace[0.5mm]
\multicolumn{12}{l}{\textbf{\ade:  UPerNet 128 epochs}}\\
\toprule

AT (ours) & clean &  5 & CN-T
& 68.0 & \textcolor{NewGray}{26.1} & 52.2 & \textcolor{NewGray}{12.8} & 24.5 & \textcolor{NewGray}{3.3} & 2.5 & \textcolor{NewGray}{0.2}\\
\RobAT (ours) & robust &  5 & CN-T
& \textbf{70.5} & \textcolor{NewGray}{\textbf{31.7}} & \textbf{55.5} & \textcolor{NewGray}{\textbf{17.2}} & \textbf{26.4} & \textcolor{NewGray}{\textbf{4.9}} & \textbf{3.1} & \textcolor{NewGray}{\textbf{0.4}}   \\
\midrule
\addlinespace[0.5mm]
\multicolumn{12}{l}{\textbf{\ade:  Segmenter 128 epochs}}\\
\toprule
AT (ours)  & clean & 5 &  ViT-S
& 67.7 & \textcolor{NewGray}{26.8}  & 49.0 & 11.1& 25.1 & \textcolor{NewGray}{3.1} & 4.8 & \textcolor{NewGray}{0.4}\\ 
\RobAT (ours) & robust & 5 & ViT-S
& \textbf{69.1} & \textcolor{NewGray}{\textbf{28.7}} & \textbf{55.3 }& \textcolor{NewGray}{\textbf{14.9}} & \textbf{33.3} & \textcolor{NewGray}{\textbf{5.3}} & \textbf{8.9} & \textcolor{NewGray}{\textbf{1.1}}\\
\bottomrule
\end{tabular}\end{table}

\input{main_figure}

\begin{table}[t]
\centering
\caption{\textbf{Ablation study AT vs \RobAT.} We show the effect of varying the number of attack steps and training epochs on the robustness (measured with \acc and \hlc[BlueGray]{\miou} at various radii) 
of the models trained with AT and \RobAT. Our \RobAT achieves similar or better robustness than AT at significantly reduced computational cost for all datasets and architectures.}
\label{tab:ablation}
\tabcolsep=1.5pt
\extrarowheight=1.5pt
\newl=7.8mm
\begin{tabular}{L{18mm} C{9mm} C{9mm} C{9mm} | 
*{4}{|
C{\newl}
>{\columncolor{BlueGray}}C{\newl}
}}
Training & Init. & Steps & Ep. & \multicolumn{2}{c}{0} & \multicolumn{2}{c}{4/255} & \multicolumn{2}{c}{8/255} & \multicolumn{2}{c}{12/255} \\
\toprule
\addlinespace[2mm]

\multicolumn{12}{l}{\textbf{\voc: \upernet with \convnext-T backbone}} \\
\toprule
 AT  & clean & 2 & 50 & \textbf{93.4} & \textcolor{NewGray}{\textbf{77.4}} & 2.6 & \textcolor{NewGray}{0.1} & 0.0 & \textcolor{NewGray}{0.0}& 0.0 & \textcolor{NewGray}{0.0} 
 \\
\RobAT  & robust & 2 & 50 & 92.9 & \textcolor{NewGray}{75.9} & \textbf{86.7} & \textcolor{NewGray}{\textbf{60.8}} & \textbf{50.2} & \textcolor{NewGray}{\textbf{21.0}} & \textbf{9.3} & \textcolor{NewGray}{\textbf{2.4}} 
\\
 AT   & clean & 2 & 300 & 93.1 & \textcolor{NewGray}{76.3} & 86.5 & \textcolor{NewGray}{59.6} & 44.1 & \textcolor{NewGray}{16.6} & 4.6 & \textcolor{NewGray}{0.1} 
\\
\midrule
 AT   & clean & 5 & 50 &   91.9 & \textcolor{NewGray}{73.1} & 86.7 & \textcolor{NewGray}{59.0} & 65.3 & \textcolor{NewGray}{28.2} & 22.0 & \textcolor{NewGray}{4.7} 
\\
\RobAT  & robust & 5 & 50 & 92.7  & \textcolor{NewGray}{75.2}  & \textbf{88.6}  &\textcolor{NewGray}{\textbf{64.9}}  & \textbf{71.7}  & \textcolor{NewGray}{\textbf{34.6}}  & \textbf{28.1} & \textcolor{NewGray}{\textbf{5.5}} \\

 AT   & clean & 5 & 300 & \textbf{92.8} & \textcolor{NewGray}{\textbf{75.5}} & \textbf{88.6} & \textcolor{NewGray}{64.8} & 71.5 & \textcolor{NewGray}{\textbf{35.1}} & 23.7 & \textcolor{NewGray}{5.1} 
\\
\midrule
\addlinespace[2mm]

\multicolumn{12}{l}{\textbf{\ade: \upernet with \convnext-T backbone}}\\
\toprule

 AT  & clean &  2 & 128 & \textbf{73.4} & \textcolor{NewGray}{\textbf{36.4}} & 0.4 & \textcolor{NewGray}{0.0} & 0.0 & \textcolor{NewGray}{0.0} & 0.0 & \textcolor{NewGray}{0.0}\\
 \RobAT & robust &  2 & 128
& 72.0 & \textcolor{NewGray}{34.7} & \textbf{45.9} & \textcolor{NewGray}{\textbf{15.1}} & \textbf{5.9} & \textcolor{NewGray}{\textbf{1.9}}& 0.0 & \textcolor{NewGray}{0.0}  \\ 

\midrule
\RobAT  & robust &  5 & 32
& 68.8 & \textcolor{NewGray}{25.2} & \textbf{56.2} & \textcolor{NewGray}{13.7} & \textbf{30.6} & \textcolor{NewGray}{\textbf{5.0}} & \textbf{4.5} & \textcolor{NewGray}{\textbf{0.6}}\\
AT  & clean &  5 & 128
& 68.0 & \textcolor{NewGray}{26.1}  & 52.2 &\textcolor{NewGray}{12.8} & 24.6 & \textcolor{NewGray}{3.3} & 2.5 & \textcolor{NewGray}{0.2} \\
\RobAT  & robust &  5 & 128
& \textbf{70.5} & \textcolor{NewGray}{\textbf{31.7}} & 55.5 & \textcolor{NewGray}{\textbf{17.2}} & 26.4 & \textcolor{NewGray}{4.9} & 3.1 & \textcolor{NewGray}{0.4}   \\
\midrule
\addlinespace[2mm]

\multicolumn{12}{l}{\textbf{\ade: Segmenter with ViT-S backbone}} \\
 \toprule
 \RobAT  & robust & 5 & 32 & 
68.1 & \textcolor{NewGray}{26.0} & \textbf{55.5} & \textcolor{NewGray}{14.2} & \textbf{34.2} & \textcolor{NewGray}{\textbf{5.3}} &8.7& \textcolor{NewGray}{0.9} \\

 AT  & clean & 5 & 128 & 
67.7 & \textcolor{NewGray}{26.8} &49.0 & \textcolor{NewGray}{11.1} & 25.1 & \textcolor{NewGray}{3.1} & 4.8 & \textcolor{NewGray}{0.4} \\

 \RobAT  & robust & 5 & 128
& \textbf{69.1} & \textcolor{NewGray}{\textbf{28.7}} & 55.3& \textcolor{NewGray}{\textbf{14.9}} & 33.3 & \textcolor{NewGray}{\textbf{5.3}} & \textbf{8.9} & \textcolor{NewGray}{\textbf{1.1}}\\
\bottomrule 
\end{tabular} \vspace{-3pt}
\end{table}

\section{Adversarially Robust Segmentation Models} \label{sec:adversarial_training}

Adversarial training (AT) \cite{MadEtAl2018} and its variants are the established techniques to get adversarially robust image classifiers. One major drawback is the significantly longer training time due to the adversarial attack steps ($k$ steps imply a factor of $k+1$ higher cost).
This can be prohibitive for large backbones and decoders in semantic segmentation models. 
As a remedy we propose Pre-trained \imagenet Robust Models AT (\RobAT), which can reduce the training time by a factor of upto 6 while improving the SOTA of robust semantic segmentation models.

\textbf{Experimental setup.} We use PGD for adversarial training with the cross-entropy loss and an $\ell_\infty$-threat model of radius $\epsilon_\infty=4/255$ (as used by robust classifiers on \imagenet). We tried AT with the losses from~\cref{sec:attacks} but got no improvements over the cross-entropy loss. This is to be expected as the good properties of the JS-divergence for attacking a semantic segmentation model are also detrimental for training such a model. For training configuration, we mostly follow standard practices for each architecture \cite{zhao2017pspnet, liu2022convnet, strudel2021segmenter}, in particular for the number of epochs (see~\cref{sec:exp_details} for
training and evaluation details).
All robustness evaluations are done with \multiloss on the entire validation set.

\subsection{\RobAT: robust models via robust initialization}
When training clean and robust semantic segmentation models it is common practice to initialize the backbone with a clean classifier pre-trained on \imagenet~\cite{strudel2021segmenter, liu2022convnet}, which are not robust to adversaries even at small radii (e.g. $\epsilon_\infty=
1/255$), see the top row in Fig.~\ref{fig:examples_voc} for an illustration.
In contrast, we propose for training robust models to use $\ell_\infty$-robust \imagenet classifiers (see \cref{app:backbone-detail} for specific models) as initialization of the backbone, whereas the decoder is always initialized randomly: we denote this approach as Pre-trained ImageNet Robust Models AT (\RobAT).
The resulting \RobAT models are robust even at a large radius such as 
$8/255$, see bottom row in Fig.~\ref{fig:examples_voc}.
This seemingly small change has huge impact on the robustness of the model.
We show in~\cref{tab:comp_robust_models_new} a direct comparison of AT and \RobAT using the same number of adversarial steps and training epochs across different decoders and architectures for the backbones. In all cases \RobAT outperforms AT in clean and robust accuracy. Up to our knowledge these are also the first robust models for \ade.
Regarding \voc, we note that DDC-AT from \cite{xu2021dynamic} as well as their AT-model are completely non-robust.
SegPGD-AT, trained for $\epsilon_\infty=8/255$ and with the larger attack budget of 7 steps, is seemingly more robust than our PSPNet, but it is evaluated with only 100 iterations of SegPGD, which is significantly weaker than \multiloss and therefore its robustness could be overestimated (this model is not publicly available).
However, our \upernet+ \convnext-T (CN-T) outperforms SegPGD-AT by at least 17.6\% in \miou at $\epsilon_\infty=8/255$ even though it is trained for $4/255$.
Interestingly, the large gains in robustness do not degrade much the clean performance, which is a typical drawback of adversarial training.  Our results for ConvNeXt-S (CN-S)  show that this also scales to larger backbones.

\subsection{Ablation study of AT vs \RobAT} \label{sec:ablation_main}

In \cref{tab:ablation} we provide a detailed comparison of AT vs \RobAT for different number of adversarial steps and training epochs.
On \voc, for 2 attack steps, AT with clean initialization for 50 epochs does not lead to any robustness. This is different to image classification where 50 epochs 2-step AT are sufficient to get robust classifiers on \imagenet \cite{singh2023revisiting}. In contrast 2-step \RobAT yields a robust model and even outperforms 300 epochs of AT with clean initialization in terms of robustness at all $\epsilon_\infty$ while being 6 times faster to train. This shows the significance of the initialization.
For 5 attack steps we see small improvements of robustness at $4/255$ compared to 2 steps but much larger ones at $8/255$. Again, 50 epochs of \RobAT mostly outperform the 300 epochs of AT with clean initialization.
Our findings generalize to \ade and across architectures (\upernet+ ConvNeXt-T and Segmenter+ViT-S): 32 epochs of \RobAT outperform 128 epochs of AT with clean initialization in terms of robustness (4 times faster). 
Robust \imagenet classifiers are now available for different architectures and sizes, thus \RobAT should become standard to train robust semantic segmentation models.

\section{Conclusion}
We have proposed novel attacks for semantic segmentation which take into account the specific properties of this problem. We obtain significant improvements over SOTA attacks and  \multiloss, an ensemble of our attacks, yields by large margin the best results regarding accuracy \emph{and} mIoU.
Moreover, we show how to train segmentation models with SOTA robustness, even at a significantly reduced cost 
by using adversarially pre-trained \imagenet classifiers. We hope that the availability of our 
 robust models, 
together with code for training and attacks, will foster research in robust semantic segmentation. 

\textbf{Limitations.} We consider \multiloss an important step towards a strong robustness evaluation of semantic segmentation models. However, similar to AutoAttack \cite{croce2020reliable}, white-box attacks should be complemented by strong black-box attacks which we leave to future work.
Moreover, several techniques, e.g. using different loss functions, unlabeled and synthetic data, adversarial weight perturbations, etc., have been shown effective to achieve more robust classifiers, and might be beneficial for segmentation too, but
this is out of scope for this work.

\section*{Acknowledgements}
We thank the International Max Planck Research School for Intelligent Systems (IMPRS-IS) for supporting NDS.
We acknowledge support from the Deutsche Forschungsgemeinschaft (DFG, German
Research Foundation) under Germany’s Excellence Strategy
(EXC number 2064/1, project number 390727645), as well
as in the priority program SPP 2298, project number 464101476.
We are also thankful for the support of Open Philanthropy and the Center for AI Safety Compute Cluster. Any opinions, findings,
and conclusions or recommendations expressed in this material are those of the author(s) and do not
necessarily reflect the views of the sponsors.

\bibliographystyle{splncs04}

\input{refs.bbl}
\ifshowappendix
\clearpage
\appendix

\section*{Contents of the Appendix}
\begin{enumerate}
\itemindent=5pt
\item Broader Impact
\item Appendix~\ref{app:JS} \ldots  Omitted proofs

   \item Appendix~\ref{sec:exp_details} \ldots   Experimental and evaluation details     
   \item Appendix~\ref{sec:additional_exps} \ldots   Additional \multiloss experiments, ablations and comparisons
     \item Appendix~\ref{app:additional-fig} \ldots   Visualizations of adversarial images generated by \multiloss
     \end{enumerate}

\section*{Broader Impact}

We propose new techniques to test the robustness of segmentation models to adversarial attacks. While we consider it important to estimate the vulnerability of existing systems, such methods might potentially be used by malicious actors.
However, we also provide insights on how to obtain, at limited computational cost, models which are robust to such perturbations.

\section{Proof of the Properties of Cross-Entropy, and the Jensen-Shannon-Divergence loss}\label{app:JS}

\textbf{Cross-entropy loss:}\\
The cross-entropy is given as: $\L_\textrm{CE}(\vp,e_y)=-\log \vp_y$, and has gradient 
\[ \frac{\partial \L_\textrm{CE}(\vu,e_y)}{\partial u_t} =  - \delta_{yt} + \vp_t(u).\]
We note that
\[ \norm{\nabla_u \L_\textrm{CE}(\vu,e_y)}^2_2 = \sum_{t \neq y} \vp_t^2 + (1-\vp_y)^2.\]
As $0\leq \vp_t\leq 1$, it holds
\[ \sum_{t \neq y} \vp_t^2 \leq \sum_{t \neq y} \vp_t = 1-\vp_y.\]
Moreover, the point of minimal $\ell_2$-distance on the surface of the $\ell_1$-ball with radius $1-\vp_y$ has equal components, and thus
\[ \sum_{t \neq y} \vp_t^2 \geq \frac{(1-\vp_y)^2}{K-1},\]
which yields 
\[ \frac{K}{K-1} (1-\vp_y)^2 \leq \norm{\nabla_u \L_\textrm{CE}(\vu,e_y)}^2_2 \leq 1-\vp_y + (1-\vp_y)^2.\]
We note that both lower and upper bounds are monotonically increasing as $\vp_y \rightarrow 0$.

\textbf{Jensen-Shannon divergence:}\\
The Jensen-Shannon-divergence between the predicted distribution $p$ and the label distribution $q$ is given by 
\[D_\textrm{JS}(\vp\, \|\, \vq) = \left(D_\textrm{KL}(\vp\, \|\, \vm) + D_\textrm{KL}(\vq\,\|\, \vm)\right)/ 2,\quad \textrm{with}\quad \vm = (\vp + \vq)/2, \]
Assuming that we have a one-hot label encoding $\vq=e_y$ (where $e_y$ is the $y$-th cartesian coordinate vector), one gets
\[D_\textrm{JS}(\vp\, \|\, e_y)=\frac{1}{2}\log\left(\frac{2}{1+\vp_y}\right) + \frac{1}{2}\sum_{i=1}^K \vp_i \log\left(\frac{2\vp_i}{\delta_{yi}+\vp_i}\right).\]
Then
\begin{align*}
\frac{\partial D_\textrm{JS}(\vp\, \|\, e_y)}{\partial \vp_r} &= \frac{1}{2}\left[- \frac{1}{1+\vp_y}\delta_{yr} + \log\left(\frac{2\vp_r}{\delta_{yr}+\vp_r}\right) + 1 -\frac{\vp_r}{\delta_{yr}+\vp_r}\right] \\
&= \frac{1}{2}\begin{cases} \log\left(\frac{2\vp_y}{1+\vp_y}\right) & \textrm{ if } r=y,\\ \log(2) & \textrm{ else}.\end{cases}\end{align*}
Given the logits $u$ we use the softmax function
\[ \vp_r = \frac{e^{u_r}}{\sum_{t=1}^K e^{u_t}}, \quad r=1,\ldots,K,\]
to obtain the predicted probability distribution $\vp$. One can compute:
\[ \frac{\partial \vp_r}{\partial u_t} = \delta_{rt}\vp_t - \vp_r \vp_t \quad \Longrightarrow \quad \sum_{r=1}^K  \frac{\partial \vp_r}{\partial u_t}=0\]
Then 
\begin{align*} 
\frac{\partial D_\textrm{JS}(\vp\, \|\, e_y)}{\partial u_t}&=\sum_{r=1}^K \frac{\partial D_\textrm{JS}(\vp\, \|\, e_y)}{\partial \vp_r} \frac{\partial \vp_r}{\partial u_t} = \frac{1}{2}\left[\log\left(\frac{2\vp_y}{1+\vp_y}\right) \frac{\partial \vp_y}{\partial u_t} + \log(2) \sum_{r\neq y} \frac{\partial \vp_r}{\partial u_t}\right]\\
&= \frac{1}{2}\left(\log\left(\frac{2\vp_y}{1+\vp_y}\frac{\partial \vp_y}{\partial u_t}\right)
- \log(2)\frac{\partial \vp_y}{\partial u_t}\right) \\
&= \frac{1}{2}\left(\log\left(\frac{\vp_y}{1+\vp_y}\right)\left[\delta_{yt}\vp_y-\vp_y\vp_t \right]\right)\\
&=\frac{1}{2}\left(\vp_y  \log\left(\frac{\vp_y}{1+\vp_y}\right)
\left[\delta_{yt}-\vp_t \right]\right)
\end{align*}
Noting that $\lim_{x \rightarrow 0}x \log(x)=0$ we get the result that: $\lim\limits_{\vp_y \rightarrow 0} \frac{\partial D_\textrm{JS}(\vp\, \|\, e_y)}{\partial u_t}=0.$
 
Thus the $\L_\textrm{JS}$ loss automatically down-weights contributions from mis-classified pixels and thus pixels which are still correctly classified get a higher weight in the gradient.

\textbf{Discussion.}
The (theoretical) discussion of benefits and weaknesses for each loss in Sec.~\ref{sec:attacks} suggests that one main difference among losses is how they balance the weight of different pixels in the objective function.
On one extreme, the plain cross-entropy maximizes the loss for all pixels independently of whether they are misclassified, and assigns them the same importance. 
Conversely, the masked losses exclude (via the mask) the misclassified pixels from the objective function, with the danger of reverting back the successful perturbations. As middle ground, losses like the JS divergence assign a weight to each pixel based on how ``confidently'' they are misclassified. We conjecture that for radii where robustness is low, masked losses help focusing on the remaining pixels, and already misclassified pixels are hardly reverted since they are far from the decision boundary.
Conversely, at smaller radii achieving confident misclassification is harder (since the perturbations are smaller), and most pixels are still correctly classified or misclassified but close to the decision boundary: then it becomes more important to balance all of them in the loss, hence losses like JS divergence are more effective. This hypothesis is in line with the empirical results in \cref{tab:comp-losses} and~\cref{tab:multiloss_individual_perf}.

\section{Experimental Details} \label{sec:exp_details}

We here provide additional details about both attacks and training scheme used in the experiments in the main part. 

\subsection{Attacks for semantic segmentation}
\label{sec:att_params}
\textbf{Baselines.} Since \cite{gu2022segpgd, agnihotri2023cospgd} do not provide code for their methods, we re-implement both SegPGD and CosPGD following the indications in the respective papers and personal communication with the authors of CosPGD.
In the comparison in 
Table~\ref{tab:comp-losses}, we use PGD with step size (8e-4, 9e-4, 1e-3, 2e-3, 3e-3, 5e-3, 6e-3) for radii (0.25/255, 0.5/255, 1/255, 2/255, 4/255, 8/255, 12/255) resp. for both CosPGD and SegPGD for 300 iterations each. The step size selection was done via a small grid-search in [2e-3, 3e-3, 5e-3, 6e-3, 1e-4] for $\eps=\nicefrac{4}{255}$ and $\nicefrac{8}{255}$, the values for other radii were extrapolated from these. Moreover, at the end we select for each image the iterate with highest loss (strongest yet generated adversary).

\textbf{APGD with masked losses.} Since APGD relies on the progression of the objective function value to e.g. select the step size, using losses which mask the mis-classified pixels might be problematic, since the loss is not necessarily monotonic.
Then, in practice we only apply the mask when computing the gradient at each iteration.

\subsection{Training robust models}
\label{app:train-setup}
In the following, we detail the employed network architectures, as well as our training procedure for the utilized datasets. All experiments are conducted in multi-GPU setting with \texttt{PyTorch}~\cite{pytorch} library.
For adversarial training we use PGD at $\epsilon=4/255$ and step size 0.01. While training clean or adversarially, the backbones are initialized with publicly available \imagenet pre-trained models, source of which are listed in Table~\ref{tab:source_models}.
\begin{table}[t]
\centering
\caption{\textbf{
Training and data configurations.} For all the models trained in this work, we list according to the dataset, the training and dataset configurations. Warmup epochs are scaled depending on the total number of epochs. Poly dec. is the polynomially decaying schedule, from~\cite{zhao2017pspnet}. The setup stays the same across all setups of adversarial training (clean init./robust init. or 2 vs 5 step).}
\vspace{2mm}
\label{tab:train-config}
\small
\tabcolsep=1.1pt
\extrarowheight=3.7pt
\begin{tabular}{L{5mm} L{40mm} | C{17mm} C{17mm} |C{17mm} C{17mm} }
& \multirow{2}{*}{Configuration} & \multicolumn{2}{c}{\voc} & \multicolumn{2}{c}{\ade}  \\
\cline{3-6} 
& & PSPNet & \upernet & \upernet & Segmenter \\
\toprule

\parbox[t]{4mm}{\multirow{4}{*}{\rotatebox[origin=c]{90}{\textcolor{NewGray}{DATA}}}} & Base size    &  512 & 512  & 520 & 520\\
& Crop size    &  473x473 & 473x473  & 512x512 & 512x512\\
& Random Horizontal Flip  &  \ding{51} & \ding{51} & \ding{51} & \ding{51} \\ 
& Random Gaussian Blur   &  \ding{51} & \ding{51} & \ding{51} & \ding{51} \\ \midrule
\parbox[t]{4mm}{\multirow{12}{*}{\rotatebox[origin=c]{90}{\textcolor{NewGray}{TRAINING}}}}& Optimizer    &  SGD & AdamW  & AdamW & SGD\\
& Base learning rate  &  5e-4 & 1e-3 & 1e-3 & 2e-3 \\ 
& Weight decay   &  0.0 & 1e-2 & 1e-2 & 1e-2\\ 
& Batch size   &  16x8 & 16x8 & 16x8 & 16x8\\ 
& Epochs   &  50/300 & 50/300 & 32/128 & 32/128\\ 
& Warmup epochs   &  5/30 & 5/30 & 5/20 & 5/20\\ 
& Momentum & 0.9 & 0.9, 0.999 & 0.9, 0.999 & 0.9 \\
& LR schedule   &  poly dec. & poly dec. & poly dec. & poly dec.\\ 
& Warmup schedule   &  linear & linear & linear & linear\\ 
& Schedule power   &  0.9 & 1.0 & 1.0 & 0.9\\ 
& LR ratio (Enc:Dec)   &  1:10 & \ding{55} & \ding{55} & \ding{55}\\ 
& Auxilary loss weight   &  0.4 & 0.4 & 0.4 & -- \\ 

\bottomrule
\end{tabular} 
\end{table}

\begin{table}
\centering
\caption{\textbf{Source of our pre-trained backbones.} We employ the same backbone for both \voc and \ade. The robust column indicates if the backbone used is adversarially robust for \imagenet and we also list the  \imagenet clean and robust accuracy at $\ell_\infty$-radius of $4/255$.}
\label{tab:source_models}
\tabcolsep=1.7pt
\extrarowheight=1.5pt
\begin{tabular}{C{20mm}  C{40mm} | C{12mm} C{10mm} | C{12mm} C{12mm}}
\multirow{2}{*}{Architecture} &
  \multirow{2}{*}{Backbone} & \multirow{2}{*}{Robust} &
  \multirow{2}{*}{Source} & \multicolumn{2}{c}{\imagenet acc.}\\
  & & & & clean & $\ell_\infty$\\
\toprule
\upernet & \convnext-T + ConvStem & \ding{55} & \cite{singh2023revisiting} &80.9\% & 0.0\%\\
\upernet & \convnext-T + ConvStem & \ding{51} & \cite{singh2023revisiting} & 72.7\%& 49.5\%\\
\upernet & \convnext-S + ConvStem & \ding{51} & \cite{singh2023revisiting} &74.1\% & 52.4\%\\
Segmenter & ViT-S & \ding{55} & \cite{rw2019timm} & 81.2\% & 0.0\%\\
Segmenter & ViT-S & \ding{51} & \cite{singh2023revisiting} & 69.2\%& 44.4\%\\
PSPNet & ResNet-50 & \ding{51} & \cite{salman2020adversarially} & 64.0\%& 35.0\%\\
\bottomrule
\end{tabular}
\end{table}

\textbf{Model architectures.} Semantic segmentation model architectures have adapted to use image classifiers in their backbone. \upernet coupled with \convnext~\cite{liu2022convnet} and transformer models like ViT~\cite{dosovitskiy2020image} with Segmenter~\cite{strudel2021segmenter} achieve SOTA segmentation results. We choose \upernet and Segmenter architectures for our experiments with \convnext and ViT  as their respective backbones. For direct comparison to existing robust segmentation works \cite{gu2022segpgd, xu2021dynamic} which only train with a PSPNet~\cite{zhao2017pspnet}, we also train a PSPNet with a ResNet-50 backbone (see Tables~\ref{tab:comp_robust_models_new} and~\ref{tab:multiloss_individual_perf}).
~\cref{tab:train-config} reports the training and data related information about the various architectures and the backbones used.

\textbf{\upernet with \convnext backbone.}
For both clean and robust initialization setups, we use the publically available \imagenet-1k pre-trained weights\footnote{\url{https://github.com/nmndeep/revisiting-at}} 
from~\cite{singh2023revisiting}, which achieve SOTA robustness for $\ell_\infty$-threat model at $\epsilon=4/255$. They propose some architectural changes, notably replacing PatchStem with a ConvStem in their most robust \convnext models, and we keep these changes intact in our \upernet models, we always use a \convnext with ConvStem in this work.
We highlight that \convnext-T, when adversarially trained for classification on \imagenet, attains significantly higher robustness than ResNet-50 at a similar parameter and FLOPs count. For example, at $\epsilon_\infty=4/255$, the \convnext-T we use has 49.5\% of robust accuracy, while ResNet-50 is reported to achieve around 35\% \cite{salman2020adversarially, bai2021are}. This supports choosing \convnext as backbone for obtaining robust segmentation models with the \upernet architecture. 
For \upernet with the \convnext backbone, we use the training setup from~\cite{liu2022convnet}, listed in \cref{tab:train-config}. We also use the same values of 0.4 or 0.3 for stochastic depth coefficient depending on the backbone, same as the original work.\footnote{\url{https://github.com/facebookresearch/ConvNeXt/blob/main/semantic\_segmentation/configs/convnext}} We do not use heavier augmentations and Layer-Decay~\cite{bao2021beit} optimizer as done by~\cite{liu2022convnet}. 

\textbf{Segmenter with ViT backbone.}
Testing with Segmenter also enables a further comparison across model size as Segmenter with a ViT-S backbone is less than half the size (26 million parameters) of \upernet with a \convnext-T backbone (60 million parameters). We define the training setup in Table~\ref{tab:train-config}, which is similar to the setup used by~\cite{strudel2021segmenter}. The decoder is a Mask transformer and is randomly initialized. Note that \cite{strudel2021segmenter} predominantly use \imagenet pre-trained classifiers at resolution of 384x384, whereas we use 224x224 resolution as no robust models at the higher resolution are available. 

\textbf{PSPNet with ResNet backbone.}
As prior works~\cite{xu2021dynamic, gu2022segpgd} use a PSPNet with a ResNet~\cite{he2016deep} backbone to test their robustness evaluations, we also train the same model for the \voc dataset. Both DDCAT~\cite{xu2021dynamic} and SegPGD-AT~\cite{gu2022segpgd} use a split of 50\% clean and 50\% adversarial inputs for training. Instead for \RobAT with PSPNet, we just use adversarial inputs. Due to this change, and due to the fact that we initialize \RobAT with \imagenet pre-trained ResNet-50 (RN50), we slightly deviate from the standard training parameters (learning rate, weight decay, warmup epochs) as in the original PSPNet work~\cite{zhao2017pspnet}. The detailed training setup is listed in~\cref{tab:train-config}.

\textbf{Training setup for \voc.} 
We use the augmentation setup from~\cite{hariharan2011semantic}. Our training set comprises of 8498 images and we validate on the original \voc validation set of 1449 images. Data and training configurations are detailed in~\cref{tab:train-config}. Adversarial training is done with either 2 or 5 steps of PGD with the cross-entropy loss. Unlike some other works in literature, we train for 21 classes (including the background class).

\begin{figure}[t]
    \centering        \includegraphics[width=0.9\columnwidth]{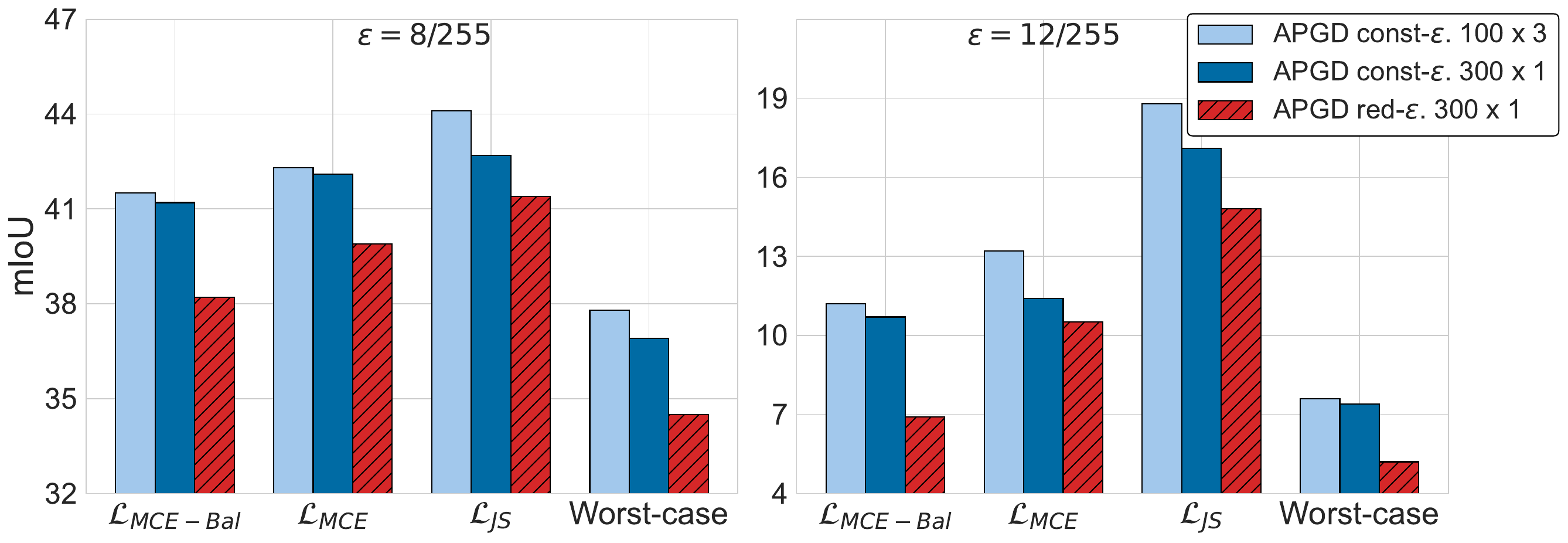}
     \vskip 0.05in
    \caption{\textbf{Comparison of const-$\epsilon$- and red-$\epsilon$ optimization schemes for \miou.} Balanced attack accuracy for the 
    robust \RobAT trained \upernet + \convnext-T model from~\cref{tab:comp_robust_models_new} trained on \voc, across different losses for the same iteration budget. The radius reduction (red-$\epsilon$) scheme performs best across all losses, and $\epsilon_\infty$ and even the worst-case over all losses improves.}
    \label{fig:loss-wise-miou}
\end{figure}

\begin{table}[!t]
\centering
\caption{\textbf{Component analysis for \multiloss.} We show the individual performance (\acc) of the runs of APGD (red-
$\epsilon$) with each loss in \multiloss for both \voc and \ade on 5-step robust models. The best results, among either individual runs, are in \textbf{bold}.}
\label{tab:multiloss_individual_perf}
\small \tabcolsep=2.5pt
\extrarowheight=1.5pt
\newl=10mm
 
\begin{tabular}{L{17mm}| *{3}{|C{\newl}
>{\columncolor{BlueGray}}C{\newl}} |
C{\newl}
>{\columncolor{BlueGray}}C{\newl}}
\multirow{2}{*}{$\epsilon_\infty$} & 
\multicolumn{6}{c|}{individual attacks} 
& \multicolumn{2}{c}{} \\
& \multicolumn{2}{c}{$\L_\textrm{MCE}$}  &\multicolumn{2}{c}{$\L_\textrm{MCE-Bal}$} & \multicolumn{2}{c|}{$\L_\textrm{JS}$} 
& \multicolumn{2}{c}{\multiloss}
\\
\toprule
\addlinespace[2mm]
\multicolumn{9}{l}{\textbf{model: PSPNet ResNet50}, \RobAT, 50 epochs, \voc}\\
\toprule

4/255 & 83.3 & \textcolor{NewGray}{48.6}& 84.7 & \textcolor{NewGray}{49.9} &\textbf{81.8} &  \textcolor{NewGray}{\textbf{47.8}} & 81.5  & \textcolor{NewGray}{47.7} \\
8/255 & \textbf{53.4} & \textcolor{NewGray}{13.5}& 56.4 & \textcolor{NewGray}{\textbf{12.2}}& 53.7 & \textcolor{NewGray}{14.1} & 50.6  & \textcolor{NewGray}{11.2} \\
12/255  & \textbf{14.9} & \textcolor{NewGray}{2.3} &17.6 & \textcolor{NewGray}{\textbf{1.7}} &20.7 & \textcolor{NewGray}{4.1} & 12.9 & \textcolor{NewGray}{1.4} 
\\

\midrule
\addlinespace[2mm]
\multicolumn{9}{l}{\textbf{model: \upernet ConvNeXt-T}, \RobAT, 50 epochs, \voc}\\
\midrule

4/255 & 89.2 &\textcolor{NewGray}{65.9} & 90.4 &\textcolor{NewGray}{67.4}& \textbf{88.7} &\textcolor{NewGray}{\textbf{64.9}}& 88.6 &\textcolor{NewGray}{64.9}\\
8/255 & 74.0 &\textcolor{NewGray}{40.6}& 77.5 &\textcolor{NewGray}{\textbf{38.4}}& \textbf{73.9} &\textcolor{NewGray}{41.3}& 71.7 &\textcolor{NewGray}{34.6}\\
12/255 & \textbf{31.5} &\textcolor{NewGray}{10.3}& 36.9 &\textcolor{NewGray}{\textbf{6.7}}& 38.6 &\textcolor{NewGray}{15.1}& 28.1 &\textcolor{NewGray}{5.5}\\
\midrule
\addlinespace[2mm]
\multicolumn{9}{l}{\textbf{model: \upernet ConvNeXt-S}, \RobAT, 50 epochs, \voc}\\
\toprule

4/255 & 89.7 &\textcolor{NewGray}{67.5}& 90.9 &\textcolor{NewGray}{68.9}& \textbf{89.3} &\textcolor{NewGray}{\textbf{66.7}} & 89.1 &\textcolor{NewGray}{66.0}\\
8/255 & \textbf{73.6} &\textcolor{NewGray}{41.0}& 77.5 &\textcolor{NewGray}{\textbf{36.9}}& 74.3 &\textcolor{NewGray}{42.7}& 71.0 &\textcolor{NewGray}{36.4} \\
12/255 & \textbf{31.2} &\textcolor{NewGray}{10.7}& 36.9 &\textcolor{NewGray}{\textbf{7.5}}& 39.0 &\textcolor{NewGray}{15.6}& 27.6 &\textcolor{NewGray}{6.2} \\ 
\midrule
\addlinespace[2mm]
\multicolumn{9}{l}{\textbf{model: \upernet ConvNeXt-T}, \RobAT, 128 epochs, \ade}\\
\toprule

4/255 & 56.8 &\textcolor{NewGray}{20.0}& 58.2 &\textcolor{NewGray}{\textbf{17.9}}& \textbf{55.9}&\textcolor{NewGray}{18.9}&55.5&\textcolor{NewGray}{17.2} \\ 

8/255 & \textbf{28.5} &\textcolor{NewGray}{6.6}& 31.1 &\textcolor{NewGray}{\textbf{5.3}} & \textbf{28.5}&\textcolor{NewGray}{7.2} & 26.4&\textcolor{NewGray}{4.9} \\ 

12/255 & \textbf{3.7} &\textcolor{NewGray}{\textbf{0.9}}& 4.5 &\textcolor{NewGray}{\textbf{0.9}} & 5.2 &\textcolor{NewGray}{1.1}& 3.1&\textcolor{NewGray}{0.4} \\ 

\midrule
\addlinespace[2mm]
\multicolumn{9}{l}{\textbf{model: \upernet ConvNeXt-S}, \RobAT, 128 epochs, \ade}\\
\toprule

4/255 & 58.6 &\textcolor{NewGray}{20.4}& 59.8 &\textcolor{NewGray}{\textbf{18.6}}& \textbf{57.6}&\textcolor{NewGray}{19.4} &56.8&\textcolor{NewGray}{17.9} \\ 

8/255 & 31.3 &\textcolor{NewGray}{8.1}& 33.3&\textcolor{NewGray}{\textbf{5.8}} & \textbf{30.9}&\textcolor{NewGray}{7.7} & 28.7&\textcolor{NewGray}{5.4} \\ 

12/255 & \textbf{4.6} &\textcolor{NewGray}{1.1}& 5.4&\textcolor{NewGray}{\textbf{0.8}} & 6.2 &\textcolor{NewGray}{1.3} & 3.1 &\textcolor{NewGray}{0.6} \\ 

\midrule
\addlinespace[2mm]
\multicolumn{9}{l}{\textbf{model: Segmenter ViT-S},  \RobAT, 128 epochs, \ade}\\
\toprule

4/255 & 56.9 &\textcolor{NewGray}{17.8}& 57.6 &\textcolor{NewGray}{\textbf{15.6}}& \textbf{55.6}&\textcolor{NewGray}{16.6} &55.3 &\textcolor{NewGray}{14.9} \\ 
8/255 & 36.2&\textcolor{NewGray}{8.5} & 37.8 &\textcolor{NewGray}{\textbf{5.6}}& \textbf{34.2}&\textcolor{NewGray}{7.7} &  33.3&\textcolor{NewGray}{5.4} \\
12/255 & \textbf{10.5}&\textcolor{NewGray}{2.2} & 11.7&\textcolor{NewGray}{\textbf{1.3}} & 11.2&\textcolor{NewGray}{2.2}  & 8.9&\textcolor{NewGray}{1.1} \\

\bottomrule
\end{tabular} 
\end{table}

\textbf{Training setup for \ade.} 
We use the full standard training and validation sets from~\cite{zhou2019semantic}.  Adversarial training is done with either 2 or 5 steps of PGD with the cross-entropy loss. Unlike the original work we train with 151 classes (including the background class).

\subsection{Initialization with pre-trained backbones}
\label{app:backbone-detail}

\RobAT uses pre-trained \imagenet models as an initialization for the backbone.
Note that in the semantic segmentation literature most modern works~\cite{liu2022convnet, strudel2021segmenter} use clean \imagenet pre-trained models as initialization for the backbone, making ours a natural choice.
The robust models are sourced from~\cite{singh2023revisiting} (see~\cref{tab:source_models}),
and more are available e.g. in RobustBench \cite{croce2020robustbench}, thus they do not cost us any additional pre-training.
One can further reduce the cost of pre-training by using robust models trained for either  1-step~\cite{debenedetti2022adversarially} or 2-step~\cite{singh2023revisiting} adversarial training, which is the common budget for robust \imagenet training.
For our \upernet + \convnext-S \RobAT model for \voc, we use the 2-step 50 epoch \imagenet trained model from~\cite{singh2023revisiting} as initialization.
Using such low-cost pre-trained backbones works well, as this model in~\cref{tab:comp_robust_models_new} achieves better or similar robust accuracy as the 300 epoch 2-step \imagenet pre-trained \convnext-T in the same table.

\section{Additional Experiments and Discussion} \label{sec:additional_exps}

We present additional studies of the properties of \multiloss and of the robust models.

\subsection{Analysis of \multiloss}
\label{sec:app-loss-ablation}

\textbf{Effect of reducing the radius.} We complement the comparison of const-$\epsilon$ and red-$\epsilon$ schemes provided in Sec.~\ref{sec:multiloss_scheme} by showing the different robust \miou achieved by the various algorithms.
In Fig.~\ref{fig:loss-wise-miou} one can observe that, consistently with what reported for average pixel accuracy in Fig.~\ref{fig:loss-wise-teaser}, reducing the value of $\epsilon$ (red-$\epsilon$ APGD) outperforms in all cases the other schemes.

\textbf{Analysis of individual components in \multiloss.}
To assess how much each loss contributes to the final performance of \multiloss, we report the individual performance (both accuracy and \miou) at different $\epsilon_\infty$ in~\cref{tab:multiloss_individual_perf}, using robust models on \voc and \ade. We recall that each loss is optimized with 300 iterations of red-$\epsilon$ APGD. A common trend across all models is that either $\L_\textrm{MCE}$ or $\L_\textrm{JS}$ are best individual attacks for accuracy whereas $\L_\textrm{MCE-BAL}$ attacks the \miou the best. Overall, \multiloss significantly reduces the worst case over individual attacks.

\textbf{Analysing attack pairs in \multiloss.}
Further insights into \multiloss are given by looking at how different pairs of the components of \multiloss perform. \cref{tab:pairs} presents such evaluation for the robust UPerNet 
on \voc from \cref{tab:comp-losses}: as expected, MCE + JS yields the best robust aAcc, while the pairs with MCE-Bal have the lowest \miou. Moreover, the worst-case over all losses (\multiloss) gives further improvements.

\begin{table}[!b]
\centering
\caption{\textbf{Effectiveness of pairs of losses.} We evaluate by pairing subset of components of \multiloss by measuring \acc and \hlc[BlueGray]{\miou}. Different pairs perform better or worse depending on perturbation strengths, while \multiloss always yields the strongest attack.} \label{tab:pairs}

\tabcolsep=0.85pt
\extrarowheight=1.5pt
\newl=13mm

\begin{tabular}{L{25mm}| *{3}{|C{\newl}
>{\columncolor{BlueGray}}C{\newl}}}

\makecell{Loss pair} & \multicolumn{2}{c}{4/255}  &\multicolumn{2}{c}{8/255} & \multicolumn{2}{c}{12/255}  \\

\toprule
$\L_\textrm{MCE}$+$\L_\textrm{MCE-Bal}$ & 88.8 & \textcolor{NewGray}{65.1}& 73.2  & \textcolor{NewGray}{35.1}& 31.6&   \textcolor{NewGray}{5.6} \\

$\L_\textrm{MCE}$+$\L_\textrm{JS}$& 88.6 & \textcolor{NewGray}{64.9}& 72.2  & \textcolor{NewGray}{35.2}& 29.4&   \textcolor{NewGray}{6.0} \\

$\L_\textrm{JS}$+$\L_\textrm{MCE-Bal}$ & 88.8 & \textcolor{NewGray}{64.9}& 73.0  & \textcolor{NewGray}{34.7}& 32.6&   \textcolor{NewGray}{5.6} \\
\midrule
\multiloss & 88.6 & \textcolor{NewGray}{64.9}& 71.7  & \textcolor{NewGray}{34.6}& 28.1&   \textcolor{NewGray}{5.5} \\

\bottomrule
\end{tabular}
\end{table}

\begin{figure}[t]
    \centering
        \includegraphics[width=\columnwidth]{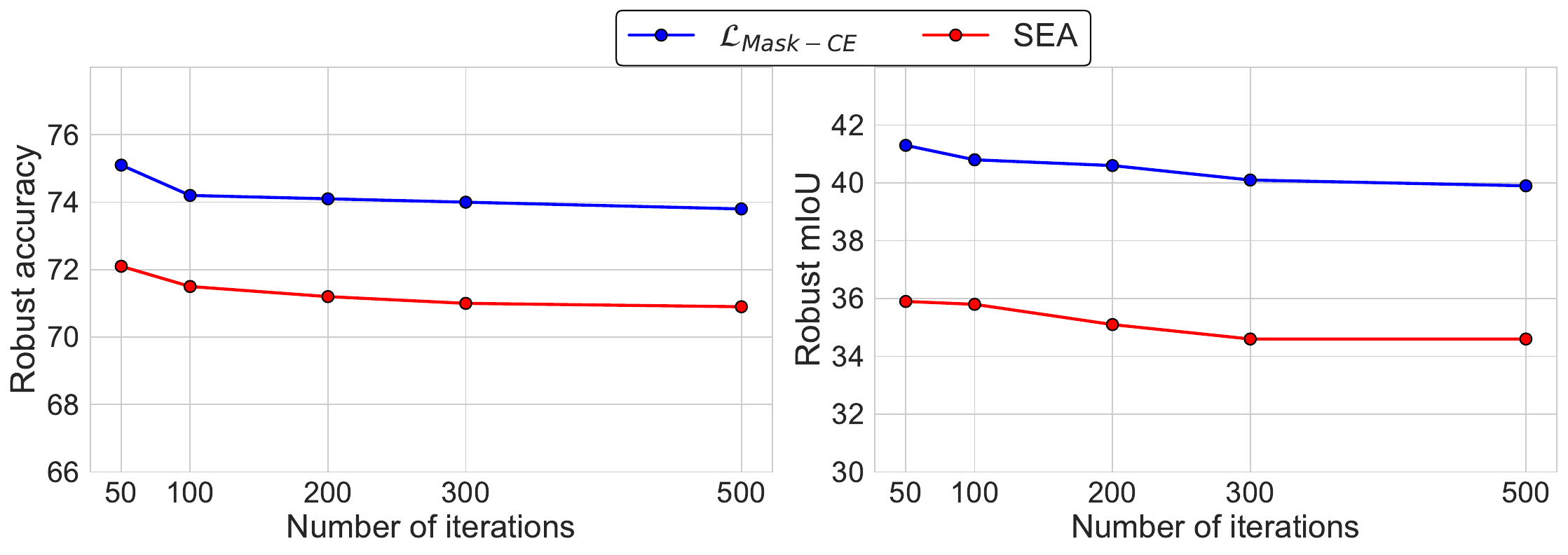}
     \vskip 0.1in
    \caption{\textbf{Influence of number of iterations in \multiloss.} We show robust average pixel accuracy (left) and \miou (right) varying the number of iterations in our attack: 300 iterations give the best compute-effectiveness trade-off.
    We use the 5 step \RobAT \voc trained \convnext-T backbone  \upernet model and the attack is done for $\ell_\infty=\nicefrac{8}{255}$.}
    \label{fig:large_eps_iterations}
\end{figure}

\textbf{More iterations.} We also explore the effect of different number of iterations in \multiloss. In Fig.~\ref{fig:large_eps_iterations} we show the performance (measured by robust accuracy and \miou) of \multiloss with 50, 100, 200, 300 and 500 iterations.
There is  a substantial improvement going from 50 to 300 iterations in all cases. On further increasing the number of attack iterations to 500, the drop in robust accuracy and \miou is around 0.1\% for both $\ell_\infty$ radii of 8/255 and 12/255. 
Since going beyond 300 iterations gives no or minimal improvement for significantly higher computational cost, we fix the number of iterations to 300 in \multiloss.

\textbf{Effect of random seed.} We study the impact of the randomness involved in our algorithm (via random starting points for each run) by repeating the evaluation on our robust model on \voc with 3 random seeds. \cref{tab:random_seed} shows that the proposed \multiloss is very stable across all perturbation strengths. It is also interesting to note that all individual losses have negligible variance across the different runs.

\begin{table}[b]
\centering\small
\caption{\textbf{Stability of \multiloss across different runs.} We report \acc computed on \voc with the 5 step \upernet model trained with  \RobAT. The mean across 3 runs is shown along with the standard deviation. Across components and perturbation strengths, \multiloss has a very low variance over random seeds.}

\label{tab:random_seed}
\tabcolsep=0.85pt
\extrarowheight=1.5pt
\newl=13mm

\begin{tabular}{L{11mm}| *{3}{|C{\newl}
>{\columncolor{BlueGray}}C{\newl}} |
C{\newl}
>{\columncolor{BlueGray}}C{\newl}}

\makecell{$\epsilon_\infty$} & \multicolumn{2}{c}{$\L_\textrm{MCE}$}  &\multicolumn{2}{c}{$\L_\textrm{MCE-Bal}$} & \multicolumn{2}{c|}{$\L_\textrm{JS}$} 
& \multicolumn{2}{c}{\multiloss} \\
\toprule
\addlinespace[2mm]
\multicolumn{9}{l}{\textbf{model:}  \textbf{\upernet ConvNeXt-T},  \RobAT, 50 epochs}\\
\toprule
4/255 & 89.2$\pm$0.2 & \textcolor{NewGray}{65.8$\pm${0.3}}&  90.4$\pm${0.1} & \textcolor{NewGray}{69.0$\pm${0.2}}&88.7$\pm${0.1} &   \textcolor{NewGray}{64.9$\pm${0.4}}&88.6$\pm${0.1}&\textcolor{NewGray}{64.9$\pm${0.4}}\\

8/255 & 73.8$\pm$0.4 &\textcolor{NewGray}{40.8$\pm${0.4}}& 77.5$\pm${0.2} & \textcolor{NewGray}{38.1$\pm${0.2}}&73.9$\pm${0.1} & \textcolor{NewGray}{41.3$\pm${0.0}}& 71.7$\pm${0.3}&\textcolor{NewGray}{34.6$\pm${0.1}}\\
12/255 & 31.5$\pm${0.3} & \textcolor{NewGray}{10.2$\pm${0.2}}&36.9$\pm${0.2} & \textcolor{NewGray}{6.6$\pm${0.1}}&38.6$\pm${0.4} &  \textcolor{NewGray}{15.0$\pm${0.1}}&28.1$\pm${0.2}&\textcolor{NewGray}{5.5$\pm${0.3}}\\

\bottomrule
\end{tabular}
\end{table}

\subsection{Excluding the background class from evaluation}
\label{app:remove_bakground-test}
For \ade, we train clean \upernet+ \convnext-T models in two settings,  
i.e. either ignoring the background class (150 possible classes), which is the standard practice while training clean semantic segmentation models, or to predict it (151 classes).
To measure the effect of the additional background class, we can evaluate the performance of both models with only 150 classes (for the one trained on 151 classes, we can exclude the score of the background class when computing the predictions).
Training on 150 classes achieves (\acc,  \miou) of (80.4\%, 43.8\%), compared to (80.2\%, 43.8\%) for 151.
This shows that we do not lose any performance when training with the background class, and the lower clean accuracy of clean trained \ade models, (\acc, \miou) of (75.5\%, 41.1\%) is due to including the background class when computing the statistics.
This also translates to the robust models trained in the 2 step \RobAT setting.
For the robust model, the two settings have (76.6\%, 37.8\%) and (76.4\%, 37.5\%) (\acc, \miou) respectively.

\subsection{Additional comparisons to existing attacks} \label{sec:comp_almaprox}

\begin{figure}[t]
    \centering
        \includegraphics[width=\columnwidth]{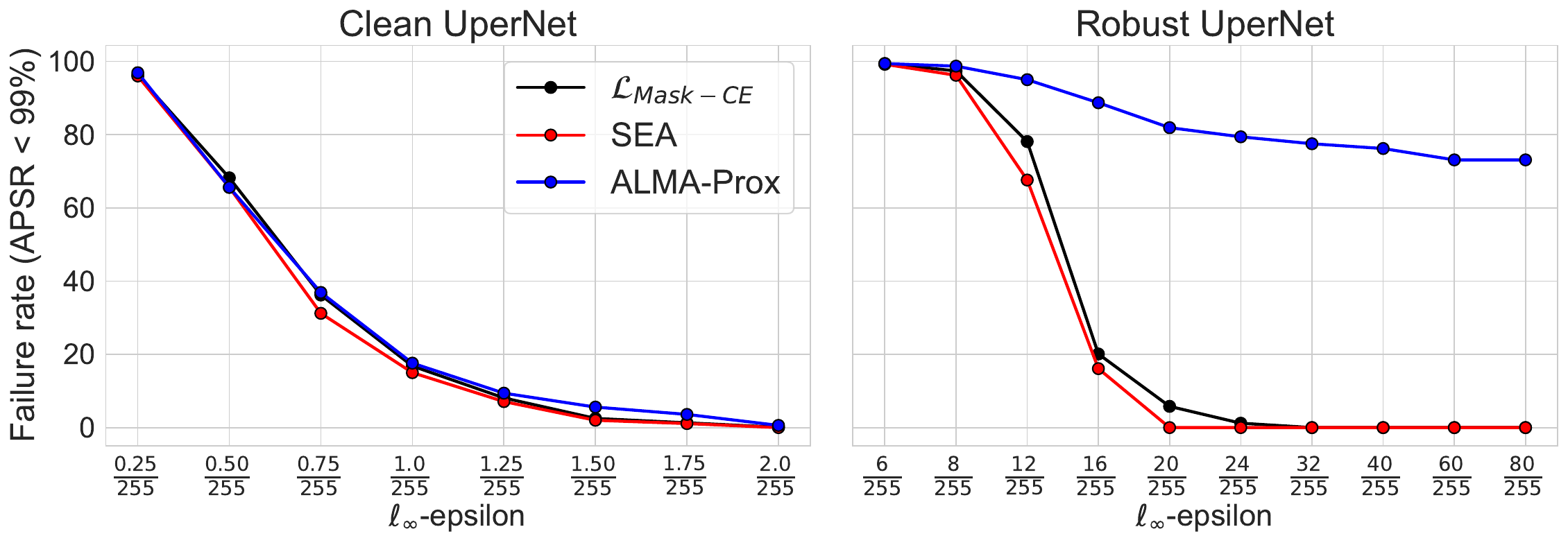}
     \vskip 0.05in
    \caption{\textbf{Comparison to ALMA prox.} We compare APGD with our novel loss ($\L_\textrm{Mask-CE}$) and the ensemble SEA according to the metric used by~\cite{rony2022proximal}, which differs from those (\acc and \miou) we use in the rest of our experiments. In the left plot, the attacks are tested on a clean trained model for the \voc dataset, and in the right plot we test against our robust \RobAT model.}
    \label{fig:alma-comparison}
\end{figure}

Rony et al. \cite{rony2022proximal} have recently proposed ALMA prox as an adversarial attack against semantic segmentation models: its goal is to reach, for each image, a fixed success rate threshold (i.e. a certain percentage of mis-classified pixels, in practice 99\% is used) with a perturbation of minimal $\ell_\infty$ norm.
Thus, the threat model considered by \cite{rony2022proximal} is not comparable to ours, which aims at reducing average pixel accuracy as much as possible with perturbations of a limited size.

In order to provide a comparison of our algorithms to ALMA prox, we measure the percentage of images for which the attack cannot make 99\% of pixels be misclassified with perturbations of $\ell_\infty$-norm smaller than a threshold $\epsilon$ (i.e. the model is considered robust on such images). In this case, lower values indicate stronger attacks.
We show in Fig.~\ref{fig:alma-comparison} the results in such metric, at various $\epsilon$, for ALMA prox (default values, 500 iterations), APGD on the Mask-CE loss (300 iterations) and \multiloss. We test for 160 random images from the \voc dataset using the clean trained \upernet with a \convnext-T backbone in the left plot and 5-step adversarially trained version of the same model in the right plot Fig.~\ref{fig:alma-comparison}.
For the clean model (left plot) the three attacks perform similarly, with a slight advantage of \multiloss at most radii. However, on the robust model (right plot), both APGD on the Mask-CE loss and \multiloss significantly outperform ALMA prox: for example, APGD, which uses even less iterations than ALMA prox, attains 0\% robustness at $32/255$, compared to 77\% of ALMA prox.
This shows that, even considering a different threat model, our attacks are effective to estimate adversarial robustness.

\subsection{Additional discussion of existing PGD-based attacks} \label{sec:discussion_pgd-based_attacks}

Recently, \cite{gu2022segpgd, agnihotri2023cospgd} revisited the loss used in the attack to improve the effectiveness of $\ell_\infty$-bounded attacks, and are closest in spirit to our work.
Since these methods represent the main baseline for our attacks, in the following we briefly summarize their approach to highlight the novelty of our proposed losses.

\begin{description}[leftmargin=0pt,itemsep=2pt,topsep=0pt,parsep=0pt]
\item[\textbf{SegPGD:}] 
\cite{gu2022segpgd} proposes to balance the importance of the cross-entropy loss of correctly and wrongly classified pixels over iterations. In particular, at iteration $t=1, \ldots, T$, they use, with $\lambda(t)=(t-1)/(2T)$, 
\begin{align*}
\L_\textrm{SegPGD}(\vu, y) =  
(&(1 - \lambda(t)) \cdot \mathbb{I}(\argmax_{j=1, \ldots, K} \vu_j = y)\\ &+  \lambda(t) \cdot \mathbb{I}(\argmax_{j=1, \ldots, K} \vu_j \neq y)
)\cdot \L_\textrm{CE}(\vu, y).
\end{align*}
In this way the algorithm first focuses only on the correctly  classified pixels and then progressively balances the attention given to the two subset of pixels: this has the goal of avoiding to make updates which find new misclassified pixels but leads to correct decisions for already misclassified pixels.

\item[\textbf{CosPGD:}] 
\cite{agnihotri2023cospgd} proposes to weigh the importance of the pixels via cosine similarity between the prediction vector (after applying the sigmoid function $\sigma(t) = 
1/(1 + e^{-t})$) and the one-hot encoding $\ve_y$ of the ground truth class. This can be written as 
\begin{align*}
\L_\textrm{CosPGD}(\vu, y) = \frac{\inner{\sigma(\vu),\ve_y}}{\norm{\sigma(\vu)}_2 \norm{\ve_y}_2}\cdot \L_\textrm{CE}(\vu, y)= \sigma(\vu_y) / \norm{\sigma(\vu)}_2 \cdot \L_\textrm{CE}(\vu, y),
\end{align*}
and again has the effect of reducing the importance of the pixels which are confidently misclassified.
\end{description}

\subsection{{Transfer attacks}}\label{app:transfer_attacks}

\begin{table}[!t]
\centering
\caption{{\textbf{Transfer attacks.} We show the robustness of \RobAT to various transfer attacks (measured with \acc and \hlc[BlueGray]{\miou} at various radii). For each case we indicate the source and target models. 
Moreover, we report the evaluation given by white-box attacks as baseline.}}
\label{tab:transfer_attacks}
\small \tabcolsep=1.3pt
\extrarowheight=1.5pt
\newl=6mm
\begin{tabular}{L{34mm} C{13mm} C{13mm} | 
*{4}{|
C{\newl}
>{\columncolor{BlueGray}}C{\newl}
}}
Attack & Source & Target & \multicolumn{2}{c}{0} & \multicolumn{2}{c}{4/255} & \multicolumn{2}{c}{8/255} & \multicolumn{2}{c}{12/255} \\
\toprule
\addlinespace[1mm]

\multicolumn{11}{l}{\textbf{\ade, Segmenter with ViT-S backbone}} \\
\toprule
APGD w/ $\L_\textrm{Mask-CE}$& clean & \RobAT & 69.1 & \textcolor{NewGray}{28.7} & 68.8 & \textcolor{NewGray}{28.3} & 68.6 & \textcolor{NewGray}{28.0} & 68.3 & \textcolor{NewGray}{27.8}\\
APGD w/ $\L_\textrm{Mask-CE}$& AT & \RobAT & 69.1 & \textcolor{NewGray}{28.7} & 66.3 & \textcolor{NewGray}{26.0} & 63.1 & \textcolor{NewGray}{23.8} & 57.4 & \textcolor{NewGray}{19.9}\\
\midrule

\multiloss (white-box) & 
\RobAT & \RobAT 
& 69.1 & \textcolor{NewGray}{28.7} & 55.3 & \textcolor{NewGray}{14.9} & 33.3 & \textcolor{NewGray}{5.4} & 8.9 & \textcolor{NewGray}{1.1}\\
\bottomrule 
\end{tabular} \end{table}

To complement the evaluation of the robustness of our \RobAT models, we further test them with transfer attacks from less robust models. 
In particular, we run APGD on the Masked-CE loss on Segmenter models obtained with either clean training or AT (5 steps) on \ade.
We then transfer the found perturbations to our PIR-AT (5 steps, 128 epochs), and report robust accuracy and \miou in \cref{tab:transfer_attacks}, together with the results of the white-box SEA on the same model (from \cref{tab:comp_robust_models_new}) as baseline. We observe that the transfer attacks are far from the performance of SEA, which further supports the robustness of the PIR-AT models.

\section{Additional Figures}
\label{app:additional-fig}

\textbf{Untargeted attacks.} Fig.~\ref{fig:examples_voc_clean} shows examples of our untargeted attacks at different radii $\epsilon_\infty$ on the clean model for \voc dataset.
In particular, we use 300 iterations of red-$\epsilon$ APGD on the $\L_\textrm{Mask-CE}$ loss.
The first column presents the original image with the ground truth segmentation mask, 
The following columns contain the perturbed images and relative predicted segmentation masks for increasing radii ($\epsilon_\infty=0$ is equivalent to the unperturbed image): one can observe that the model predictions 
progressively become farther away from the ground truth values.
We additionally report the average pixel accuracy for each image.
In Fig.~\ref{fig:examples_voc_robust}, we repeat the same visualization for the most robust 5 step 300 epochs \RobAT model.
Note that we use different values of $\epsilon_\infty$ for the two models, i.e. significantly smaller ones for the clean model, following \cref{tab:comp-losses}.
Finally, the same setup is employed on the \upernet + \convnext-T model trained for \ade dataset for the illustrations in Fig.~\ref{fig:examples_ade_clean}  (clean model) and Fig.~\ref{fig:examples_ade_robust} (5-step robust \RobAT model),
and we have similar observations as for the smaller dataset.
Again we use smaller radii for the clean model, since it is significantly less robust than the \RobAT one.

\textbf{Targeted attacks.} In Fig.~\ref{fig:teaser} we show examples of the perturbed images and corresponding predictions resulting from targeted attacks. In this case, we run APGD (red-$\epsilon$ scheme with 300 iterations) on the negative JS divergence between the model predictions and the one-hot encoding of the target class.
In this way the algorithm optimizes the adversarial perturbation to have all pixels classified in the target class (e.g. ``grass'' or ``sky'' in Fig.~\ref{fig:teaser}).
We note that other losses like cross-entropy can be adapted to obtain a targeted version of \multiloss, and we leave the exploration of this aspect of our attacks to future work.

\newh=.16\columnwidth

\begin{figure}[t] \centering
\small
\tabcolsep=1.5pt
\newl=.16\columnwidth
\begin{tabular}{c | c c c c c} 
original & 0 & 0.25/255 & 0.5/255 & 1/255 & 2/255\\
\toprule 
& \acc: 95.9\%& \acc: 94.8\%& \acc: 75.9\%& \acc: 48.3\%& \acc: 0.0\%\\
\includegraphics[width=\newl, height=\newh
]{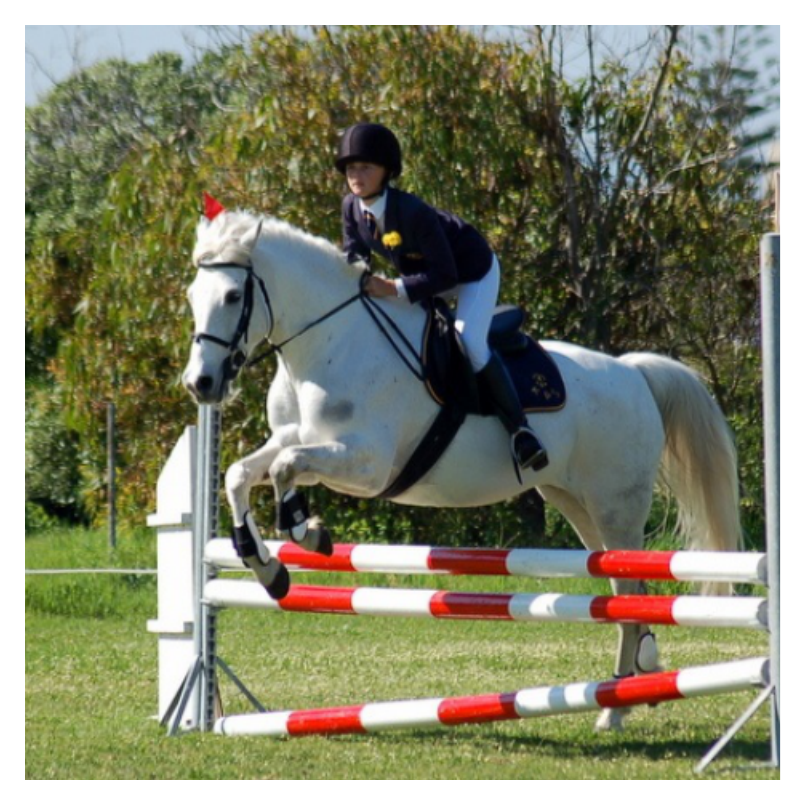} & \includegraphics[width=\newl, height=\newh
]{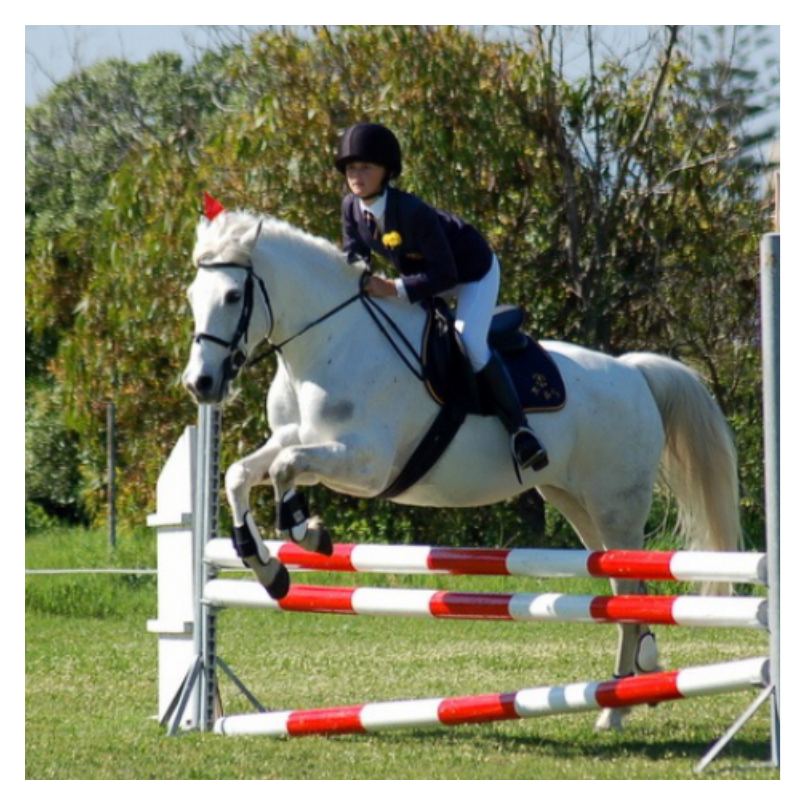} & \includegraphics[width=\newl, height=\newh
]{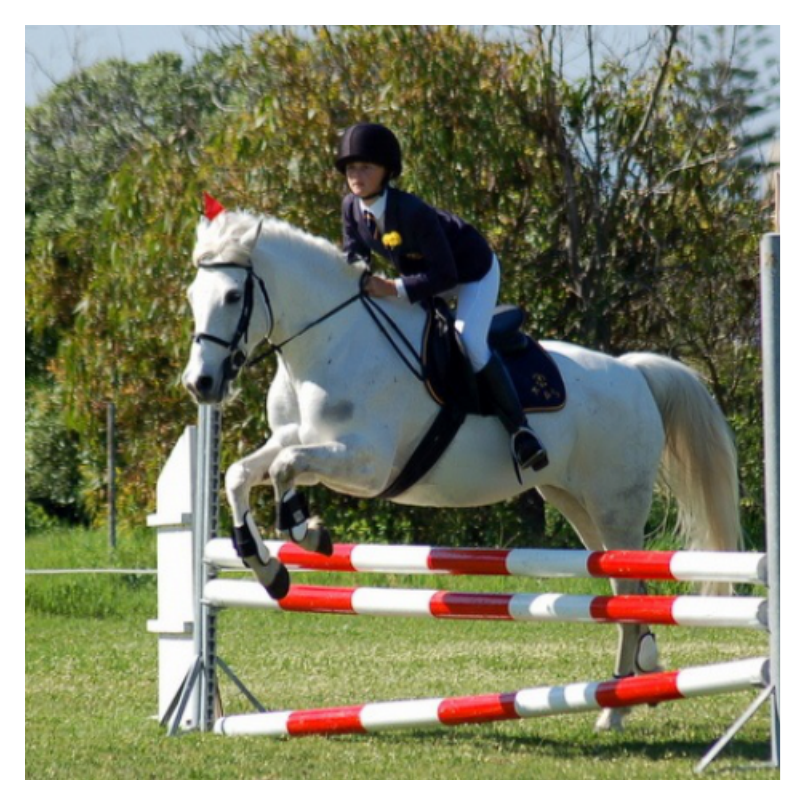} & \includegraphics[width=\newl, height=\newh
]{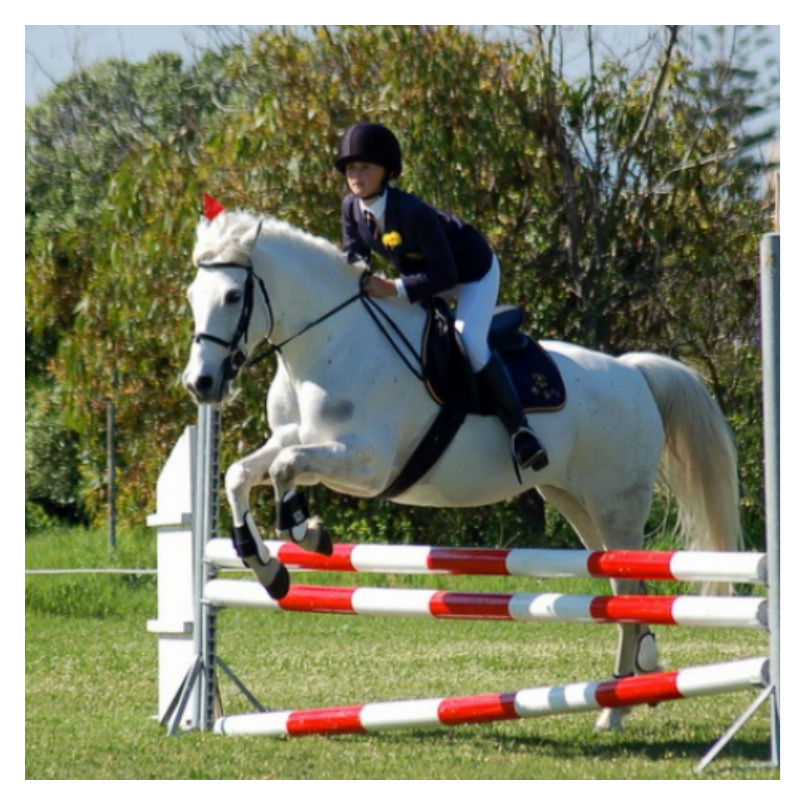} & \includegraphics[width=\newl, height=\newh
]{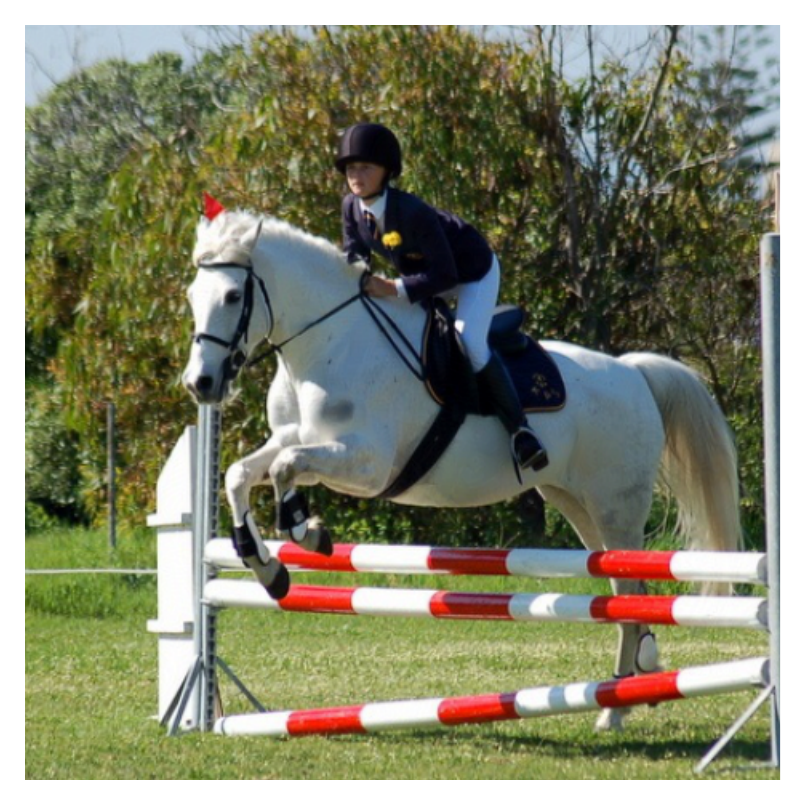} & \includegraphics[width=\newl, height=\newh
]{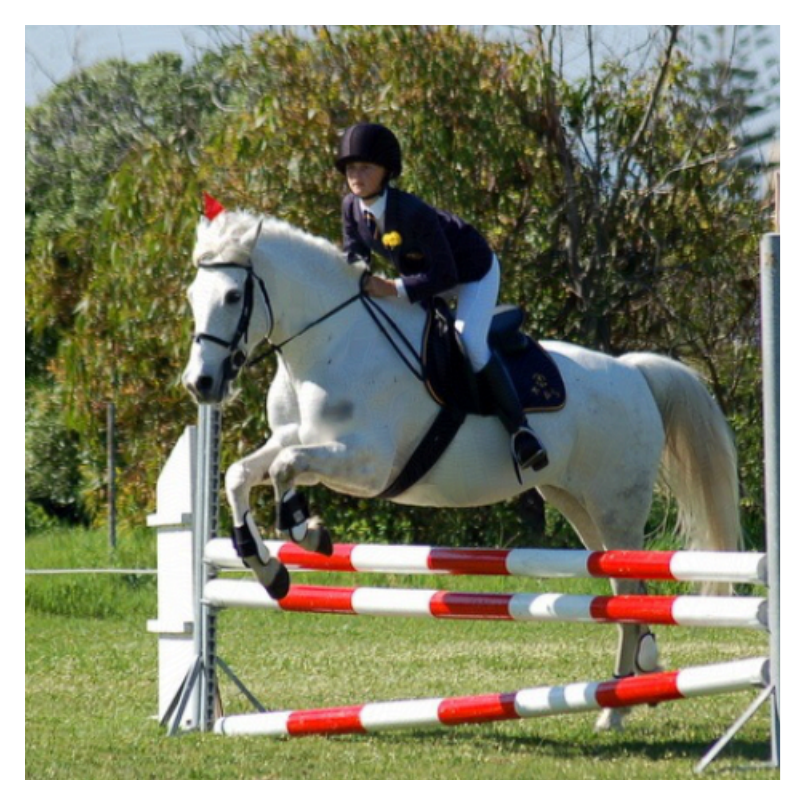} \\ \includegraphics[width=\newl, height=\newh
]{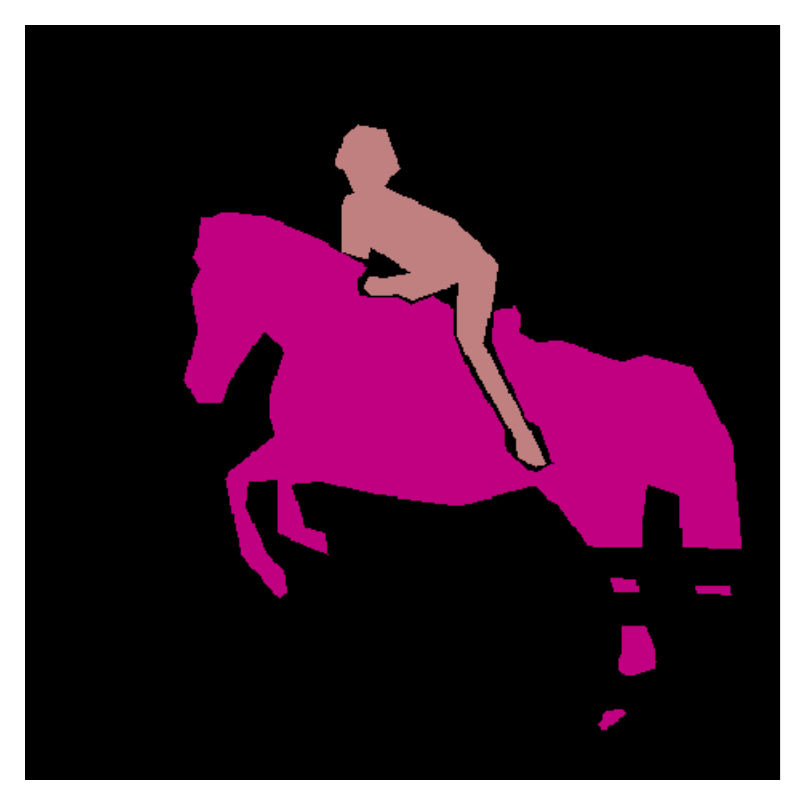} & \includegraphics[width=\newl, height=\newh
]{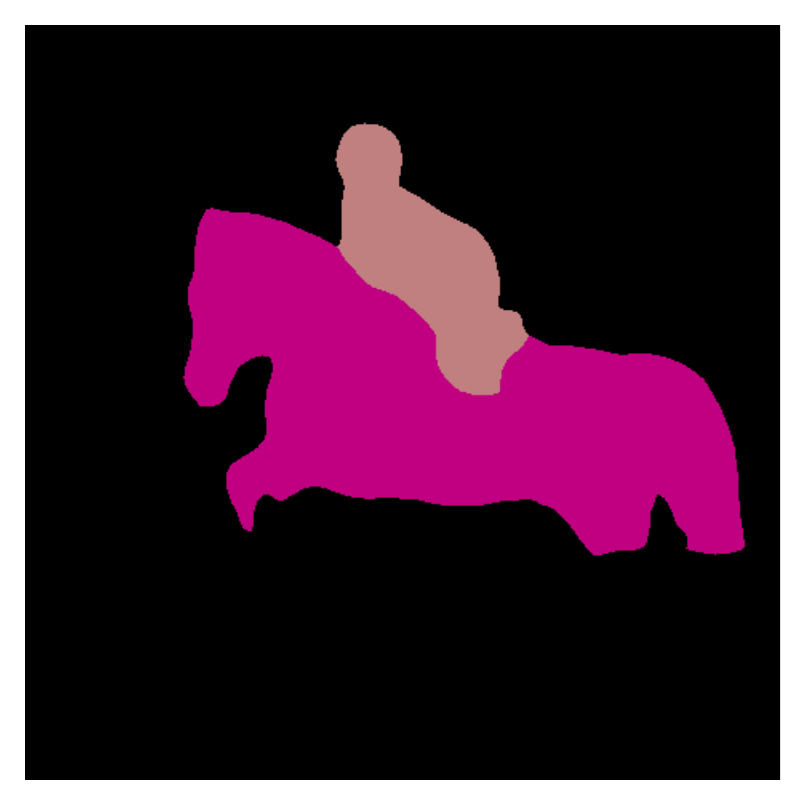} & \includegraphics[width=\newl, height=\newh
]{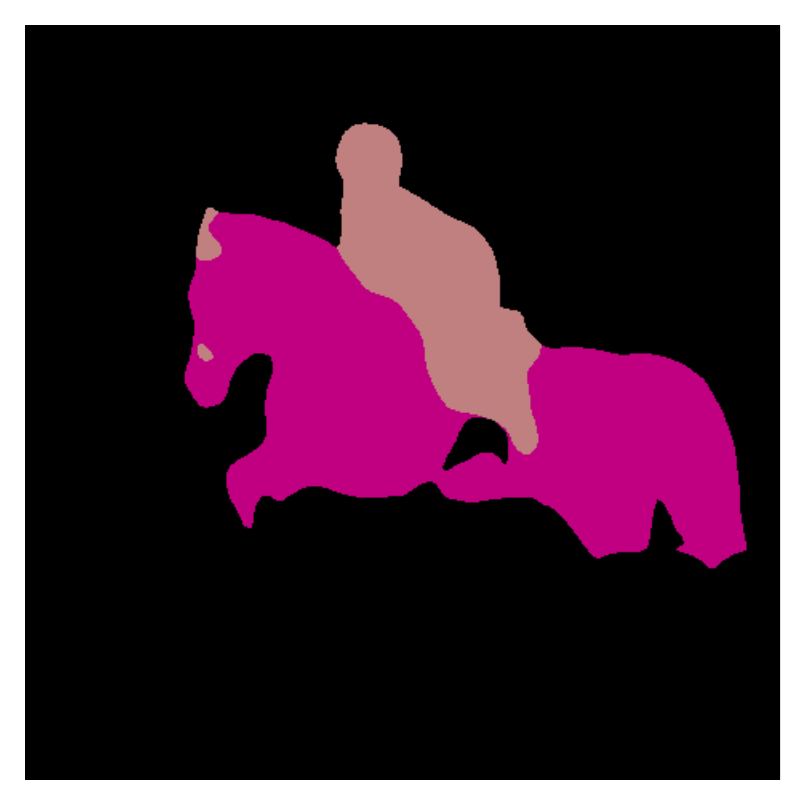} & \includegraphics[width=\newl, height=\newh
]{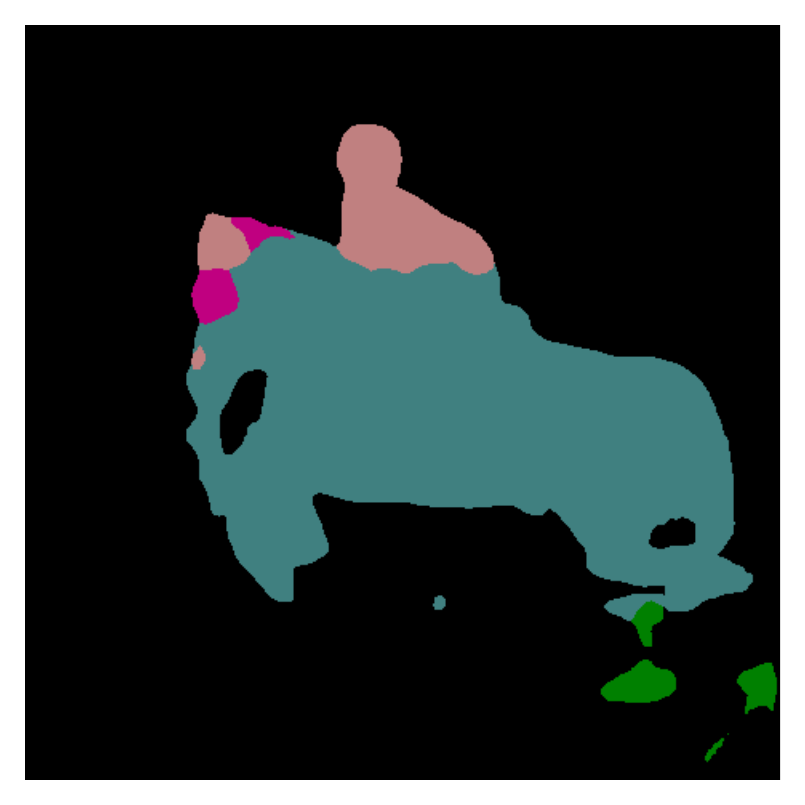} & \includegraphics[width=\newl, height=\newh
]{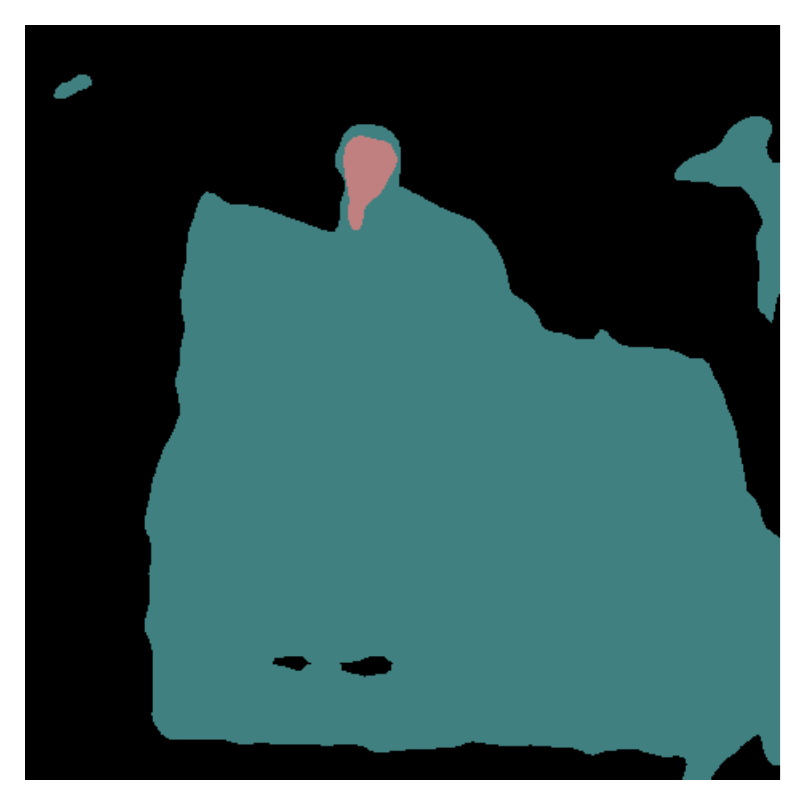} & \includegraphics[width=\newl, height=\newh
]{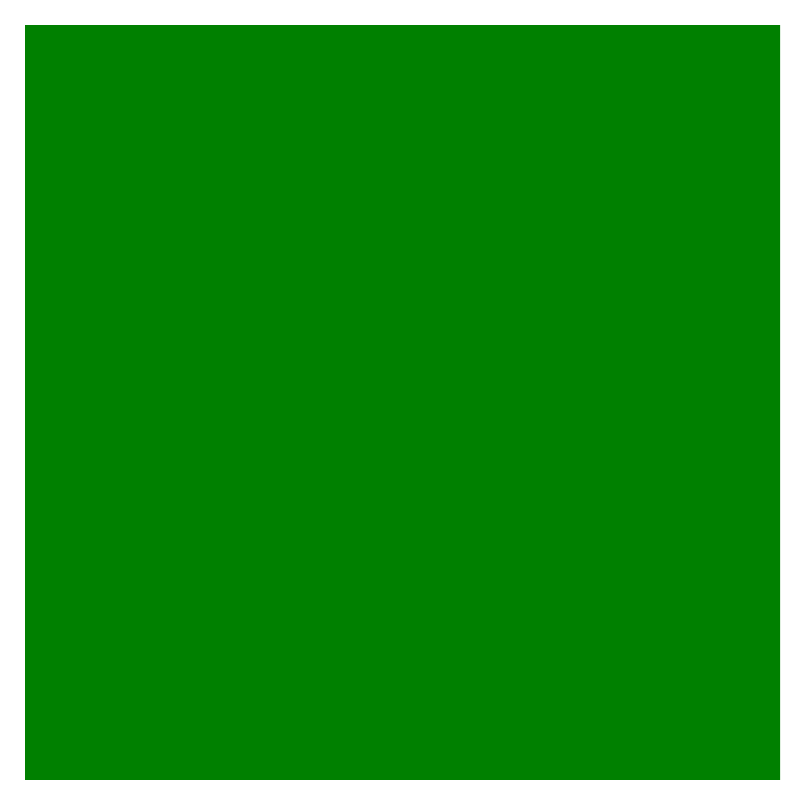}\\
\midrule

& \acc: 96.1\%& \acc: 61.4\%& \acc: 0.0\%& \acc: 0.0\%& \acc: 0.0\%\\
\includegraphics[width=\newl, height=\newh
]{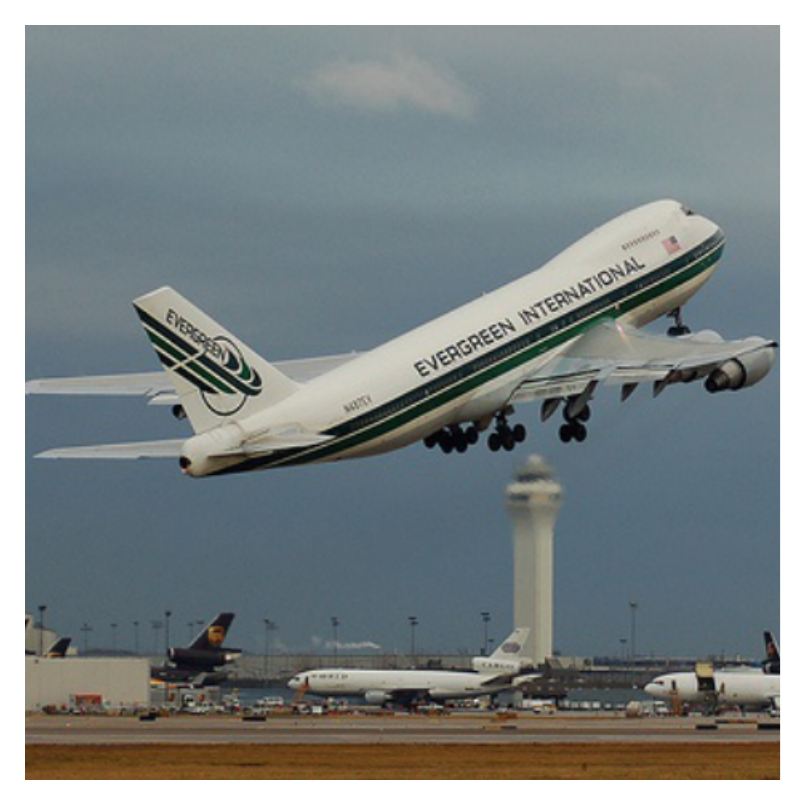} & \includegraphics[width=\newl, height=\newh
]{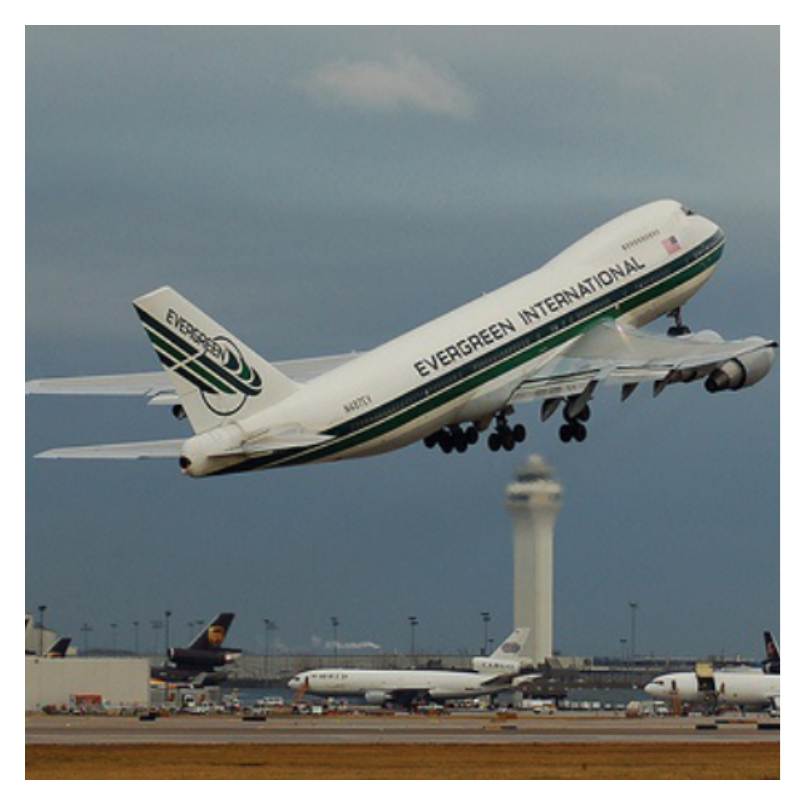} & \includegraphics[width=\newl, height=\newh
]{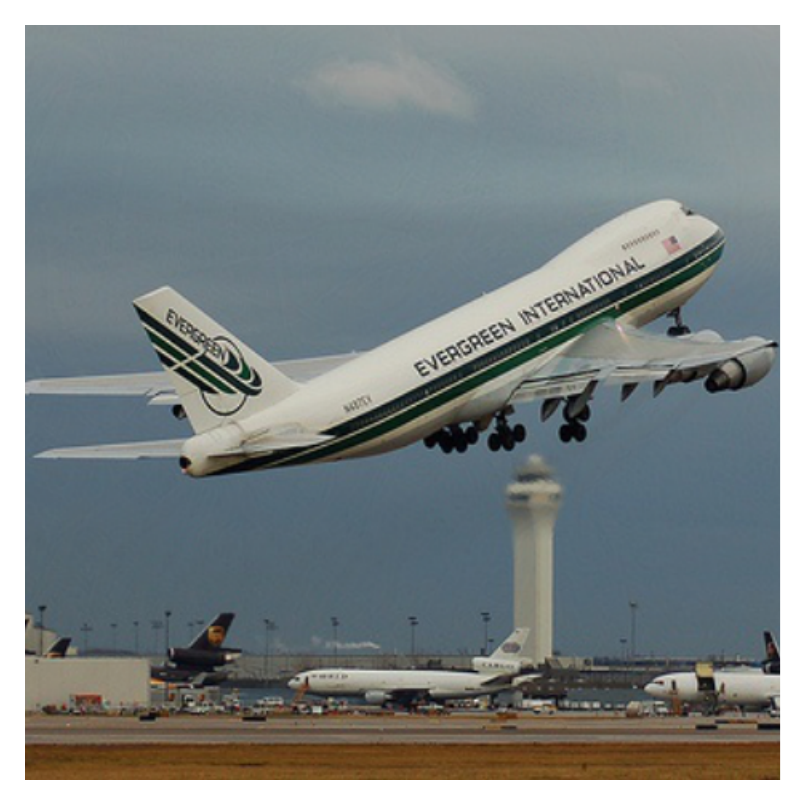} & \includegraphics[width=\newl, height=\newh
]{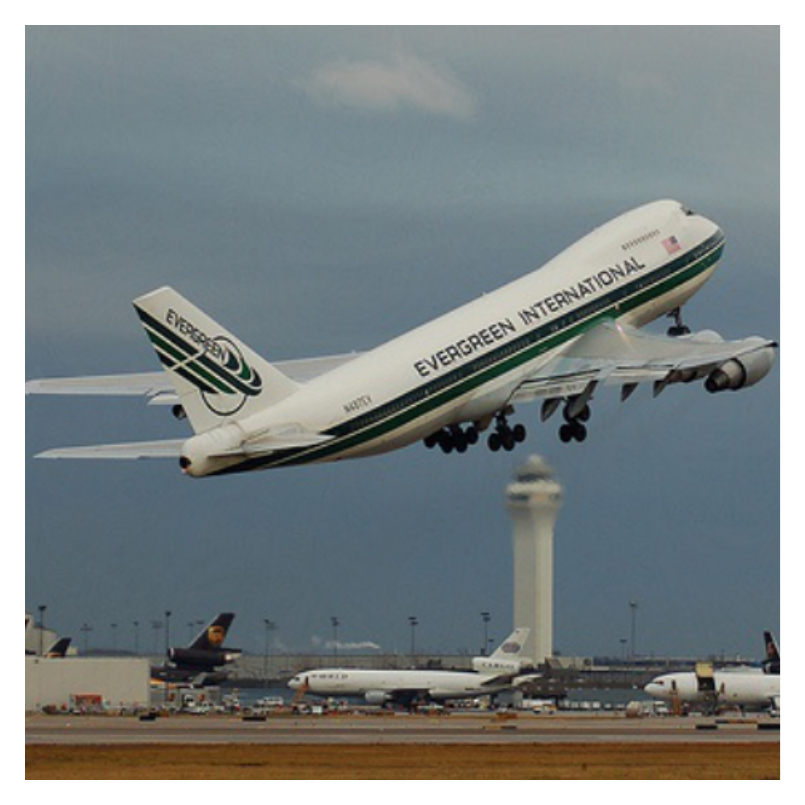} & \includegraphics[width=\newl, height=\newh
]{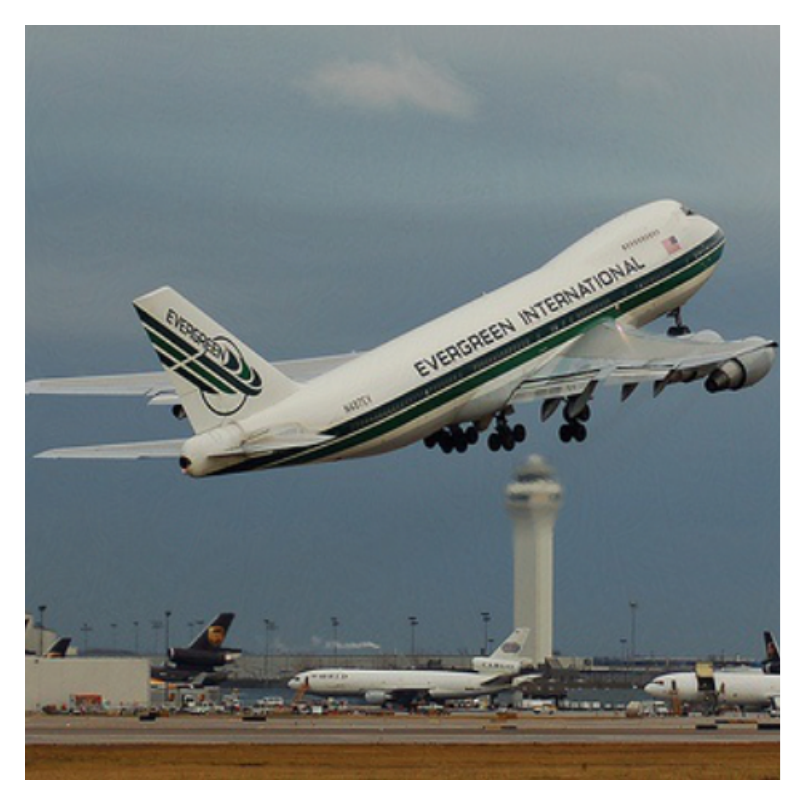} & \includegraphics[width=\newl, height=\newh
]{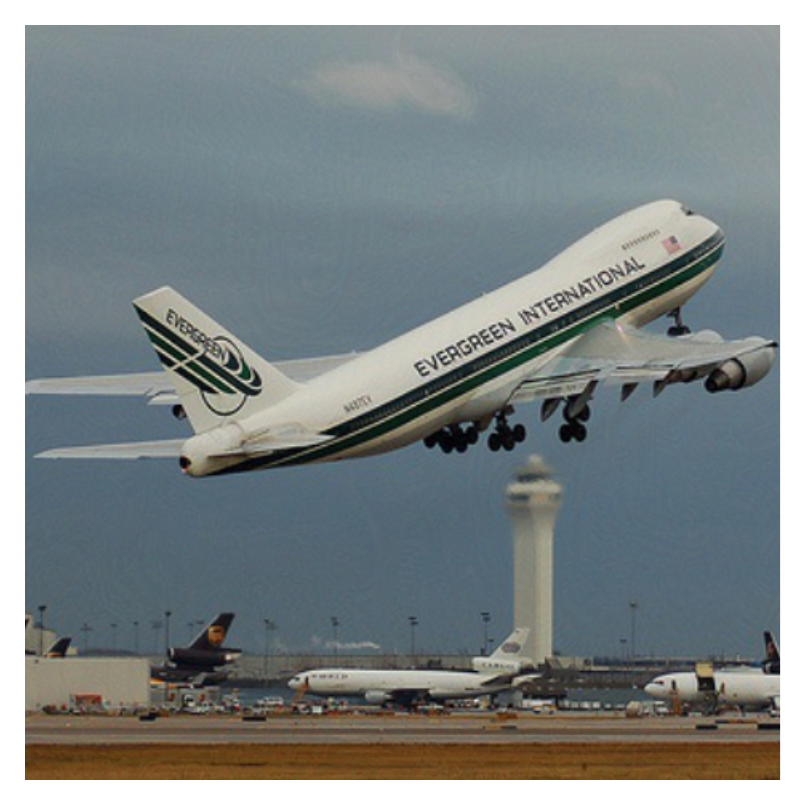} \\ \includegraphics[width=\newl, height=\newh
]{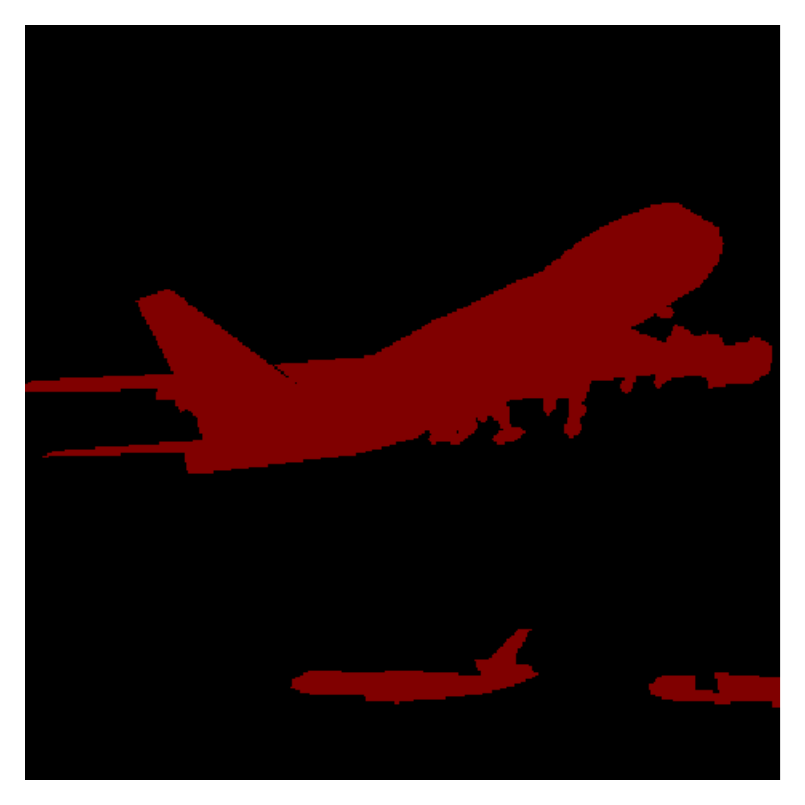} & \includegraphics[width=\newl, height=\newh
]{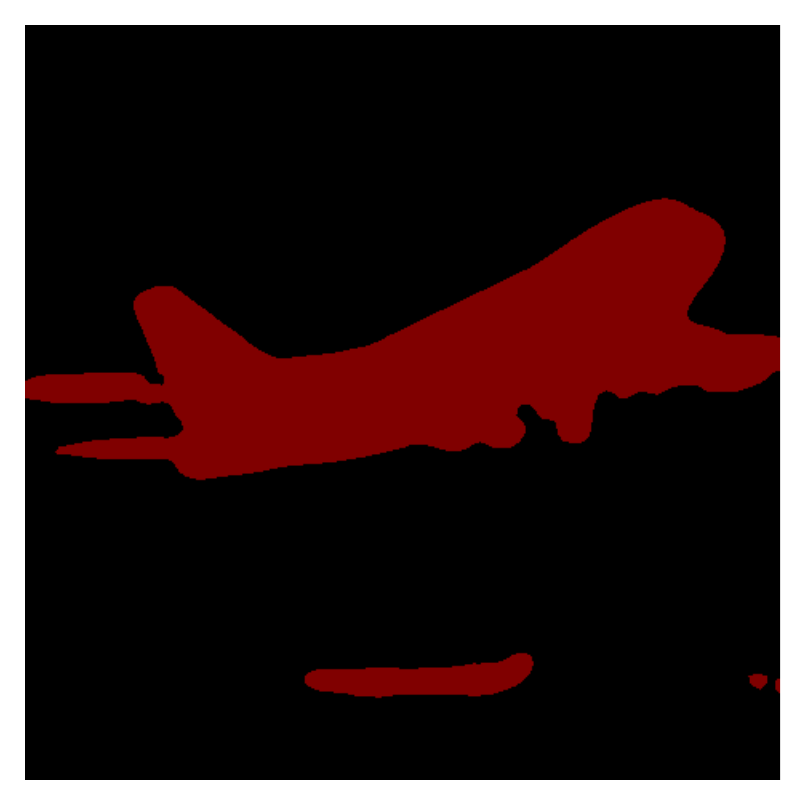} & \includegraphics[width=\newl, height=\newh
]{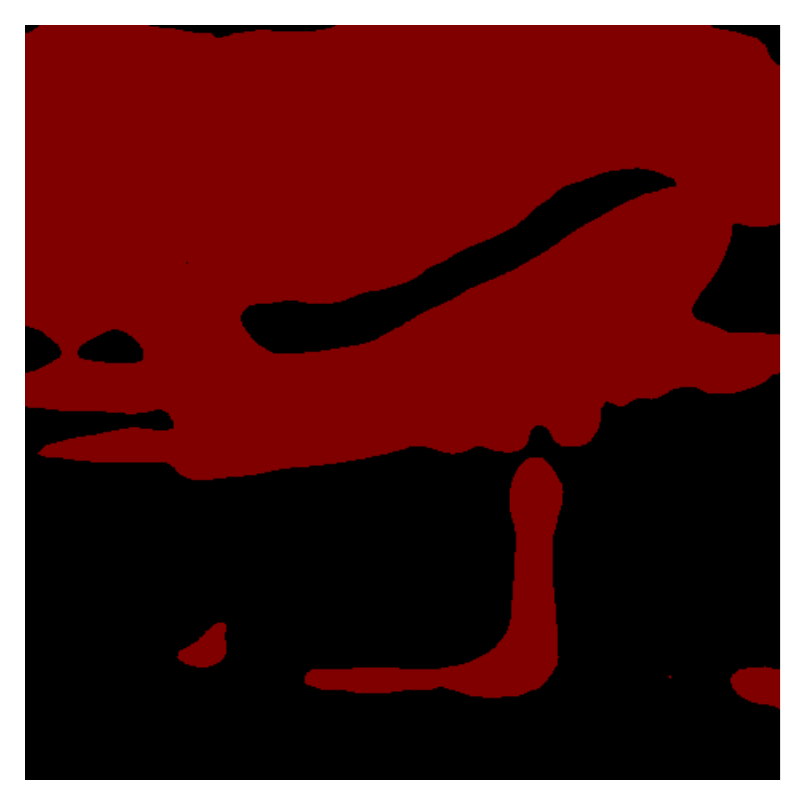} & \includegraphics[width=\newl, height=\newh
]{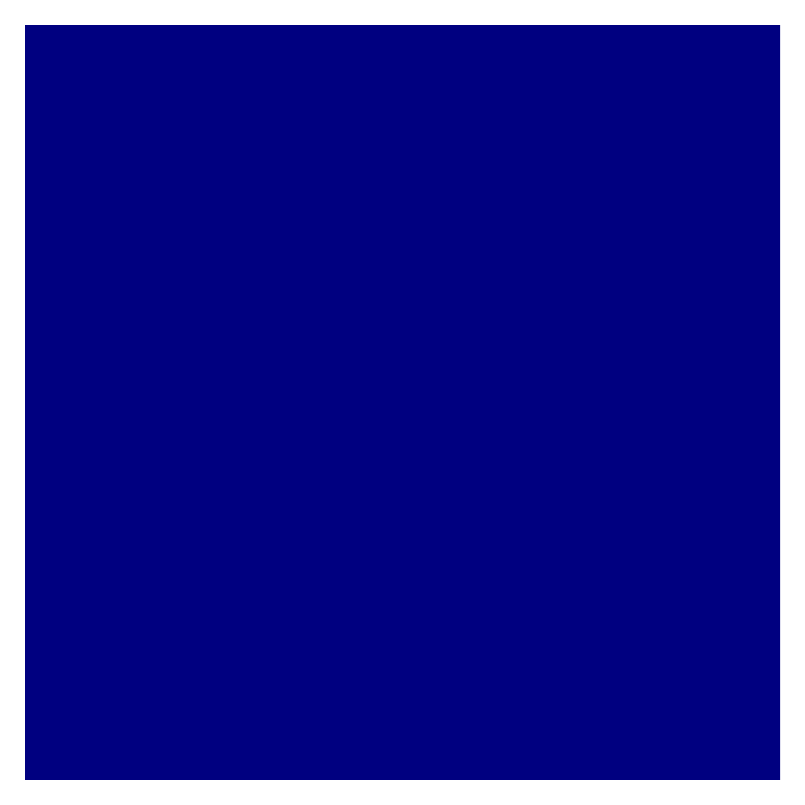} & \includegraphics[width=\newl, height=\newh
]{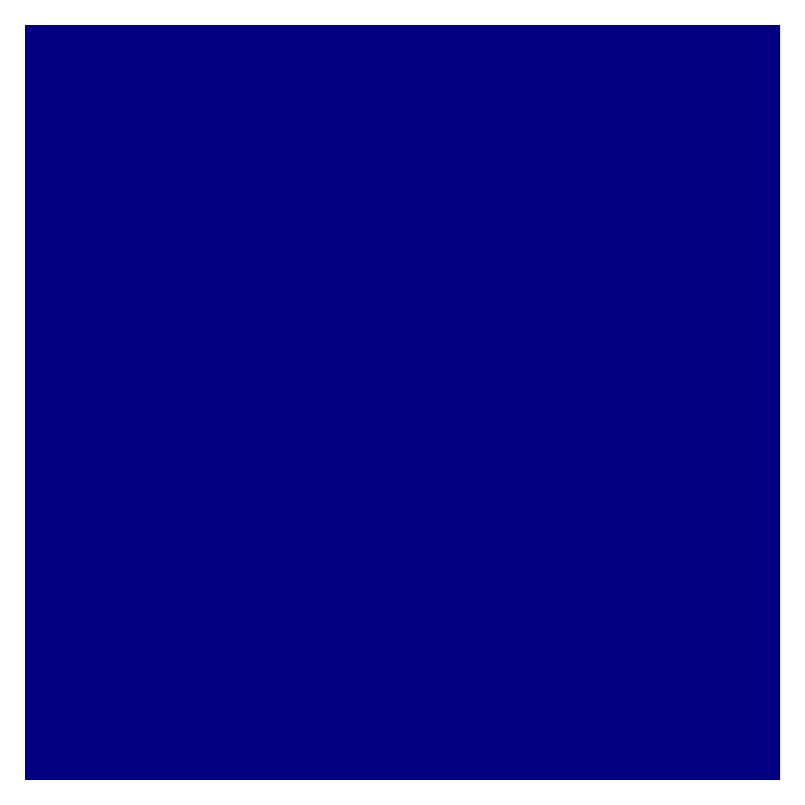} & \includegraphics[width=\newl, height=\newh
]{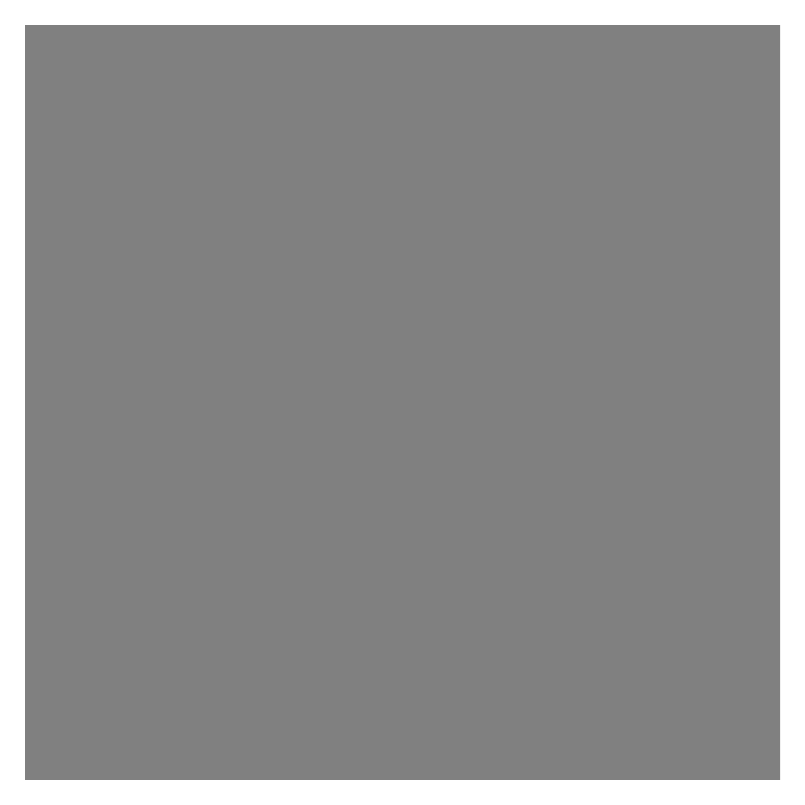}
\end{tabular}
\caption{
Visualizing the perturbed images, corresponding predicted masks and \acc for increasing radii. The attacks are generated on the clean model on \voc with APGD on $\L_\textrm{Mask-CE}$. Original image and ground truth mask in the first column.} \label{fig:examples_voc_clean}
\end{figure}

\begin{figure}[b] \centering
\small
\tabcolsep=1.5pt
\newl=.16\columnwidth
\begin{tabular}{c | c c c c c} 
original & 0 & 4/255 & 8/255 & 12/255 & 16/255\\
\toprule 
& \acc: 95.5\%& \acc: 94.6\%& \acc: 90.8\%& \acc: 49.2\%& \acc: 0.0\%\\
\includegraphics[width=\newl, height=\newh]{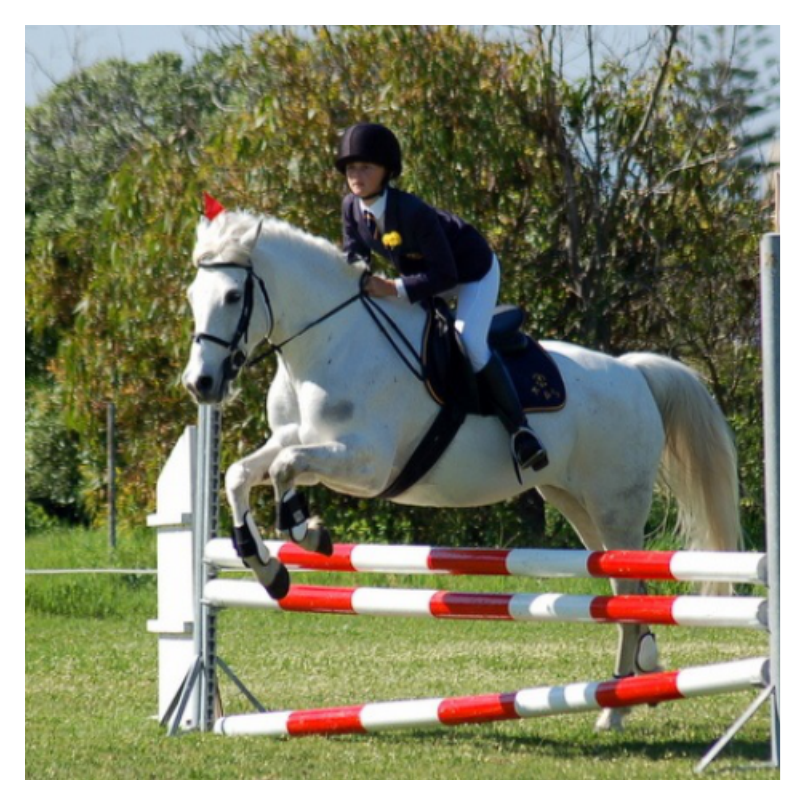} & \includegraphics[width=\newl, height=\newh]{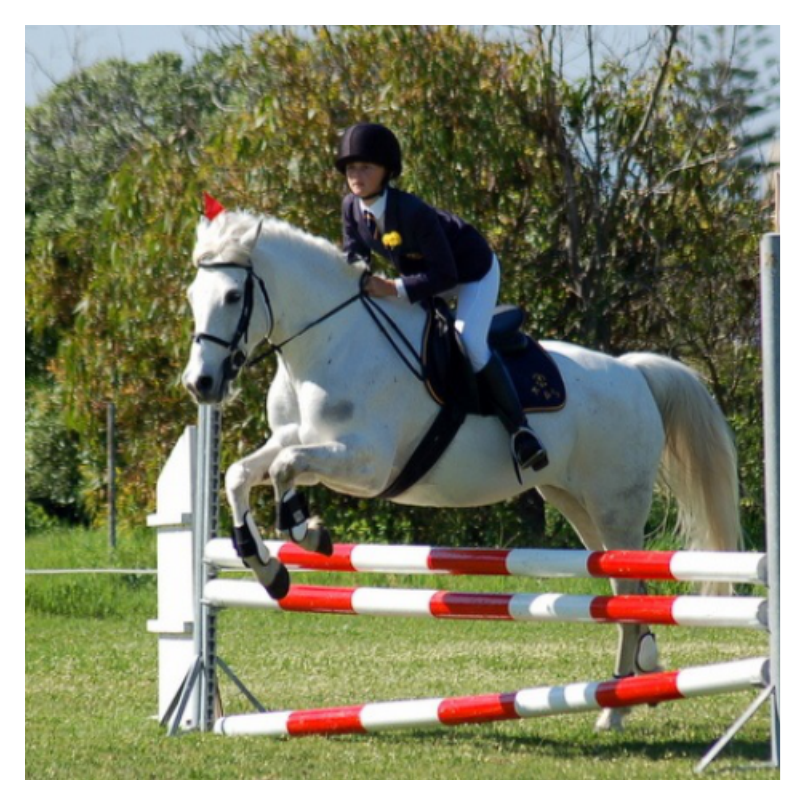} & \includegraphics[width=\newl, height=\newh]{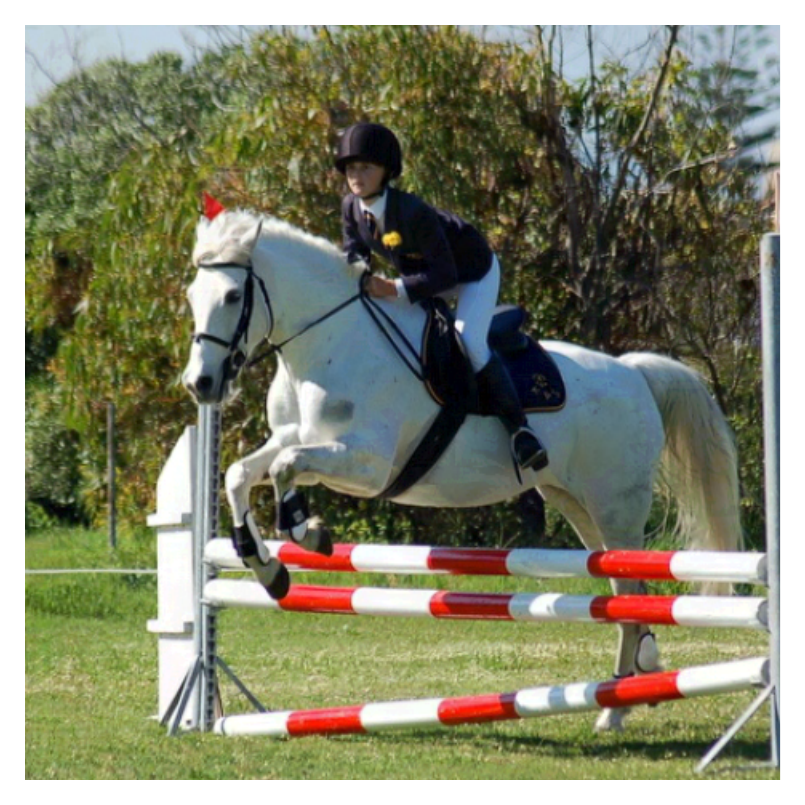} & \includegraphics[width=\newl, height=\newh]{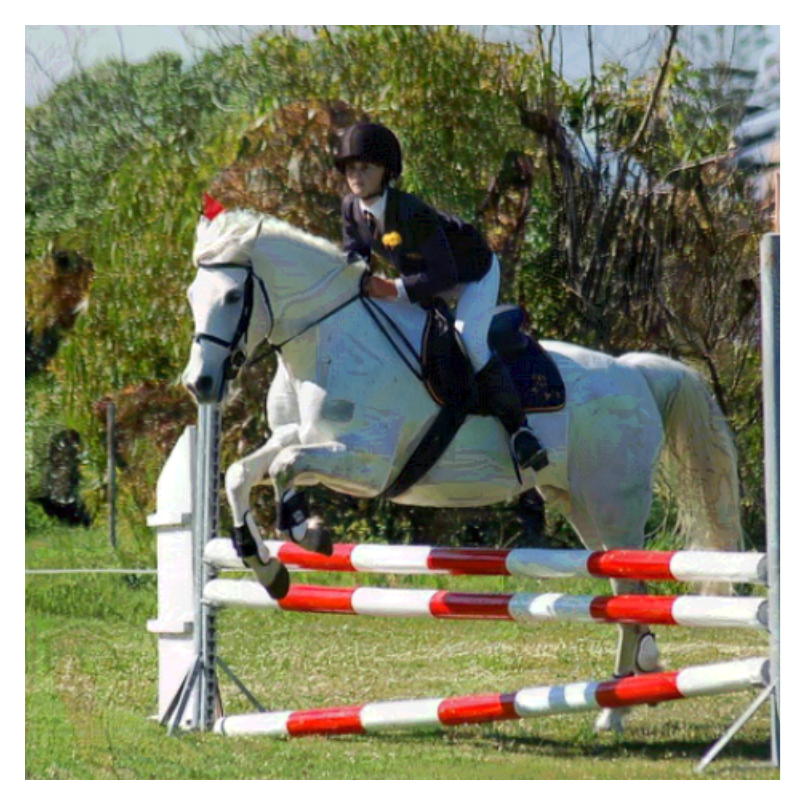} & \includegraphics[width=\newl, height=\newh]{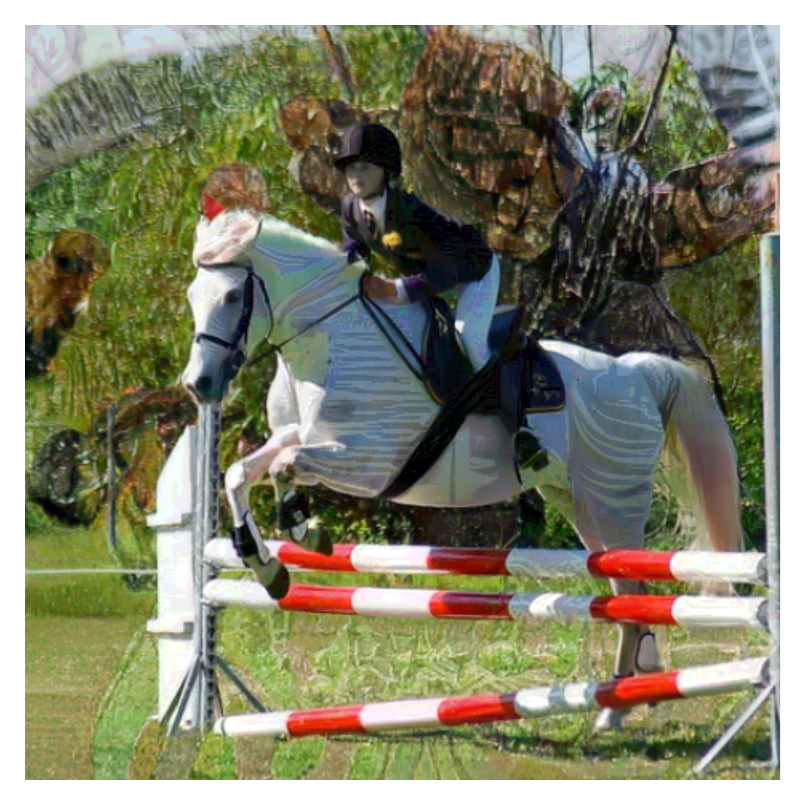} & \includegraphics[width=\newl, height=\newh]{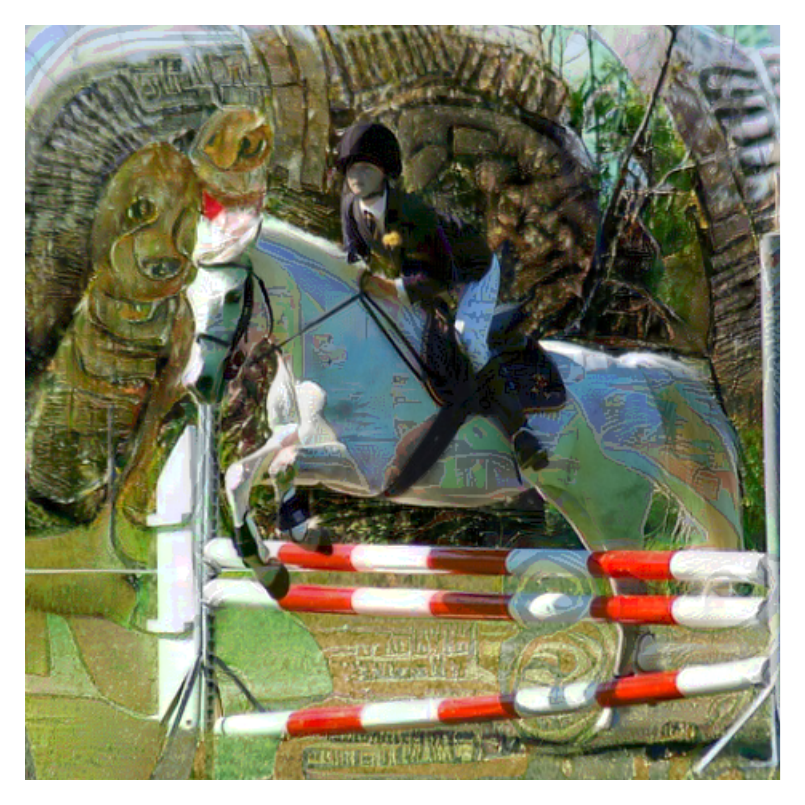} \\ \includegraphics[width=\newl, height=\newh]{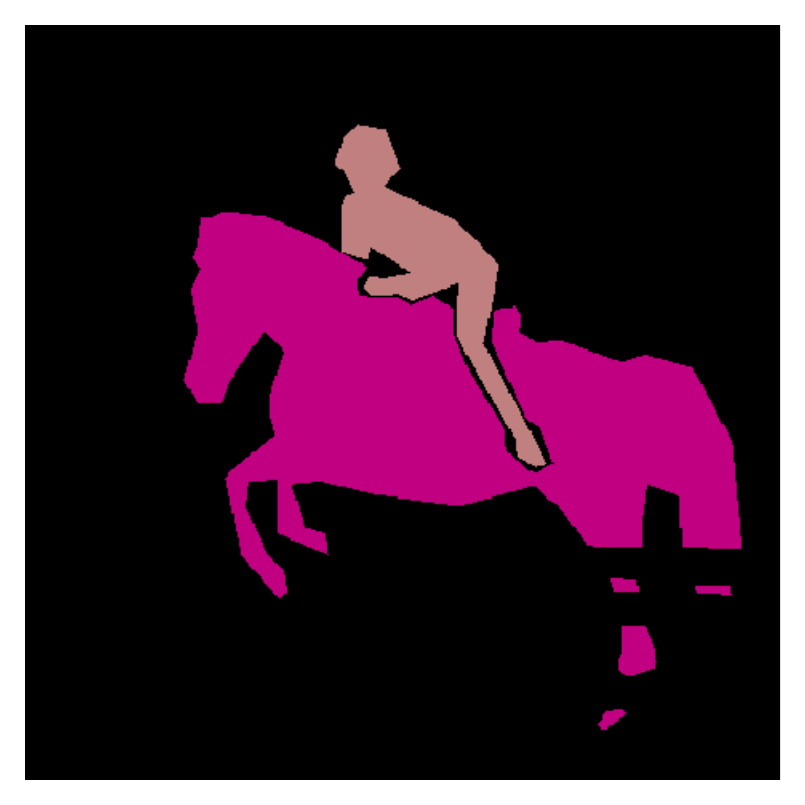} & \includegraphics[width=\newl, height=\newh]{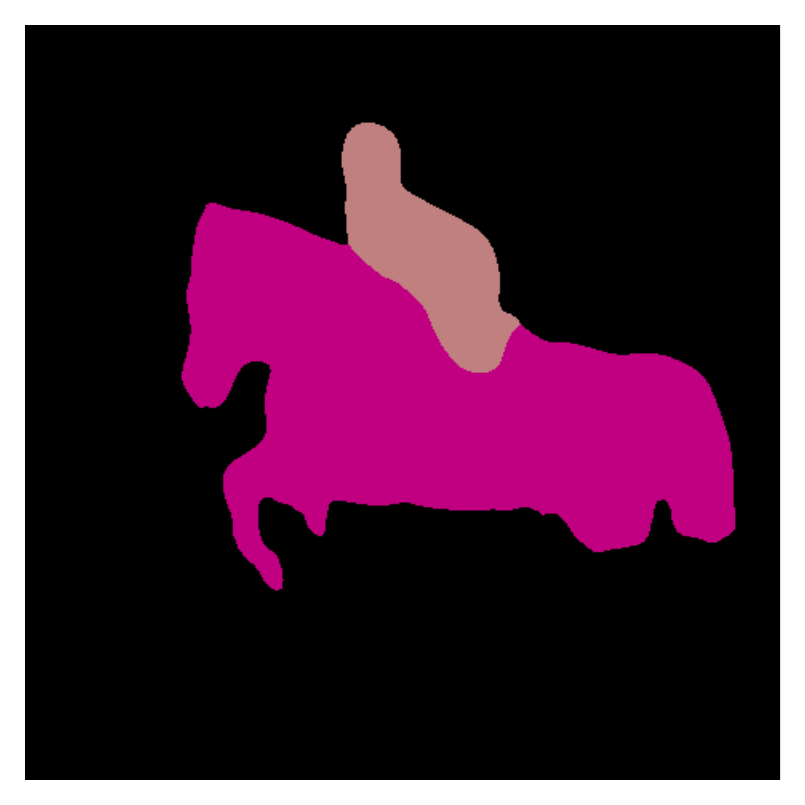} & \includegraphics[width=\newl, height=\newh]{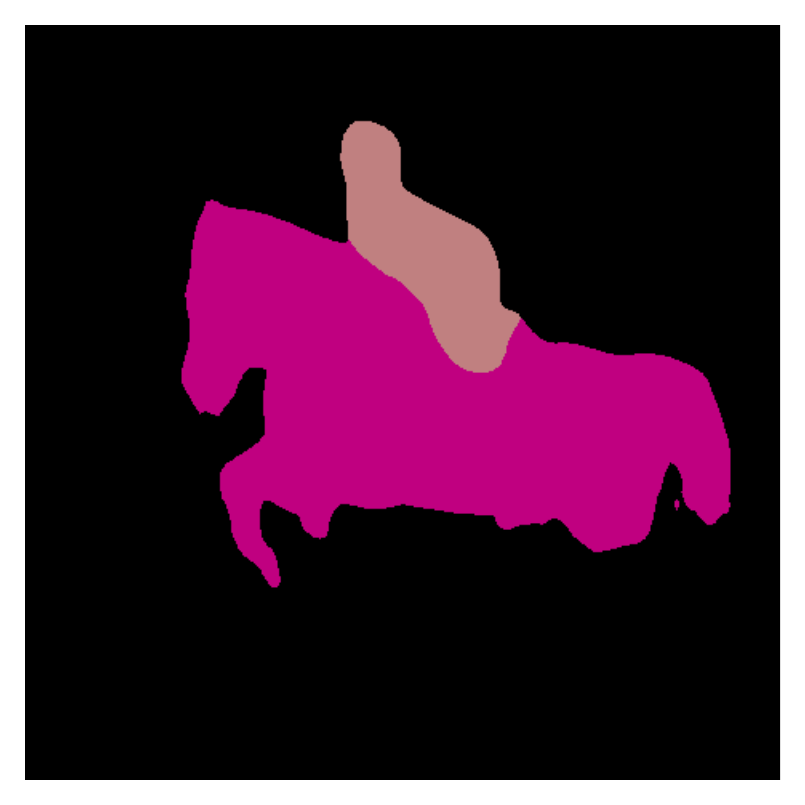} & \includegraphics[width=\newl, height=\newh]{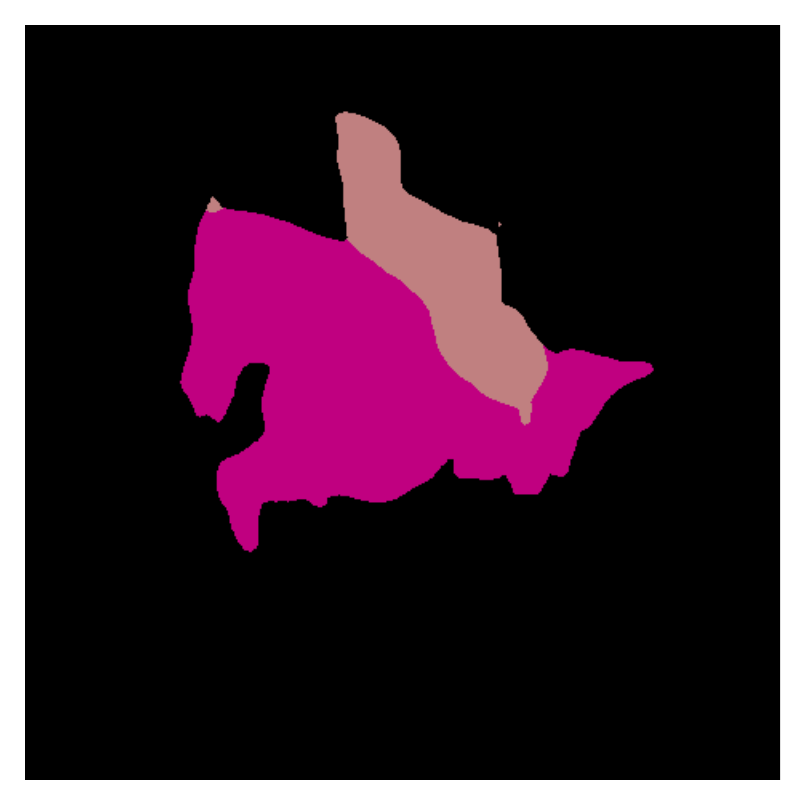} & \includegraphics[width=\newl, height=\newh]{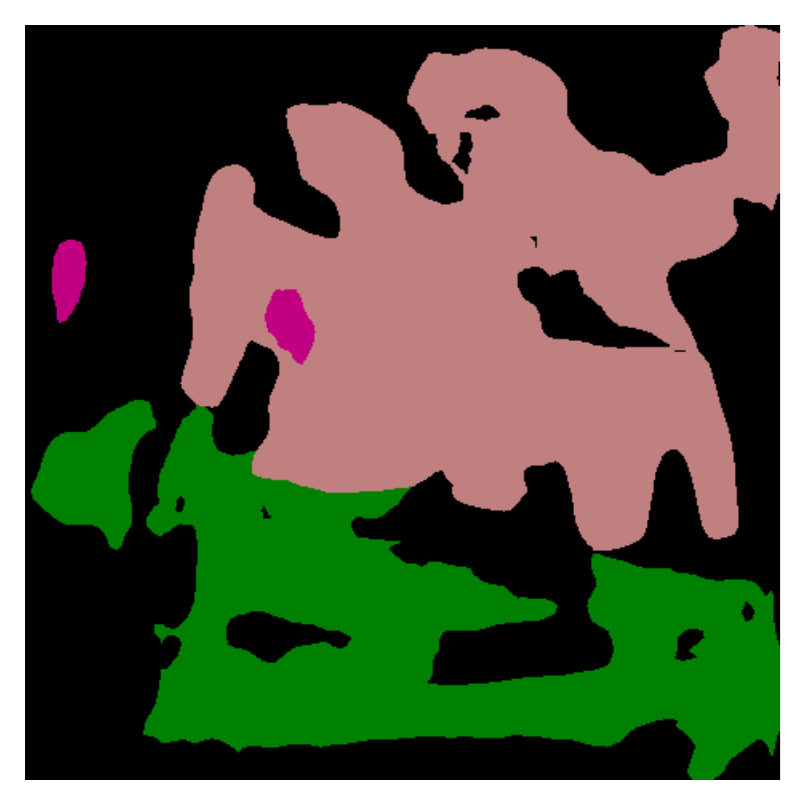} & \includegraphics[width=\newl, height=\newh]{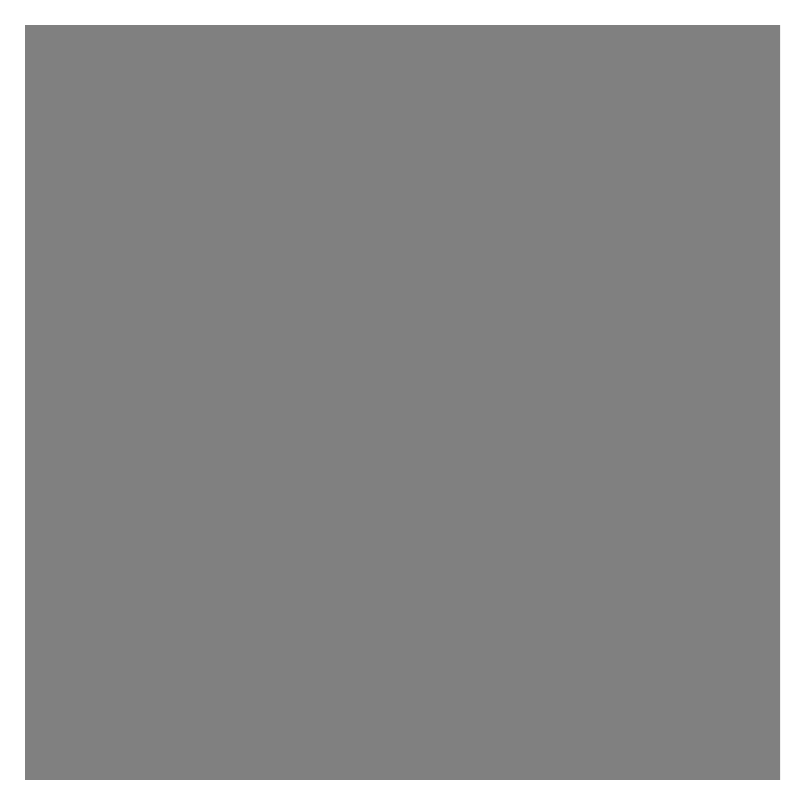}\\
\midrule
& \acc: 93.7\%& \acc: 92.7\%& \acc: 83.3\%& \acc: 6.8\%& \acc: 0.0\%\\
\includegraphics[width=\newl, height=\newh]{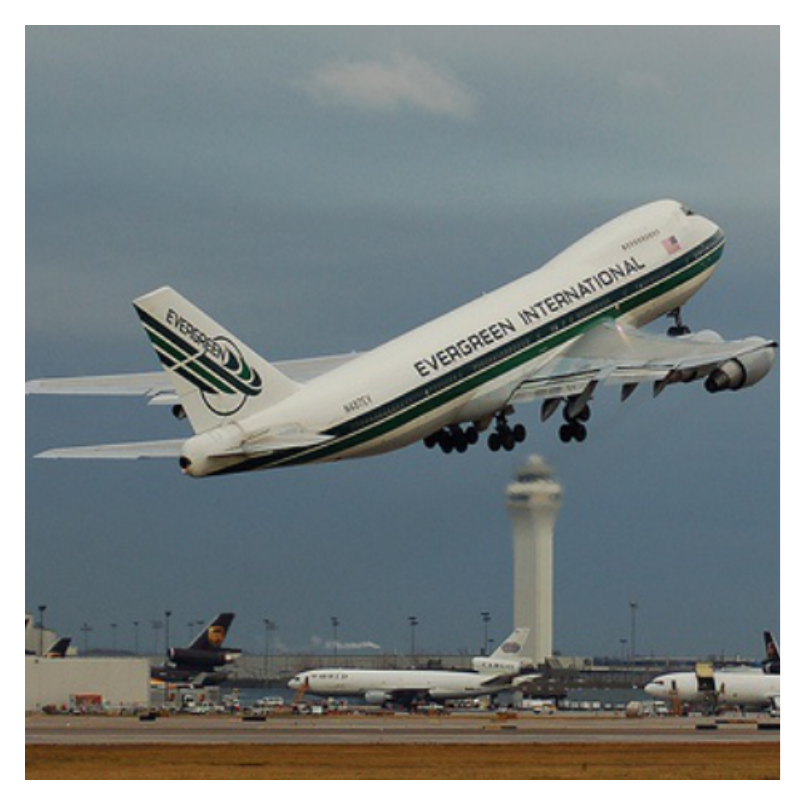} & \includegraphics[width=\newl, height=\newh]{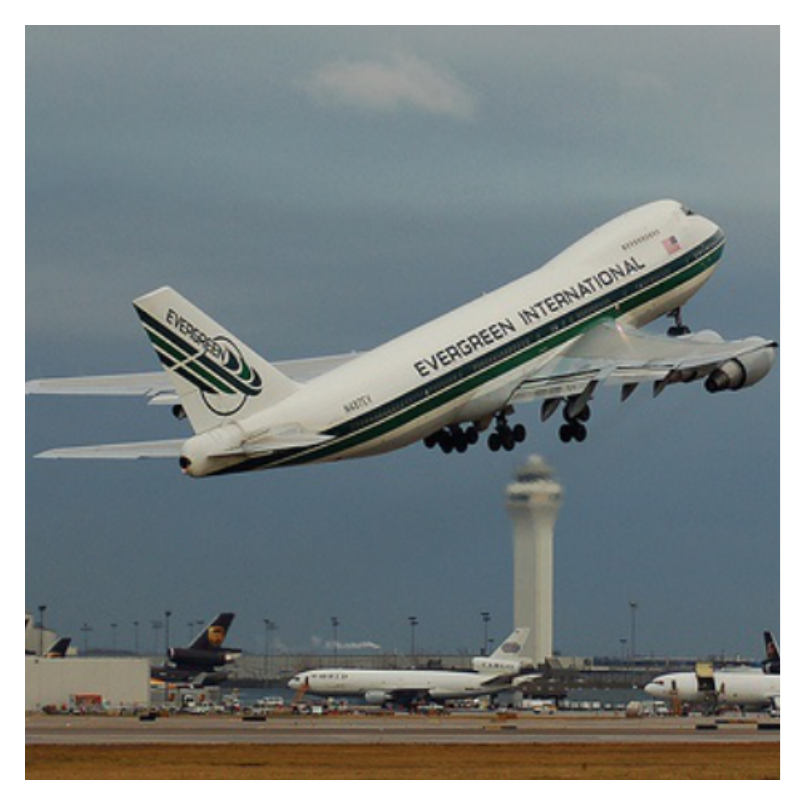} & \includegraphics[width=\newl, height=\newh]{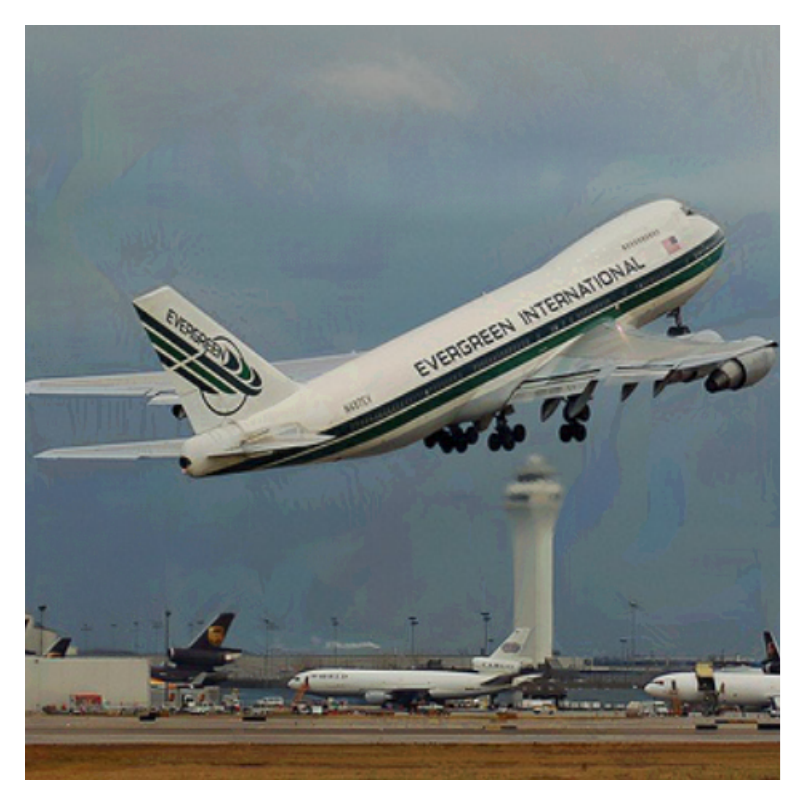} & \includegraphics[width=\newl, height=\newh]{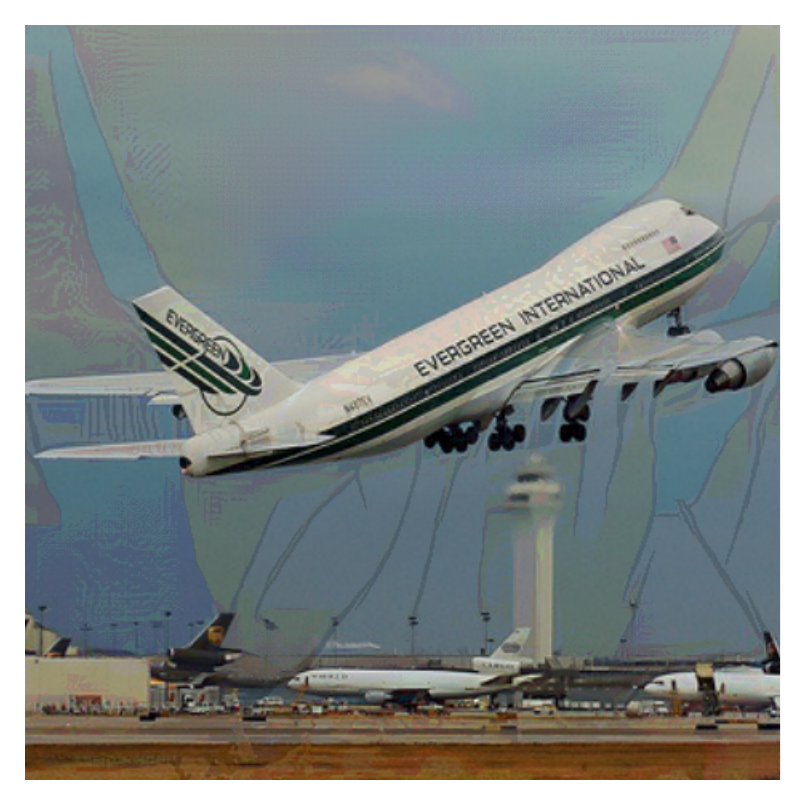} & \includegraphics[width=\newl, height=\newh]{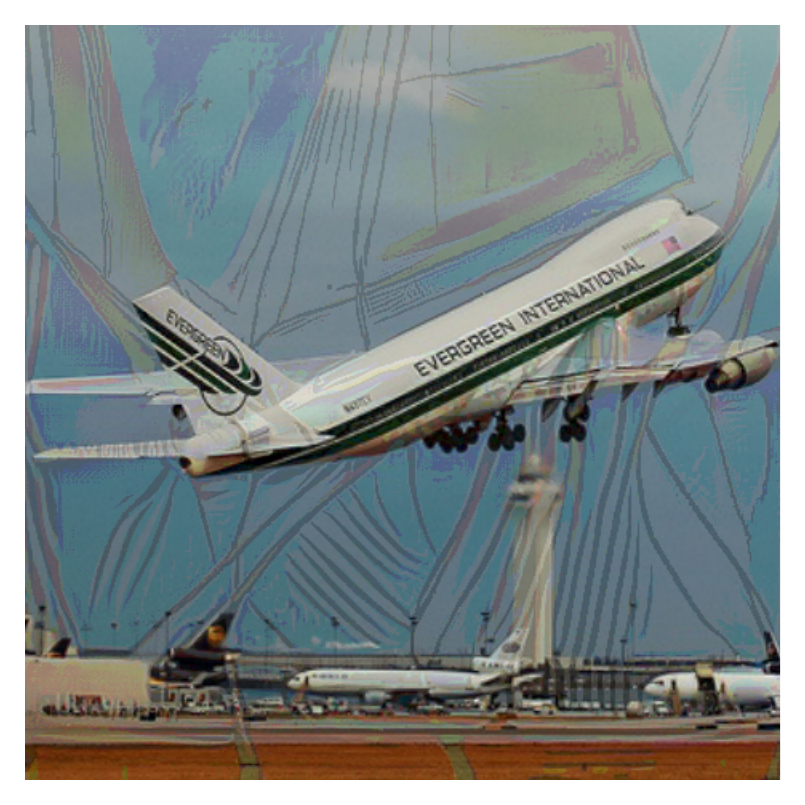} & \includegraphics[width=\newl, height=\newh]{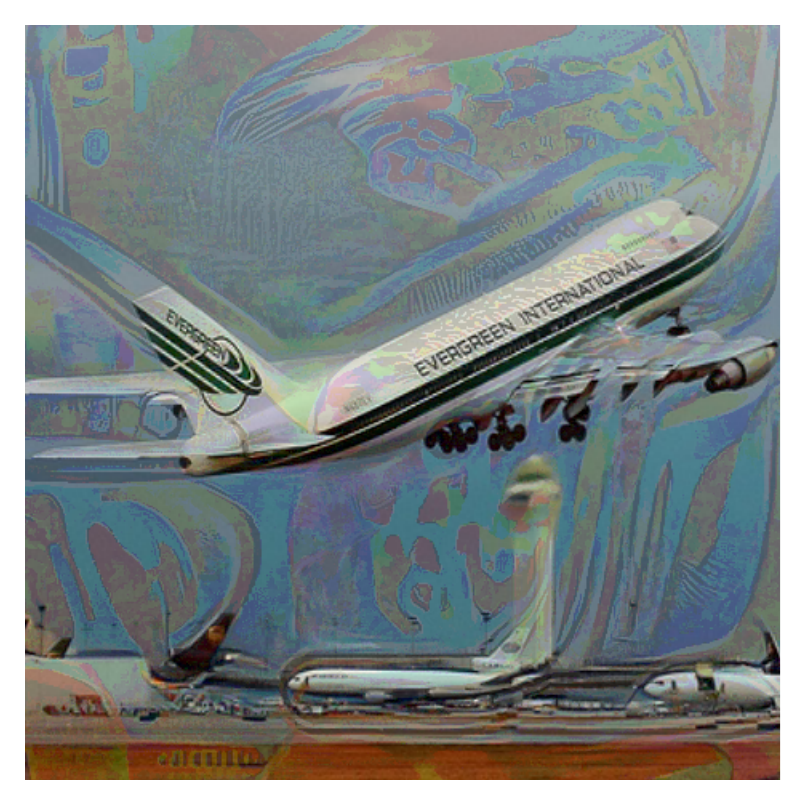} \\ \includegraphics[width=\newl, height=\newh]{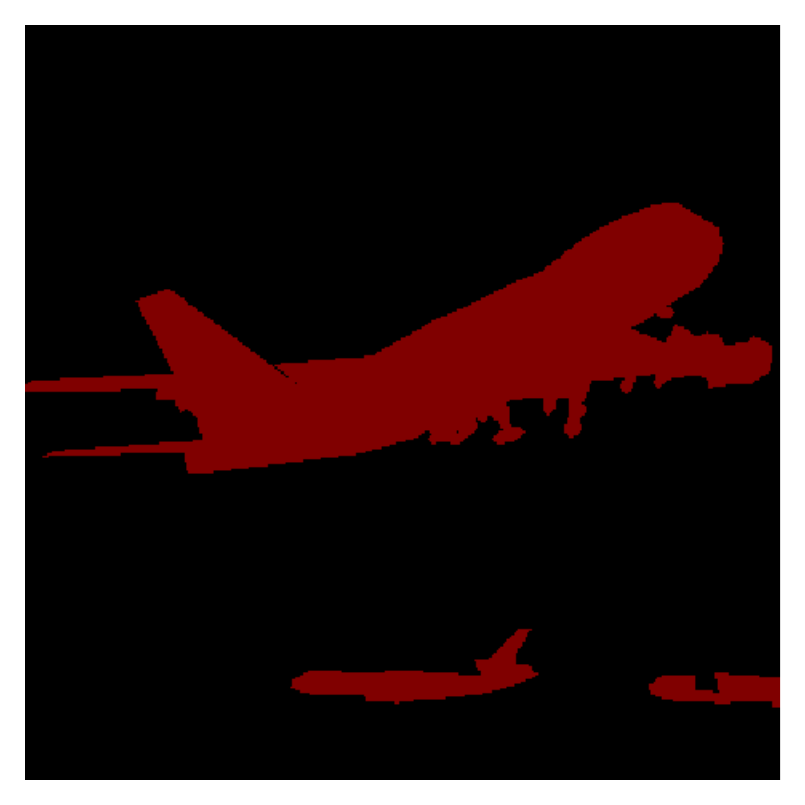} & \includegraphics[width=\newl, height=\newh]{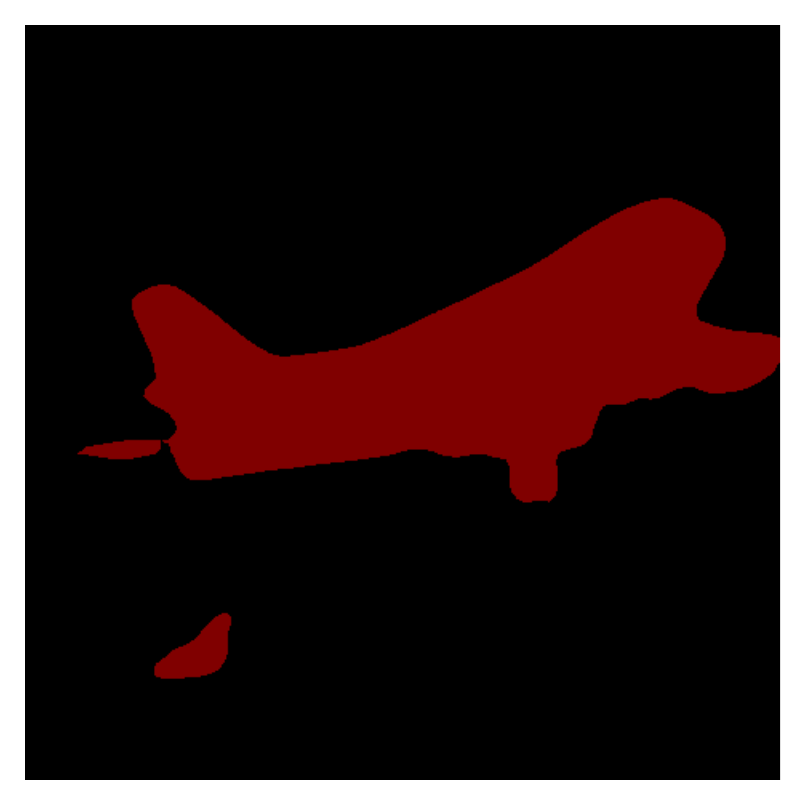} & \includegraphics[width=\newl, height=\newh]{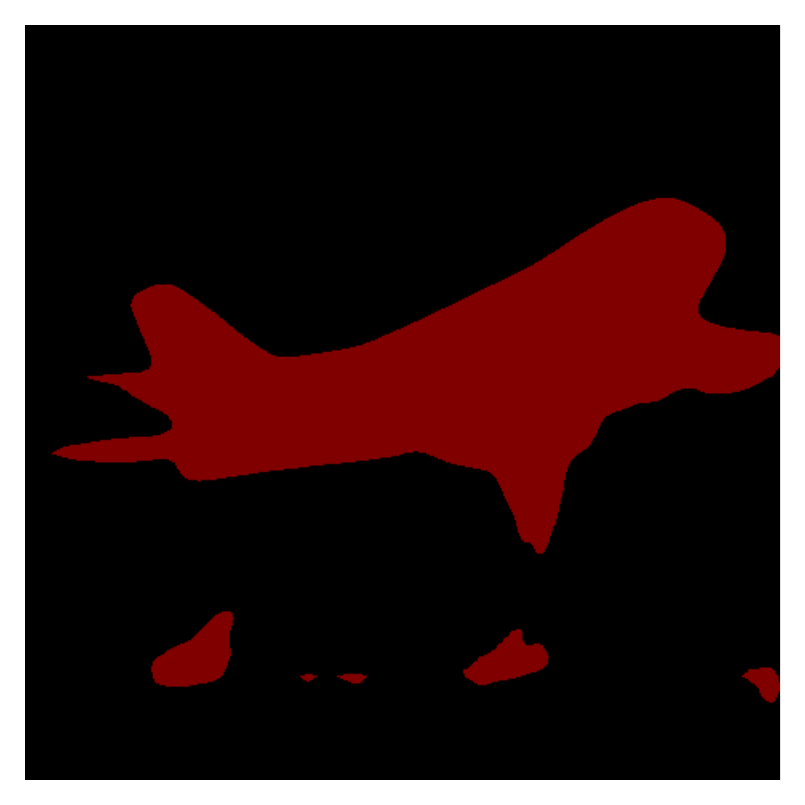} & \includegraphics[width=\newl, height=\newh]{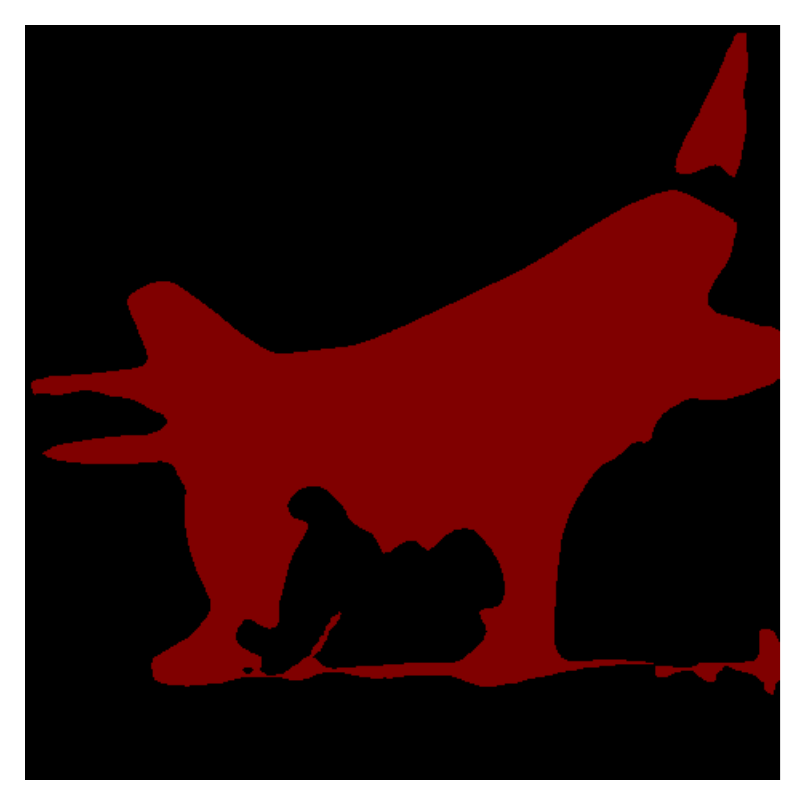} & \includegraphics[width=\newl, height=\newh]{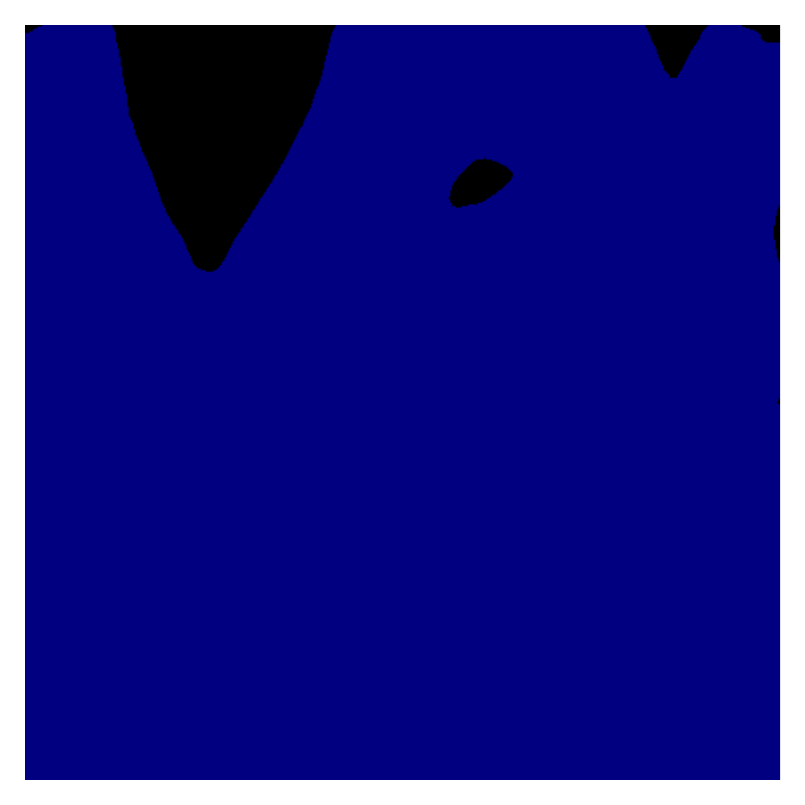} & \includegraphics[width=\newl, height=\newh]{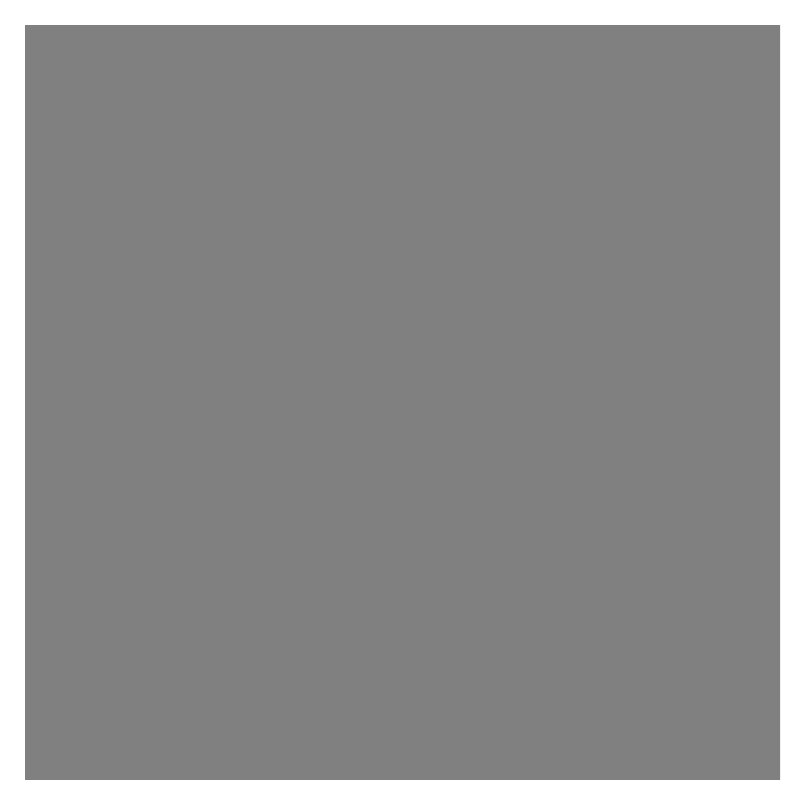}
\end{tabular}
\caption{
Same setting as in Fig.~\ref{fig:examples_voc_clean} for the 5-step \RobAT  model
} 
\label{fig:examples_voc_robust}
\end{figure}

\begin{figure}[t] \centering
\small
\tabcolsep=1.5pt
\newl=.16\columnwidth
\newh=\newl
\begin{tabular}{c | c c c c c} 
original & 0 & 0.25/255 & 0.5/255 & 1/255 & 2/255\\
\toprule
& \acc: 65.9\%& \acc: 54.9\%& \acc: 4.9\%& \acc: 0.0\%& \acc: 0.0\%\\
\includegraphics[width=\newl, height=\newh]{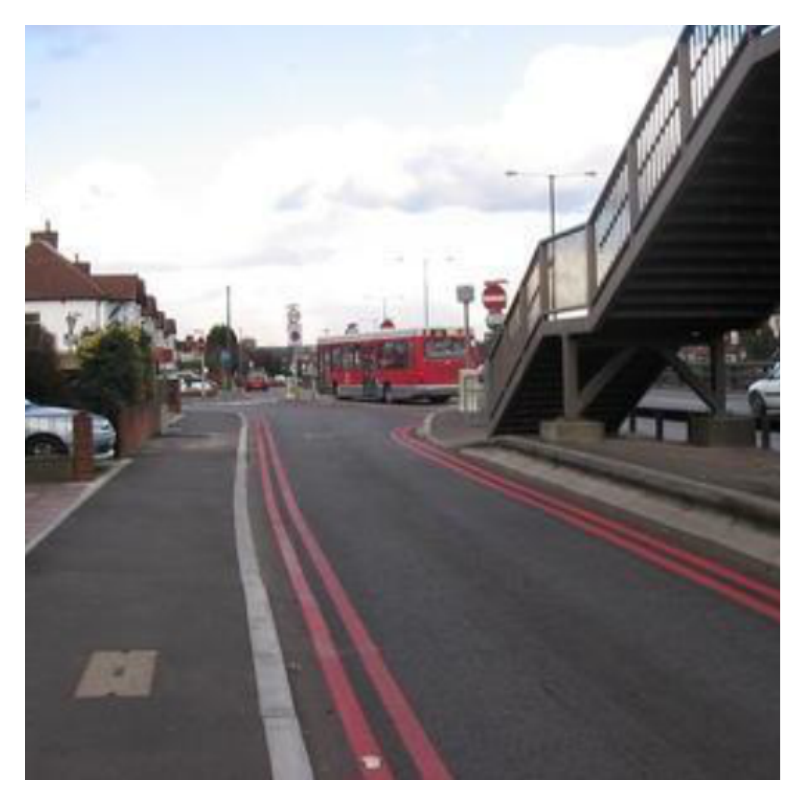} 
& \includegraphics[width=\newl, height=\newh]{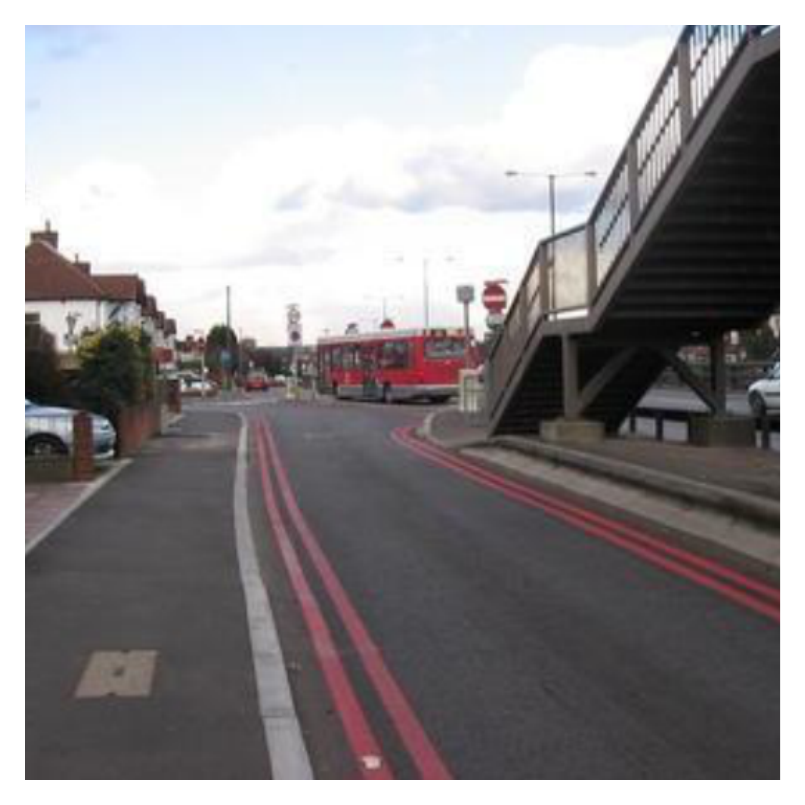} 
& \includegraphics[width=\newl, height=\newh]{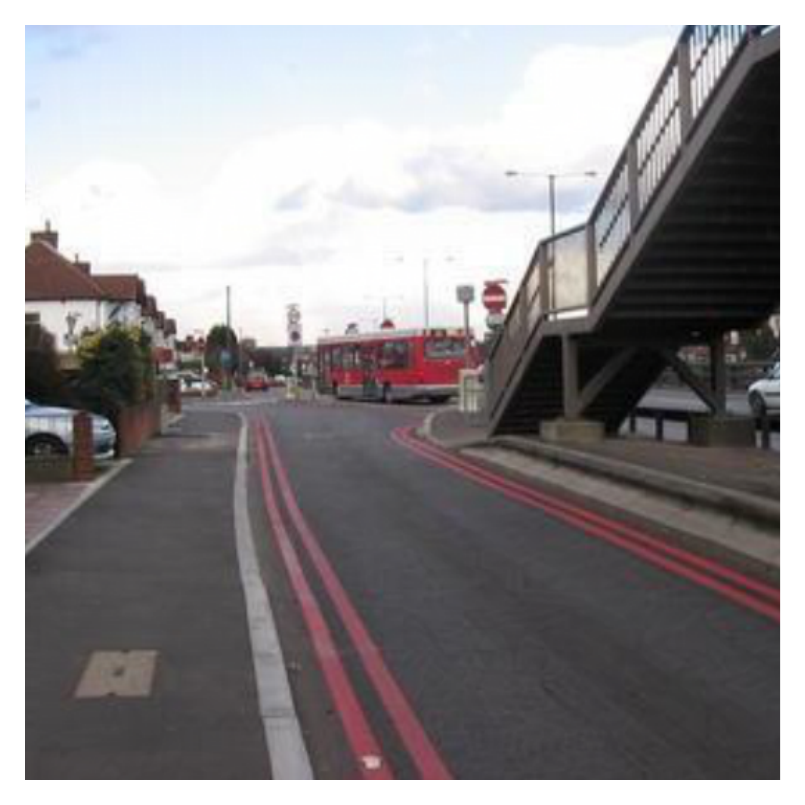} 
& \includegraphics[width=\newl, height=\newh]{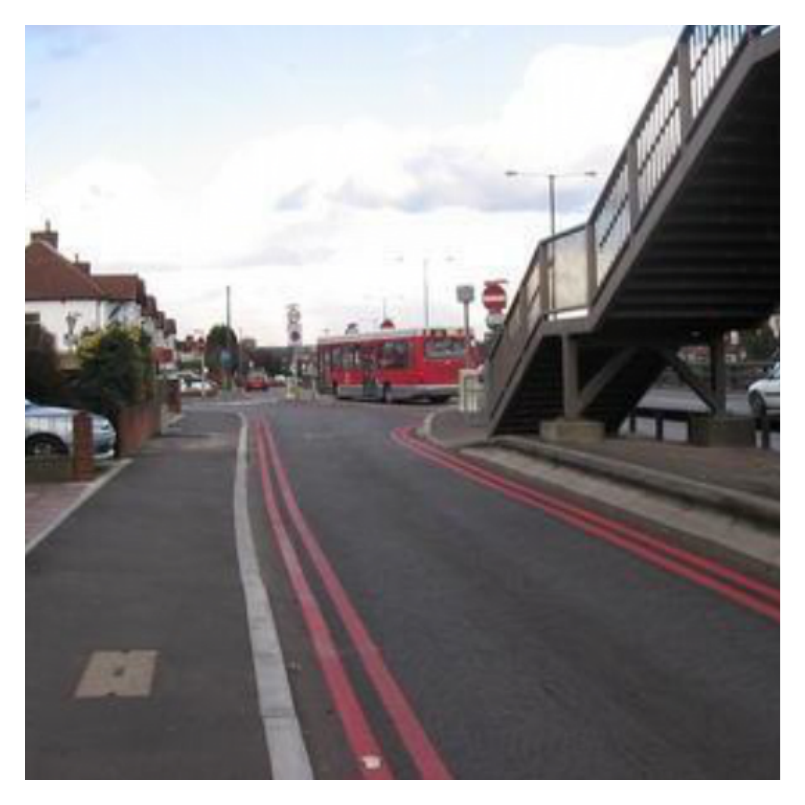} 
& \includegraphics[width=\newl, height=\newh]{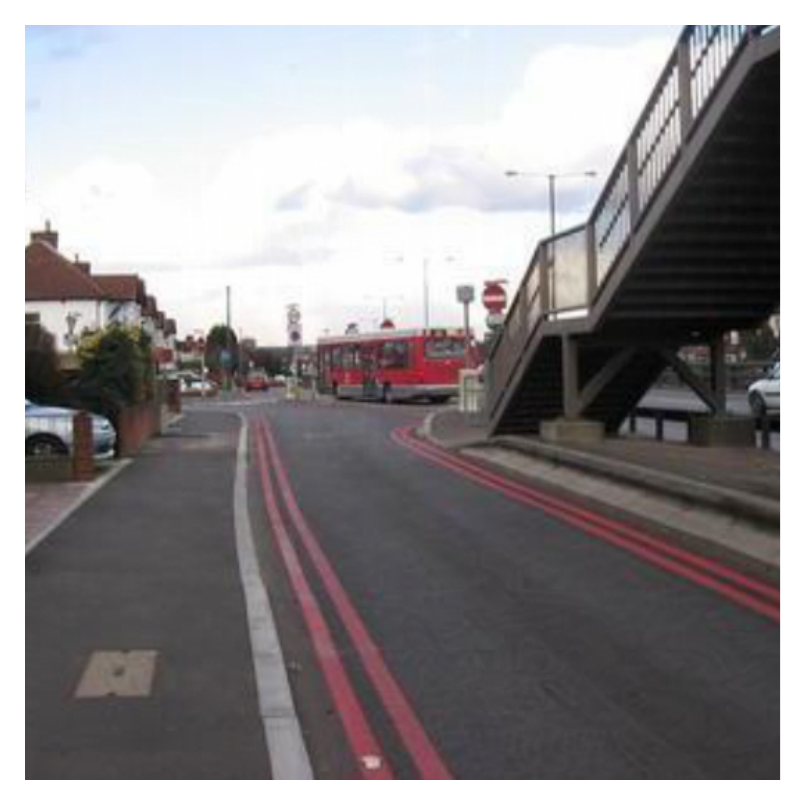} 
& \includegraphics[width=\newl, height=\newh]{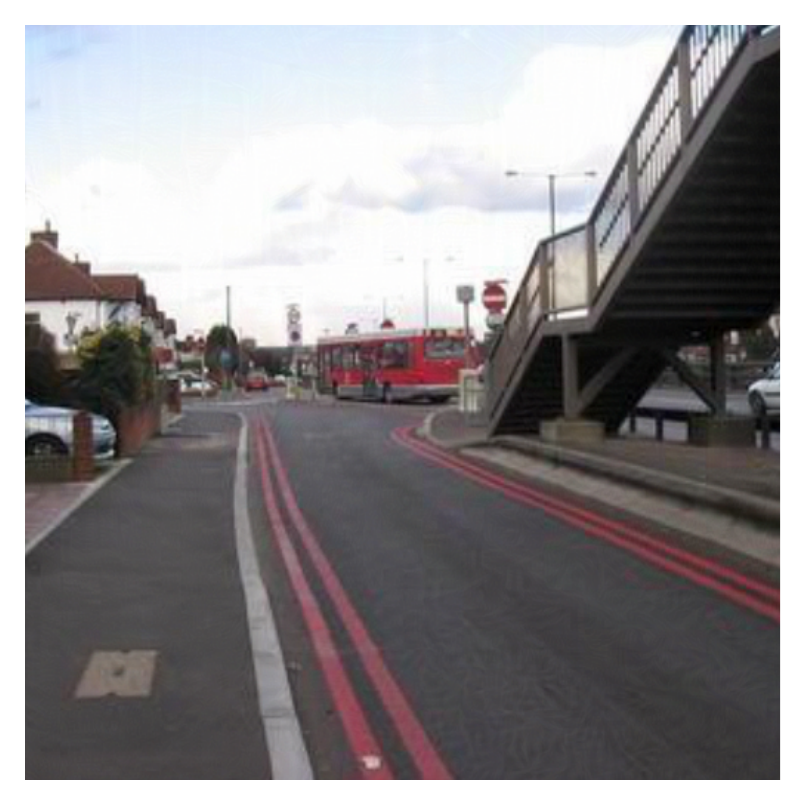} \\ 

\includegraphics[width=\newl, height=\newh]{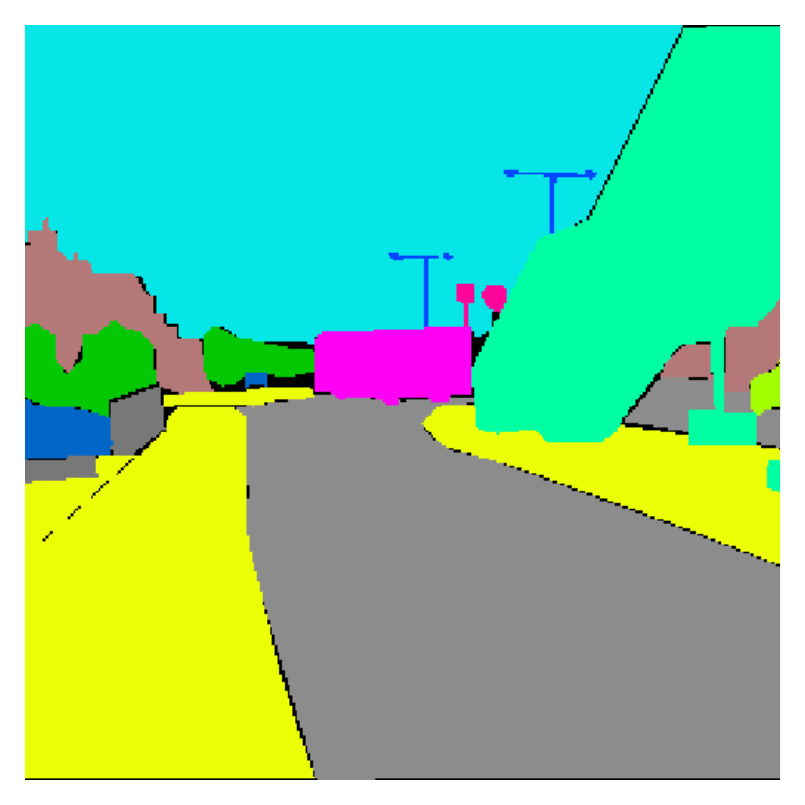} 
& \includegraphics[width=\newl, height=\newh]{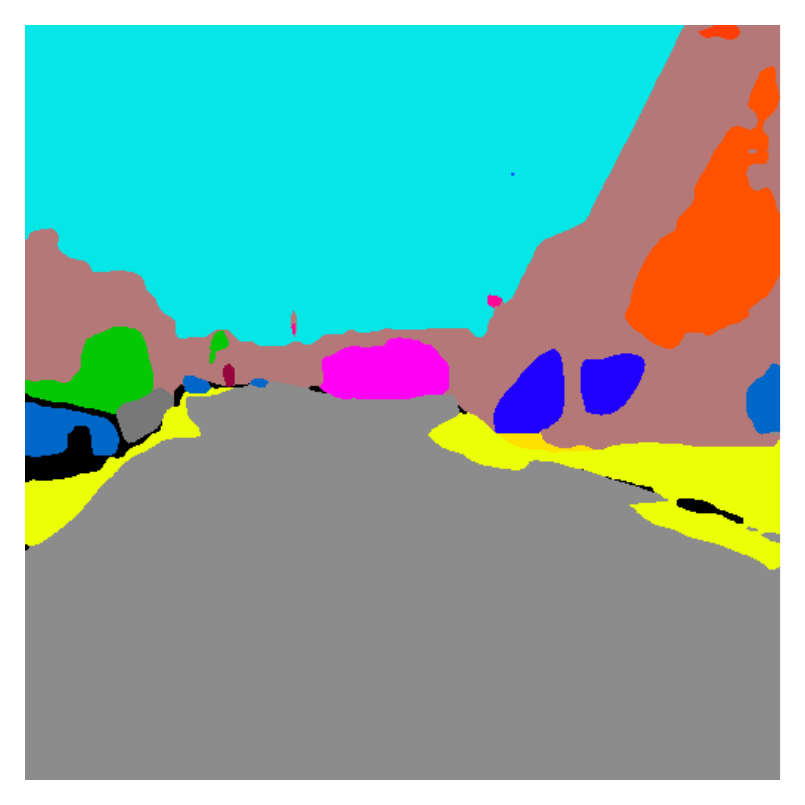} 
& \includegraphics[width=\newl, height=\newh]{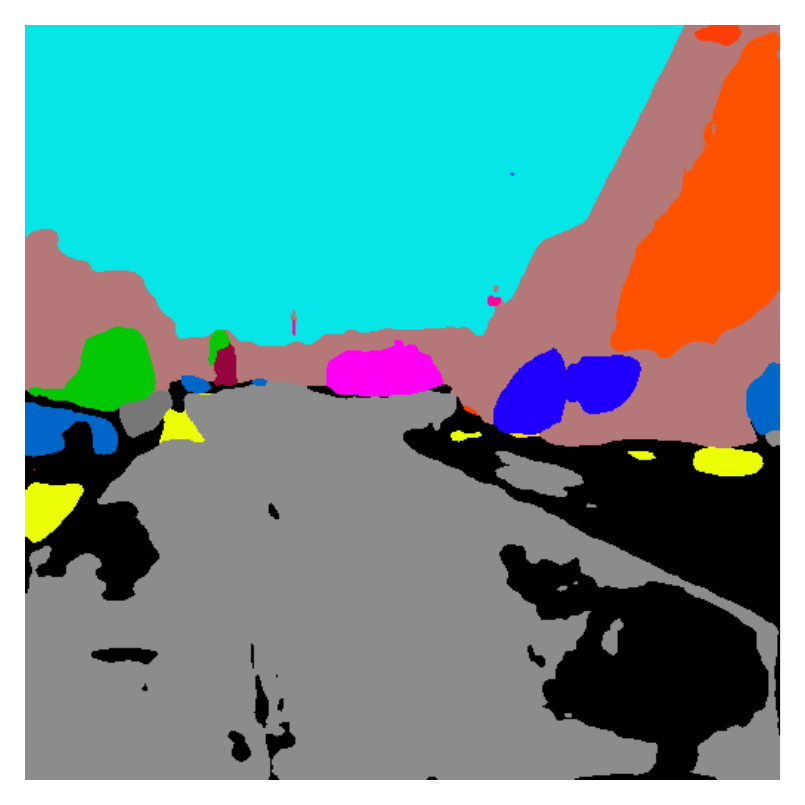} 
& \includegraphics[width=\newl, height=\newh]{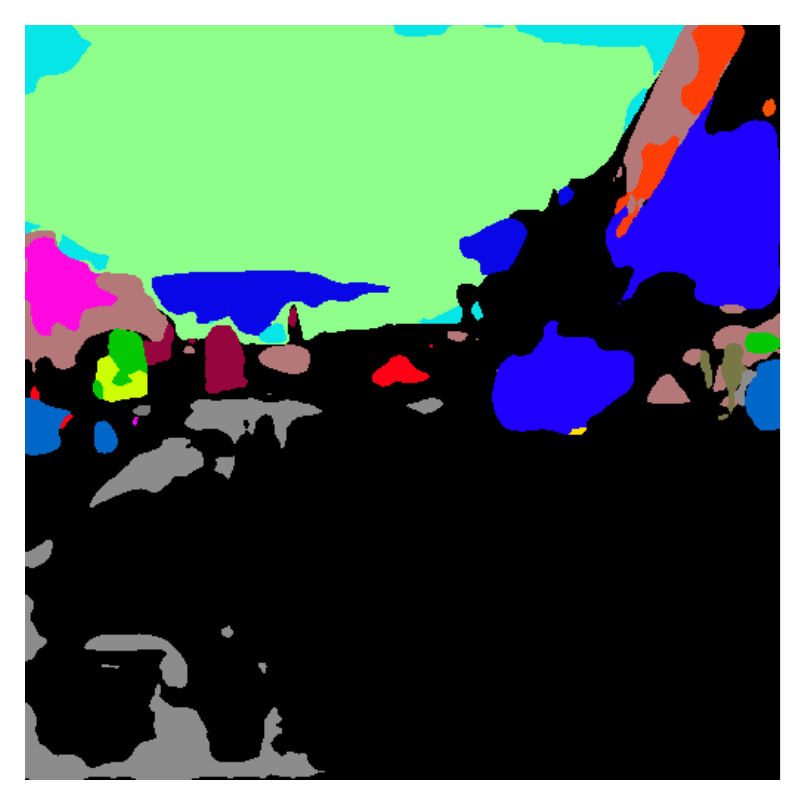}
& \includegraphics[width=\newl, height=\newh]{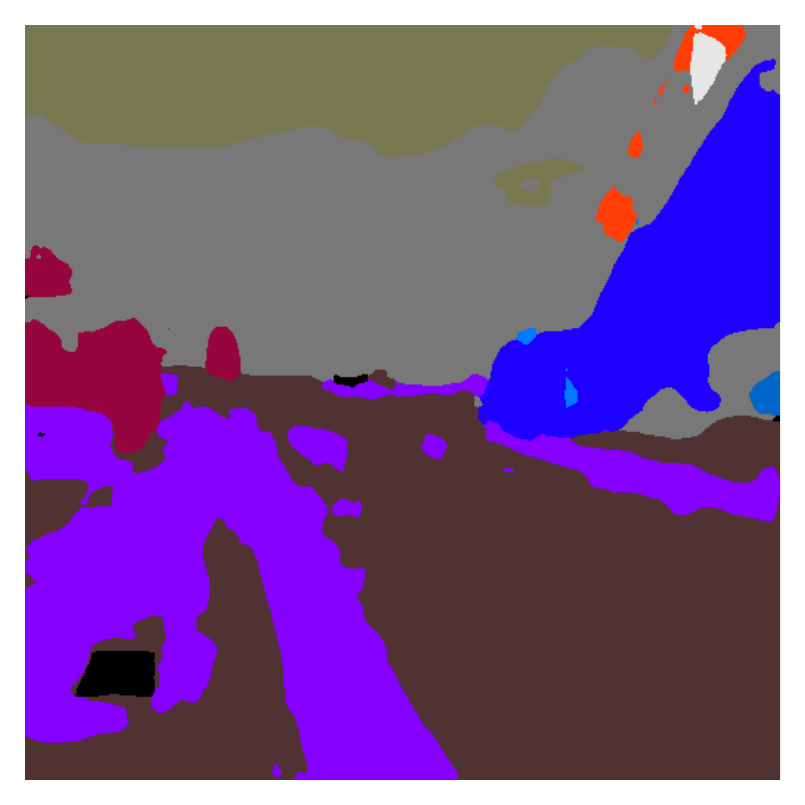} 
& \includegraphics[width=\newl, height=\newh]{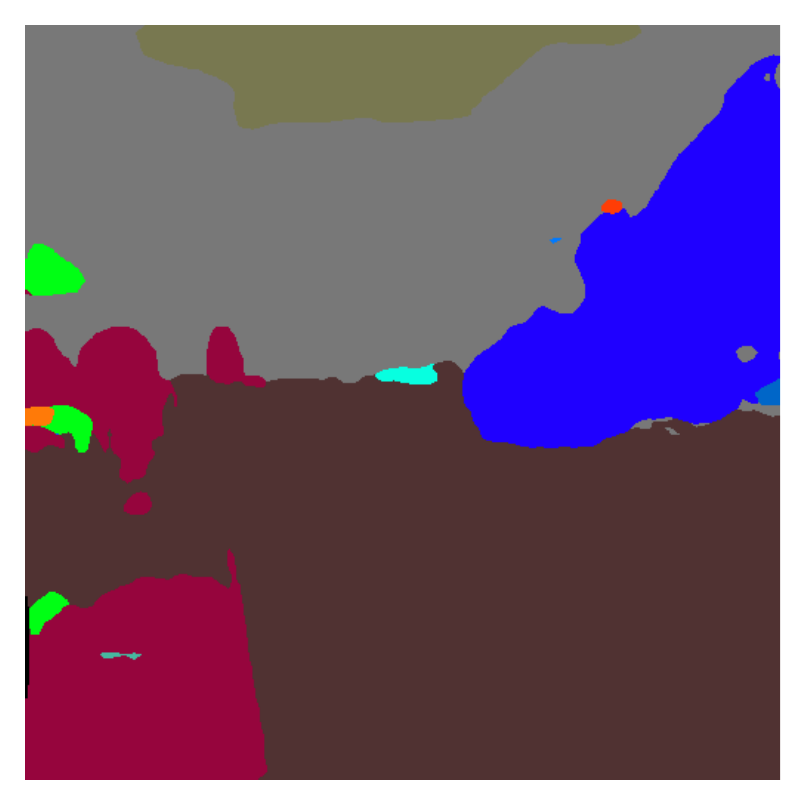}\\
\midrule
& \acc: 81.2\%& \acc: 47.9\%& \acc: 21.9\%& \acc: 2.6\%& \acc: 0.0\%\\
\includegraphics[width=\newl, height=\newh]{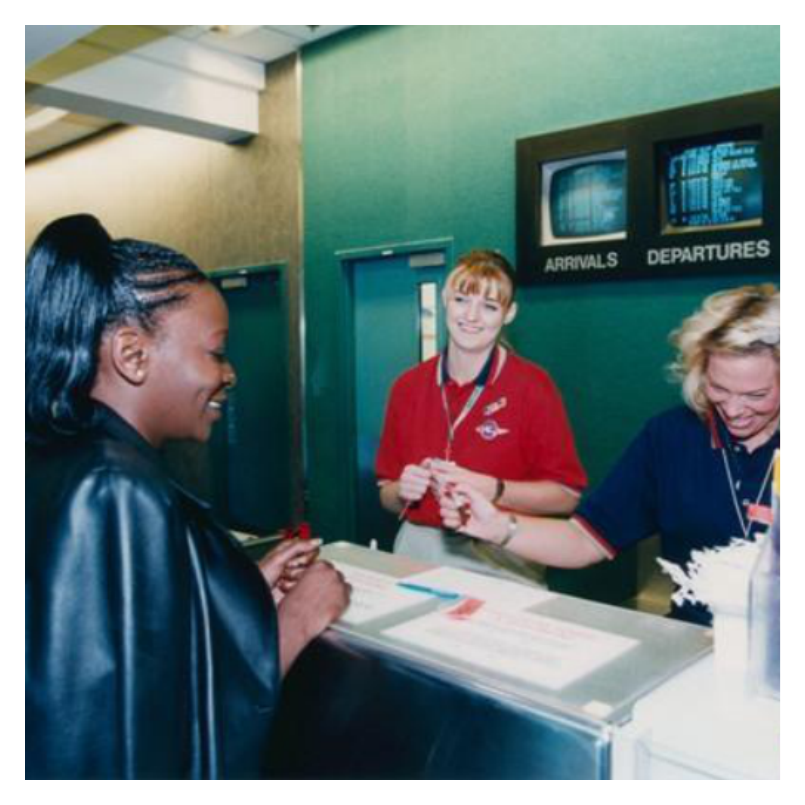} & \includegraphics[width=\newl, height=\newh]{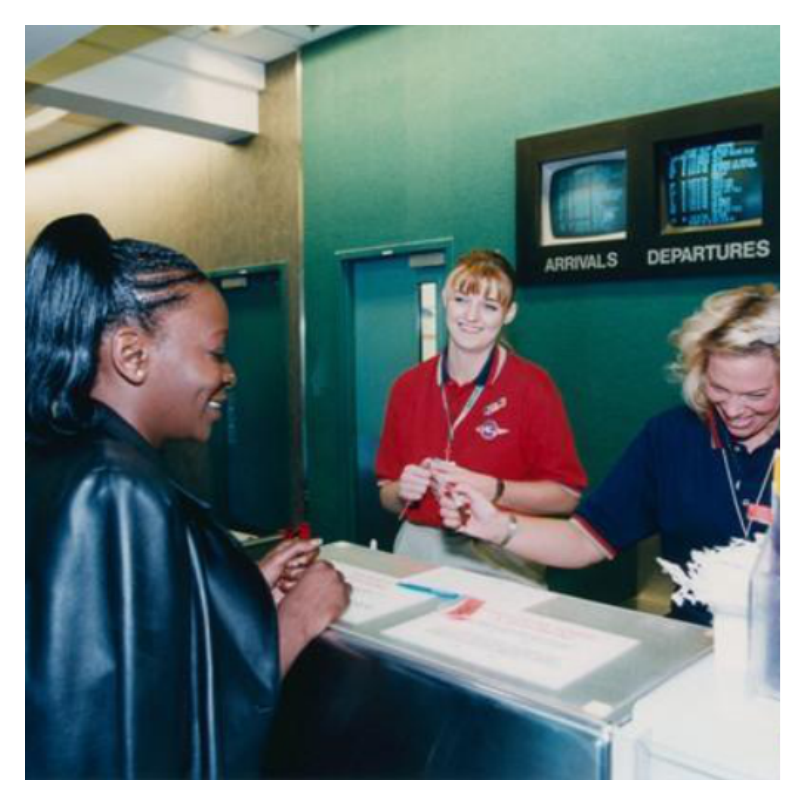} & \includegraphics[width=\newl, height=\newh]{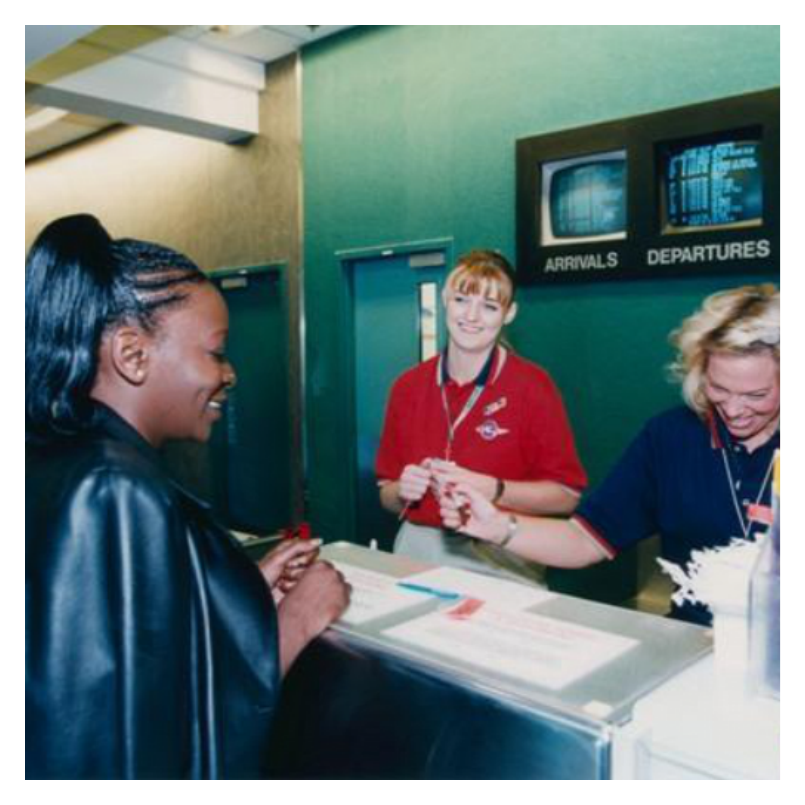} & \includegraphics[width=\newl, height=\newh]{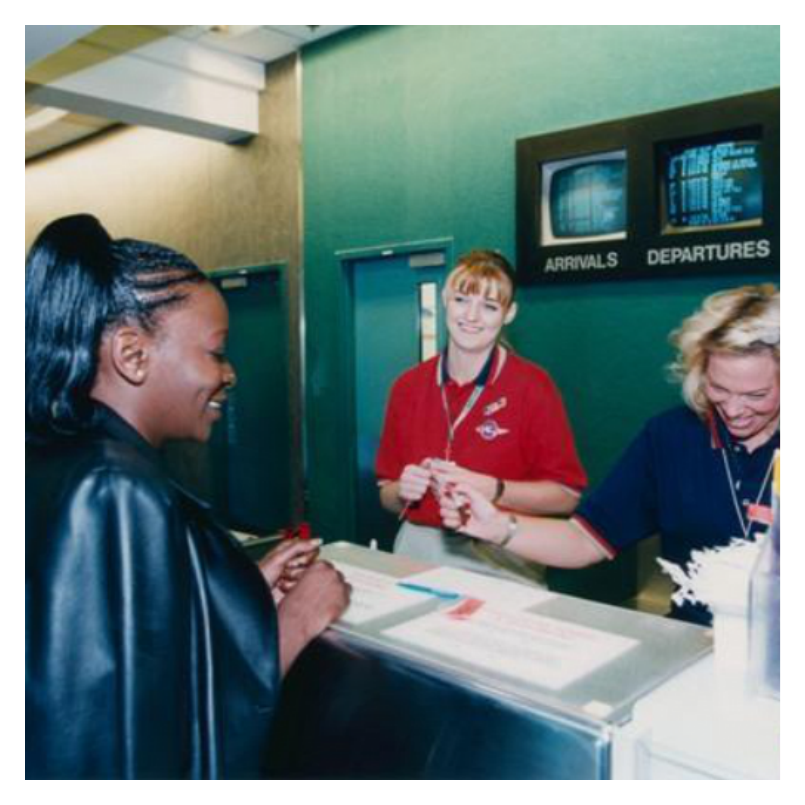} & \includegraphics[width=\newl, height=\newh]{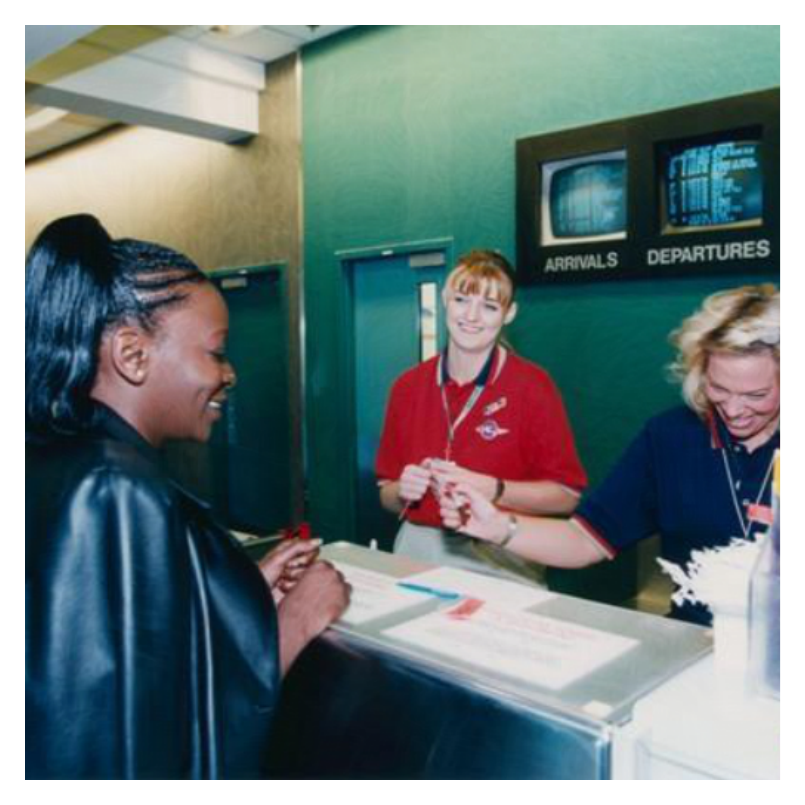} & \includegraphics[width=\newl, height=\newh]{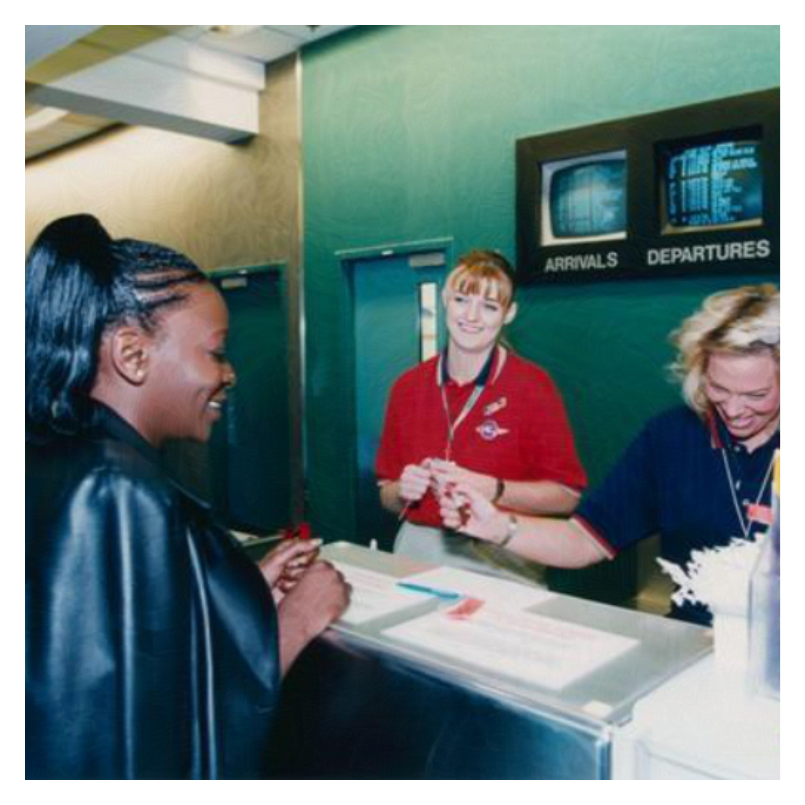} \\ \includegraphics[width=\newl, height=\newh]{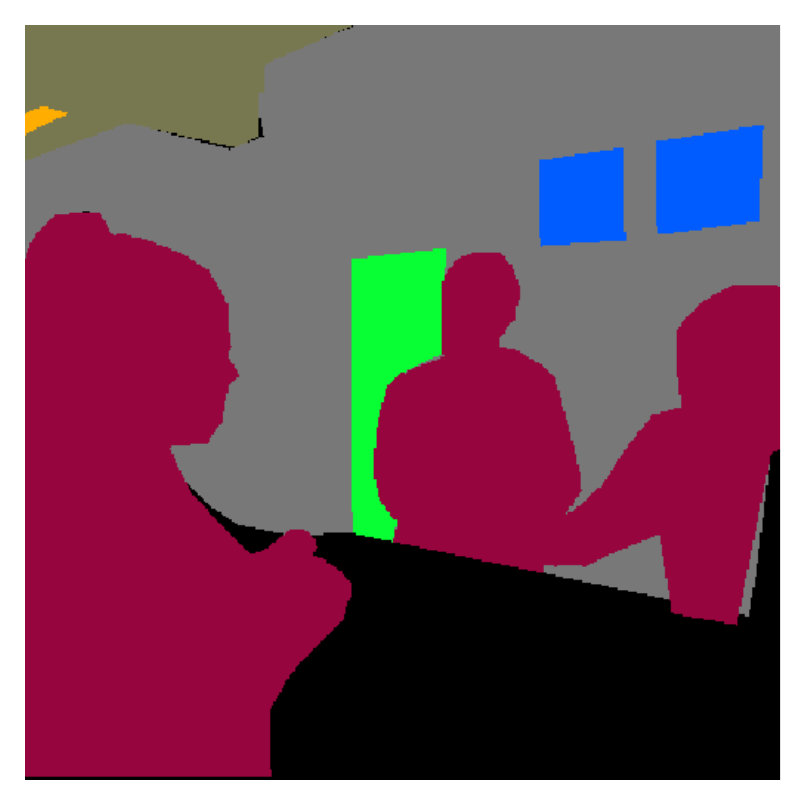} & \includegraphics[width=\newl, height=\newh]{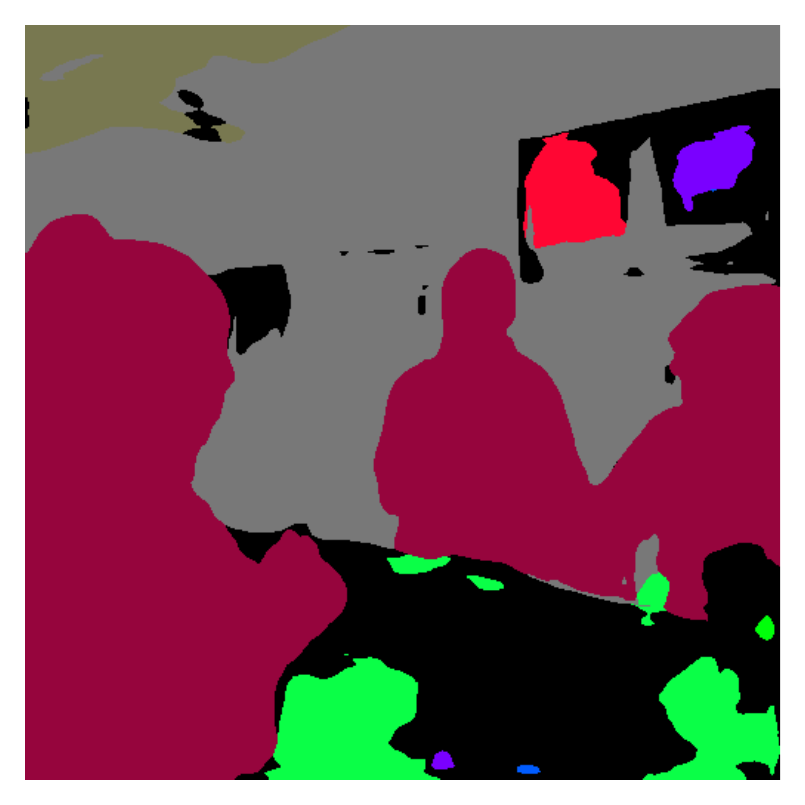} & \includegraphics[width=\newl, height=\newh]{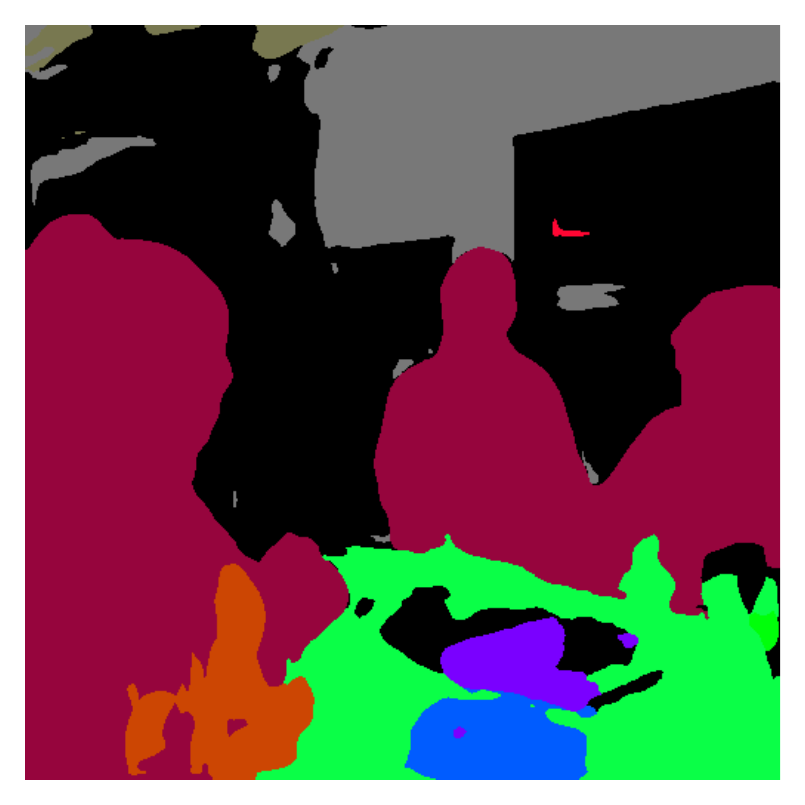} & \includegraphics[width=\newl, height=\newh]{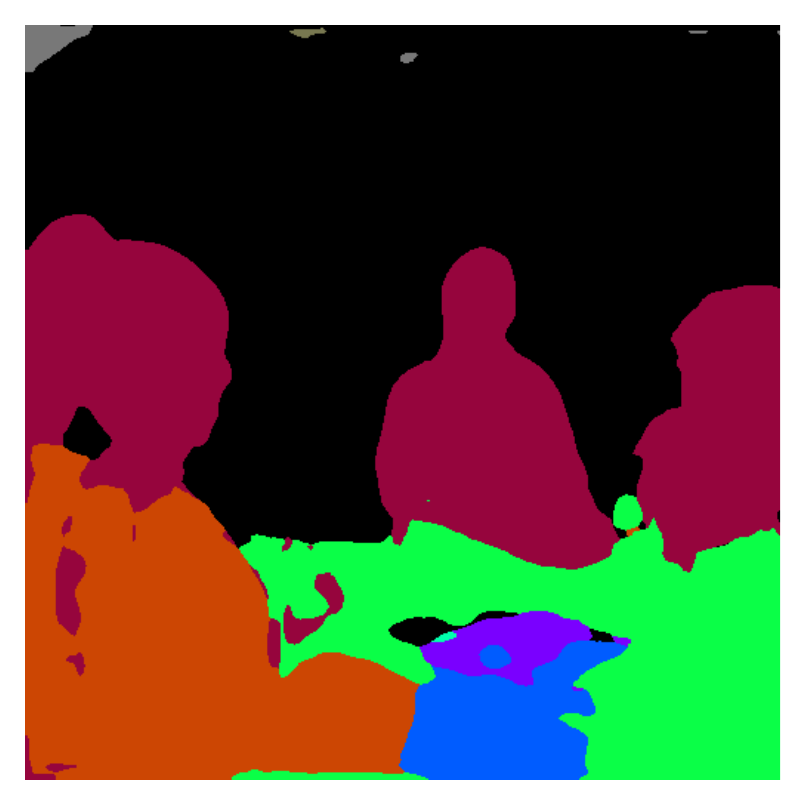} & \includegraphics[width=\newl, height=\newh]{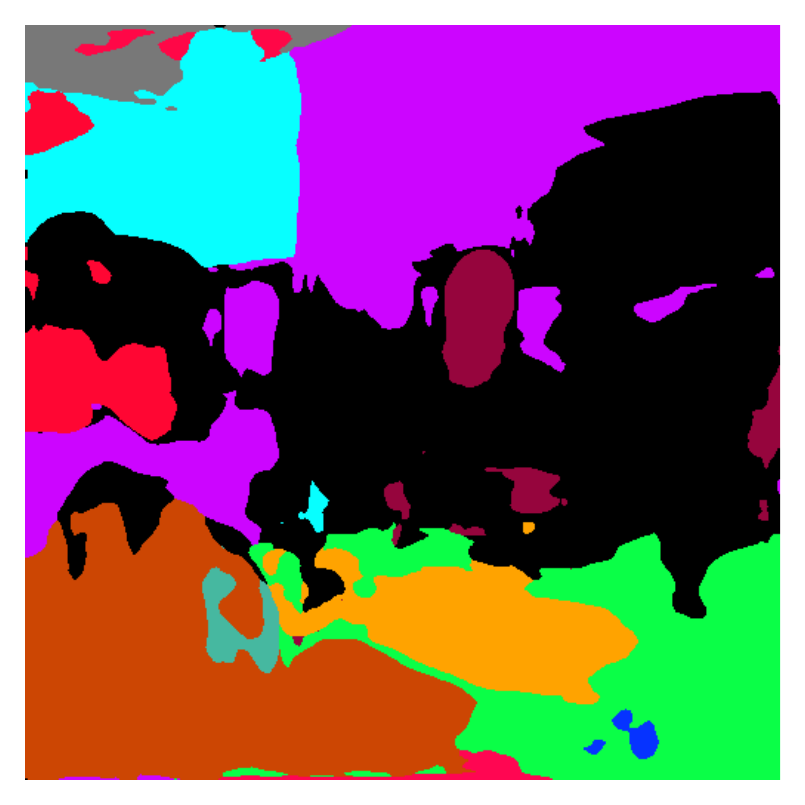} & \includegraphics[width=\newl, height=\newh]{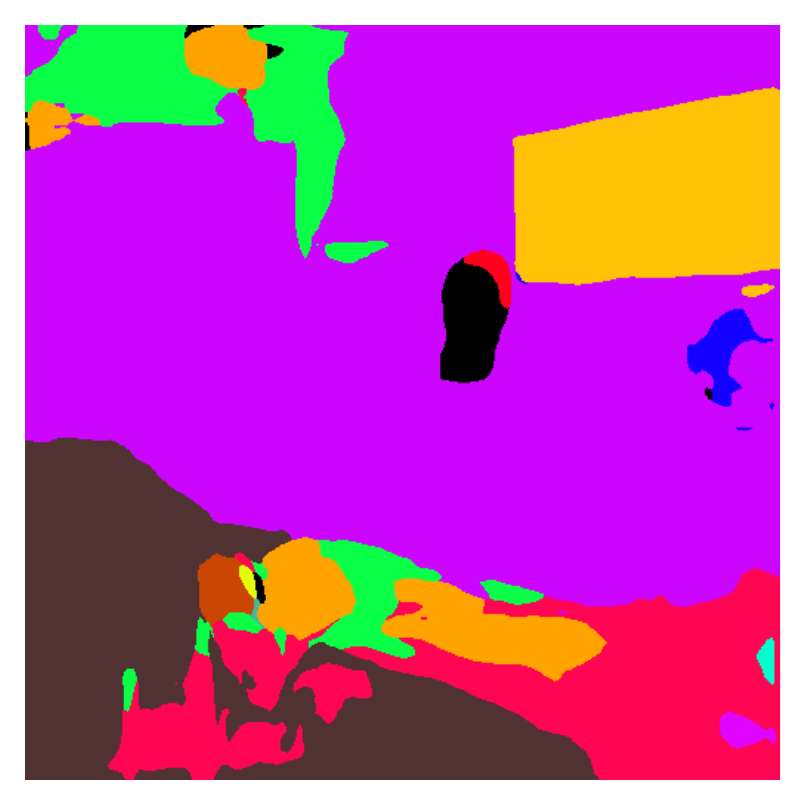}
\end{tabular}
\caption{
Visualizing the perturbed images, corresponding predicted masks and \acc for increasing radii. The attacks are generated on the clean model on \ade with APGD on $\L_\textrm{Mask-CE}$. Original image and ground truth mask in the first column.}
\label{fig:examples_ade_clean}

\end{figure}

\begin{figure}[b] \centering
\small
\tabcolsep=1.5pt

\newl=.16\columnwidth
\newh=\newl
\begin{tabular}{c | c c c c c} 
original & 0 & 4/255 & 8/255 & 12/255 & 16/255\\
\toprule 
& \acc: 61.3\%& \acc: 58.6\%& \acc: 29.7\%& \acc: 1.6\%& \acc: 0.0\%\\
\includegraphics[width=\newl, height=\newh]{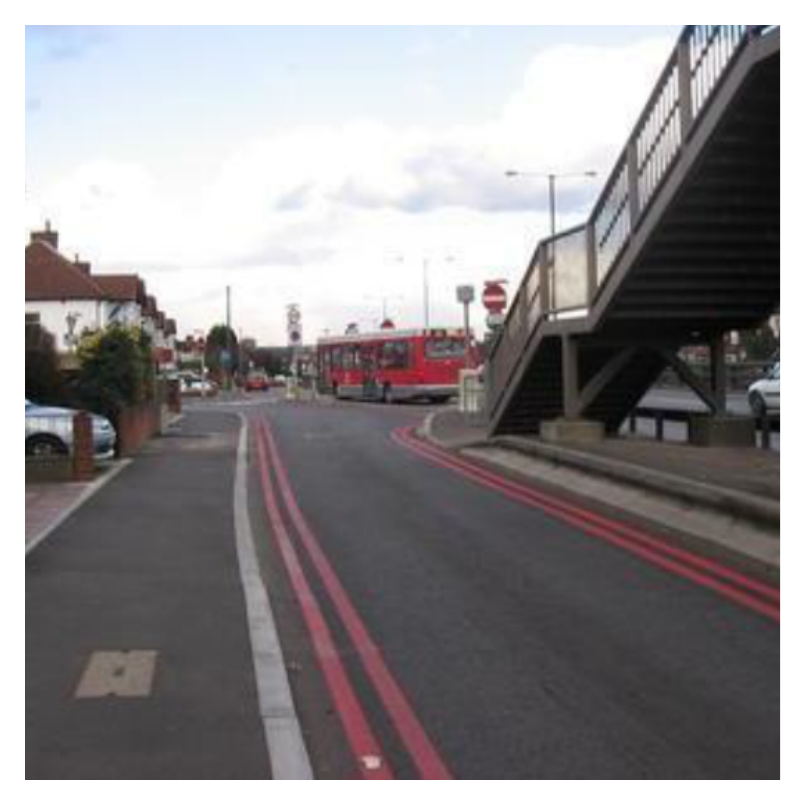} & \includegraphics[width=\newl, height=\newh]{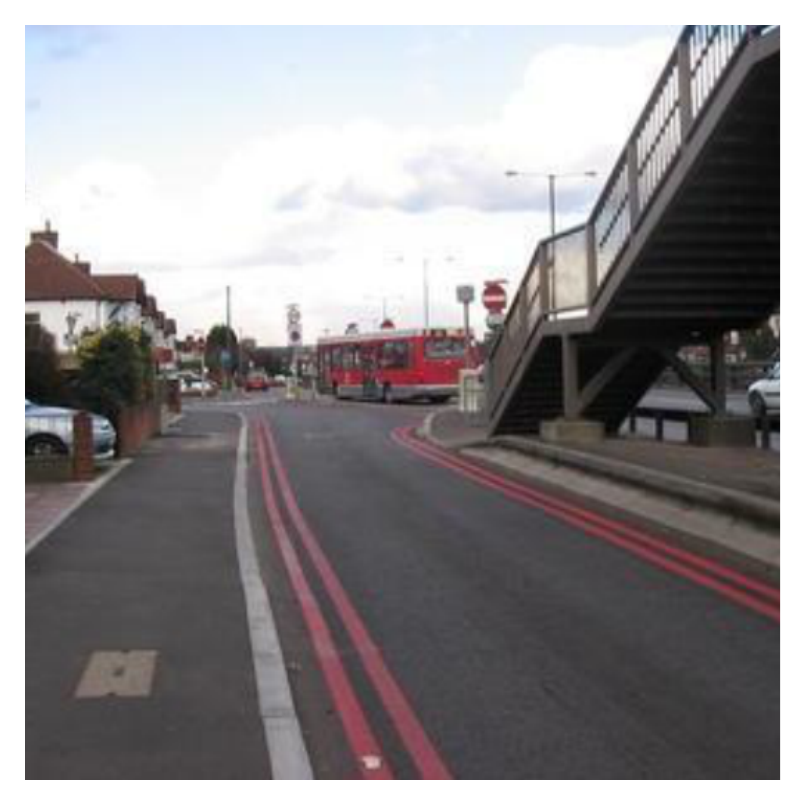} & \includegraphics[width=\newl, height=\newh]{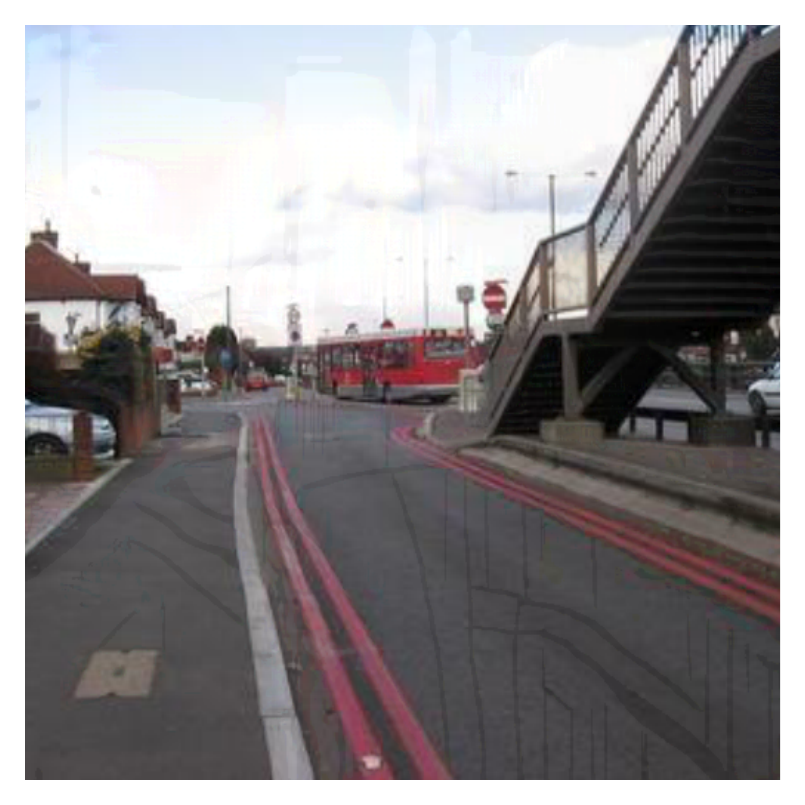} & \includegraphics[width=\newl, height=\newh]{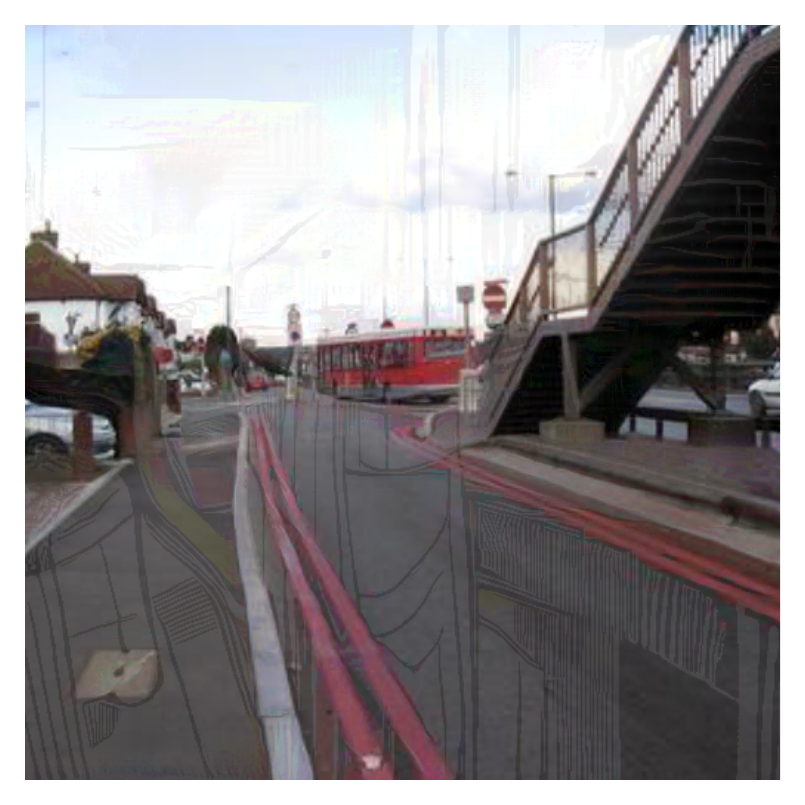} & \includegraphics[width=\newl, height=\newh]{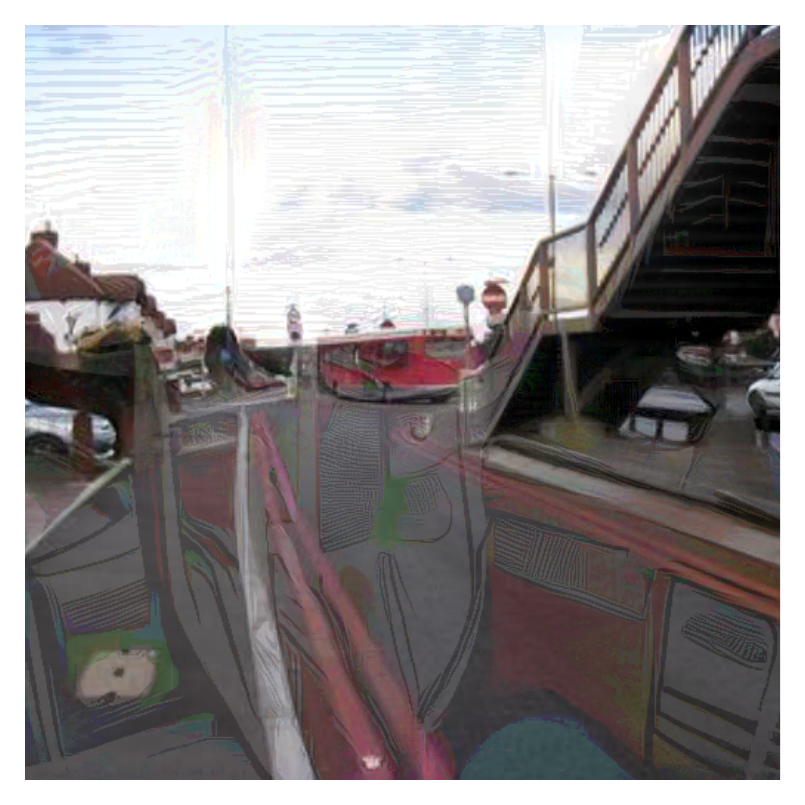} & \includegraphics[width=\newl, height=\newh]{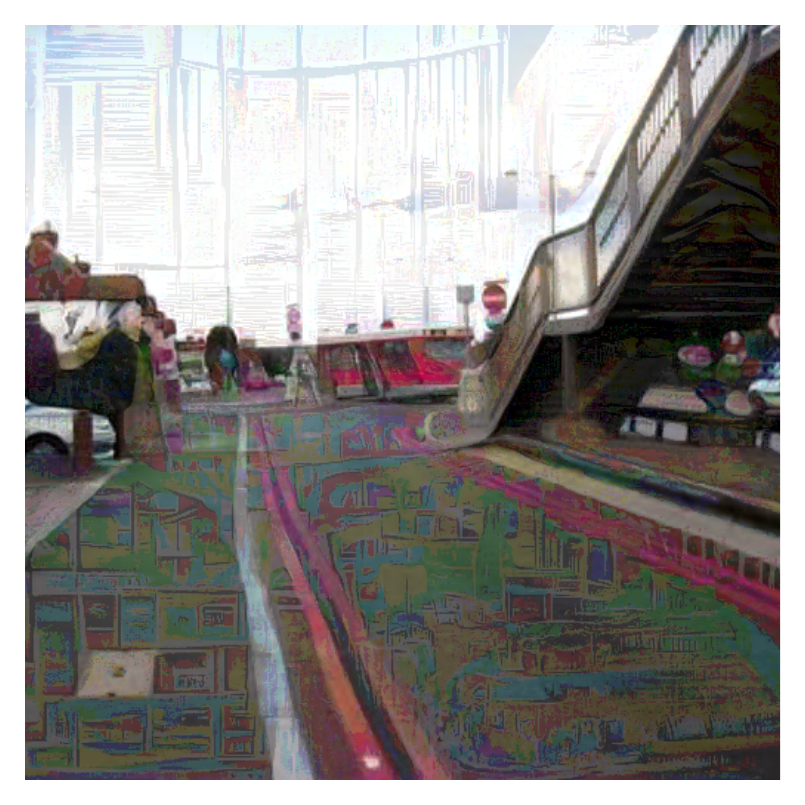} \\
\includegraphics[width=\newl, height=\newh]{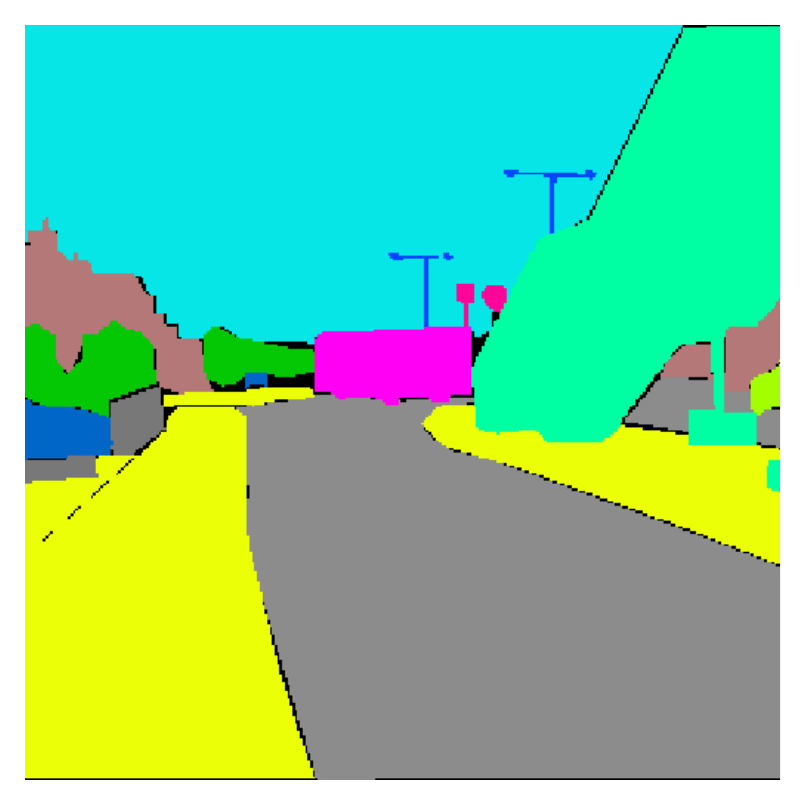} & \includegraphics[width=\newl, height=\newh]{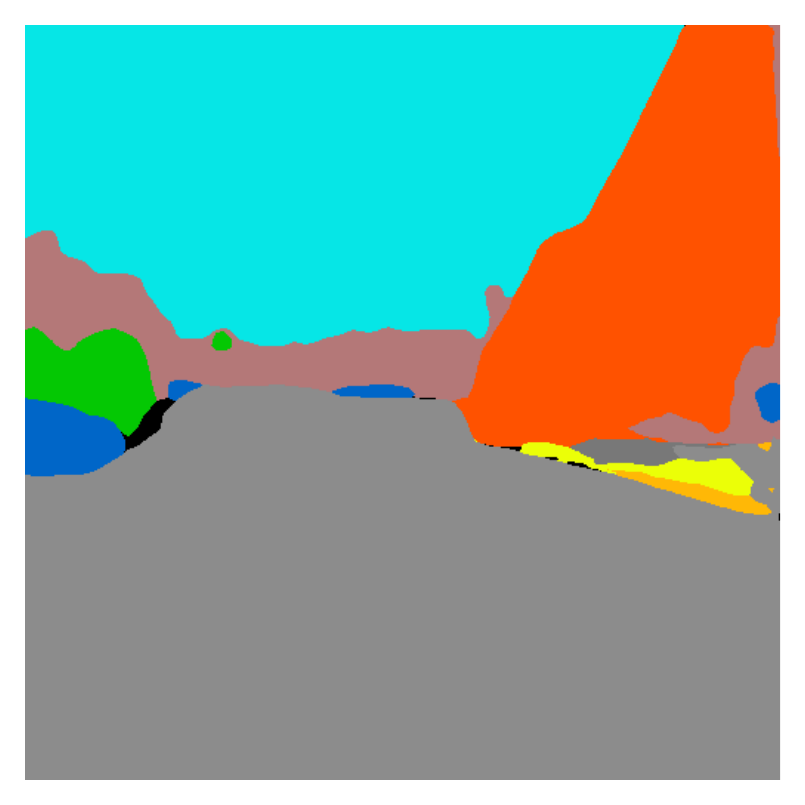} & \includegraphics[width=\newl, height=\newh]{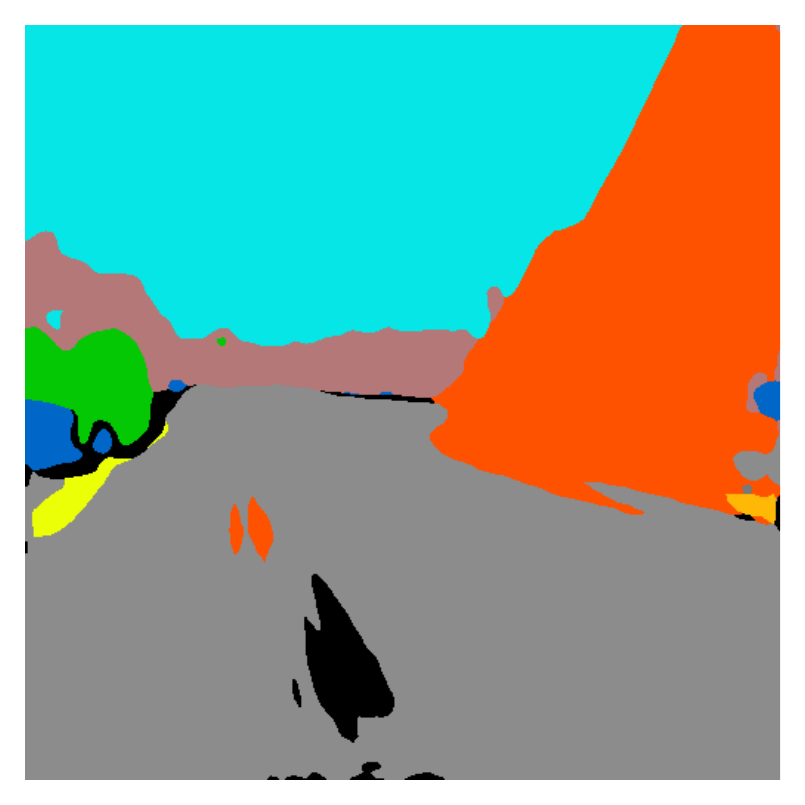} & \includegraphics[width=\newl, height=\newh]{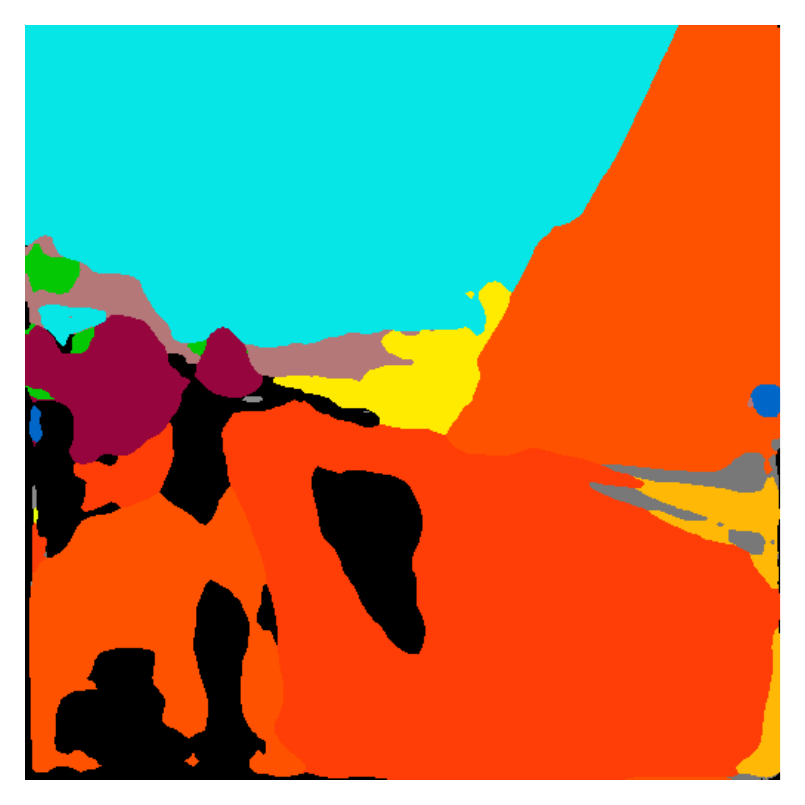} & \includegraphics[width=\newl, height=\newh]{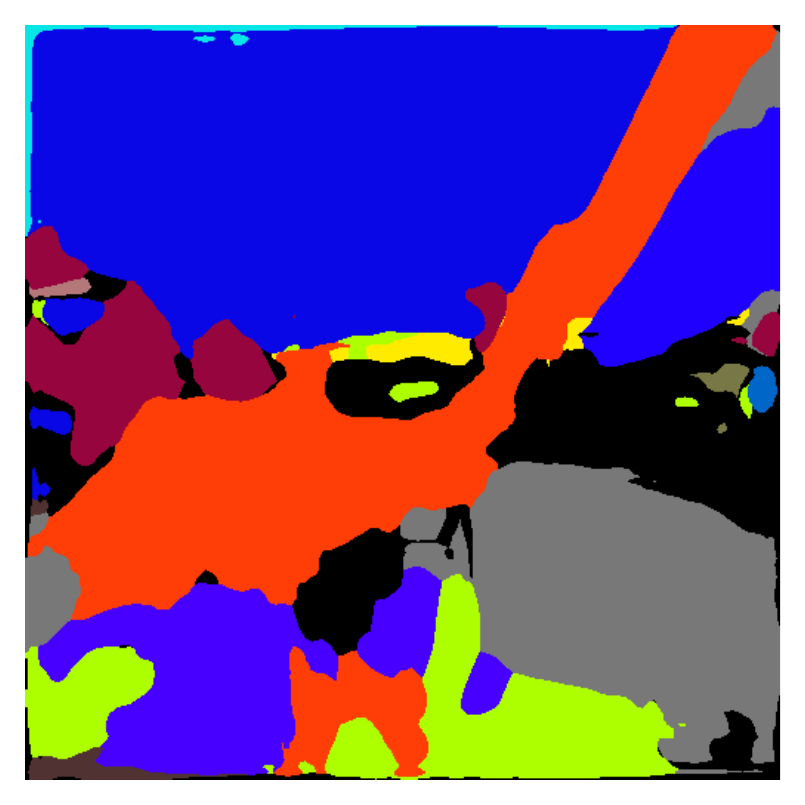} & \includegraphics[width=\newl, height=\newh]{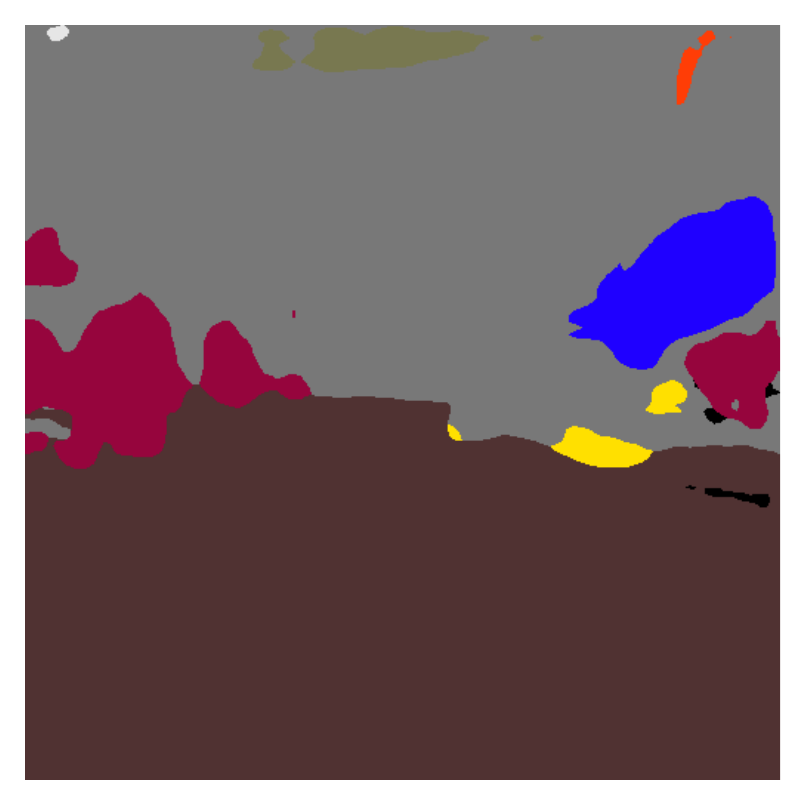}\\
\midrule
& \acc: 84.4\%& \acc: 67.3\%& \acc: 32.8\%& \acc: 6.0\%& \acc: 0.0\%\\
\includegraphics[width=\newl, height=\newh]{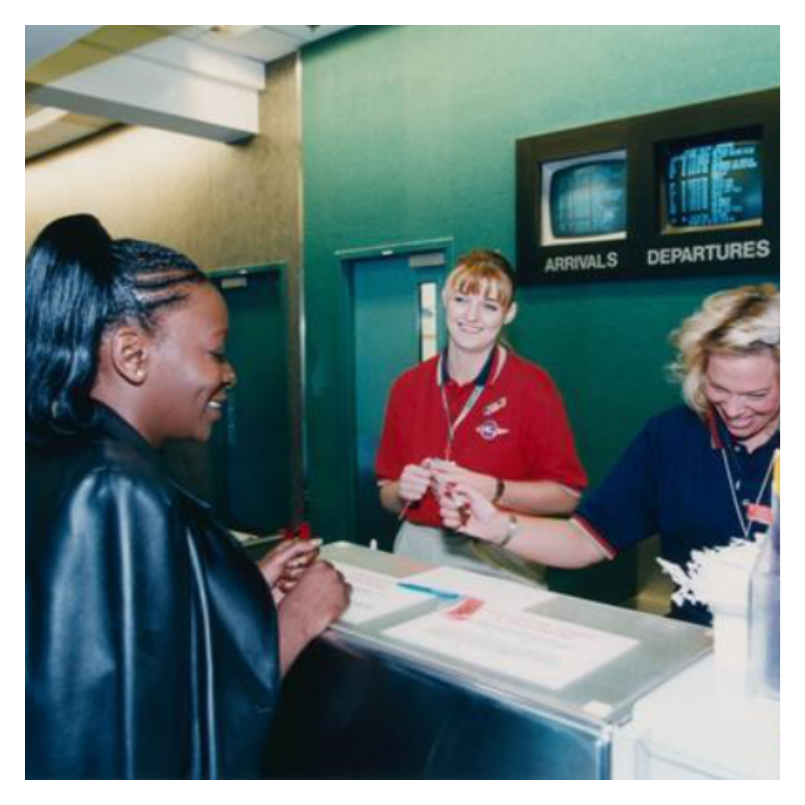} & \includegraphics[width=\newl, height=\newh]{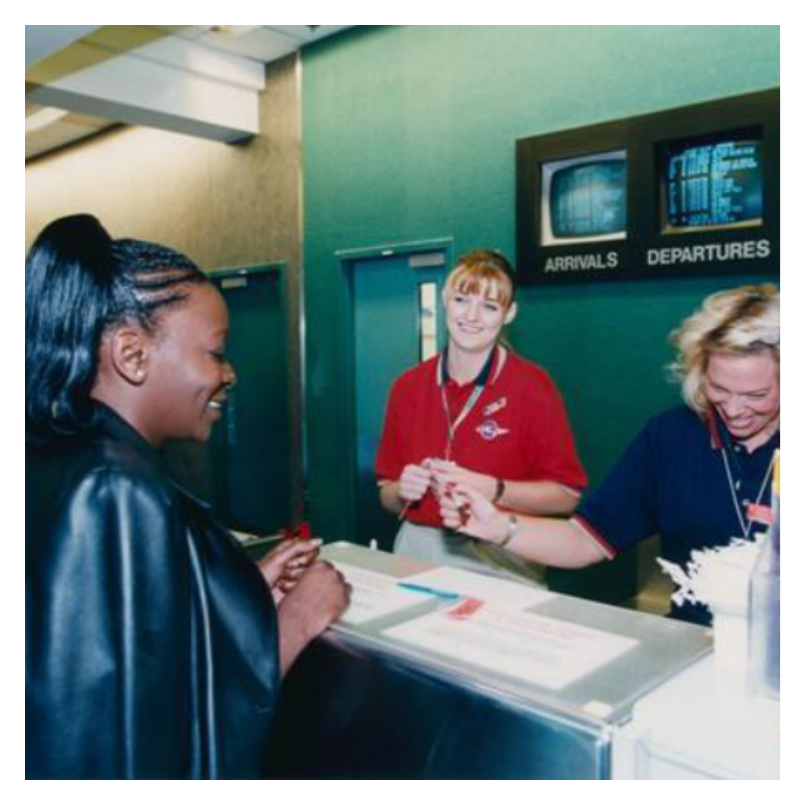} & \includegraphics[width=\newl, height=\newh]{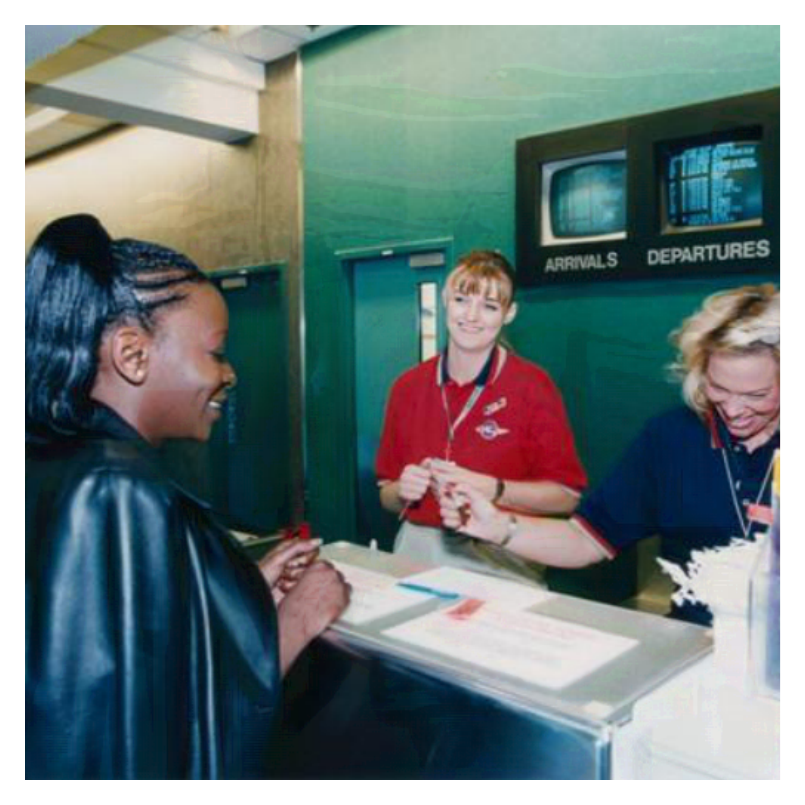} & \includegraphics[width=\newl, height=\newh]{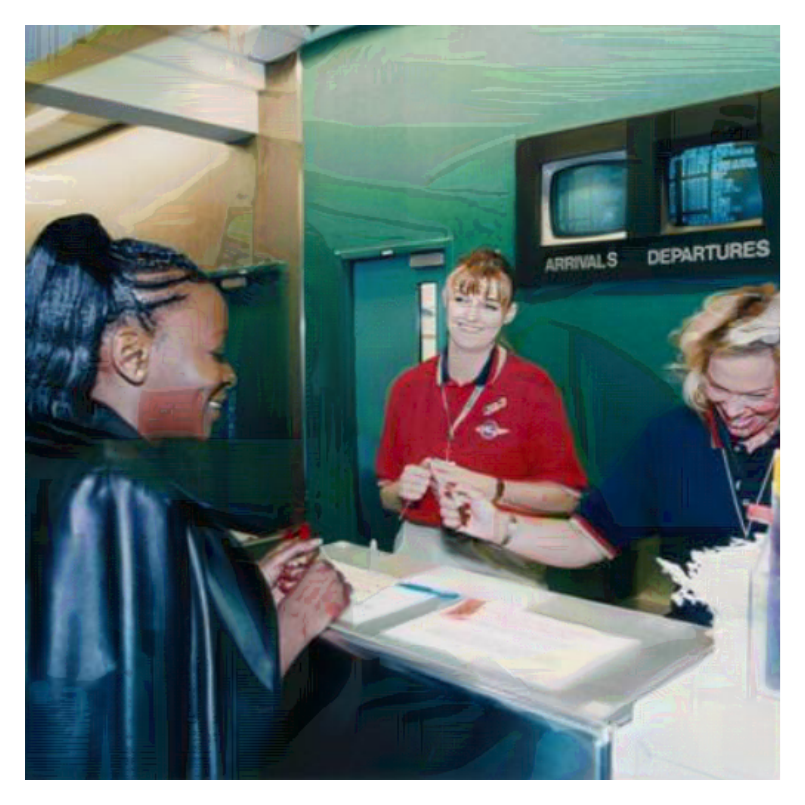} & \includegraphics[width=\newl, height=\newh]{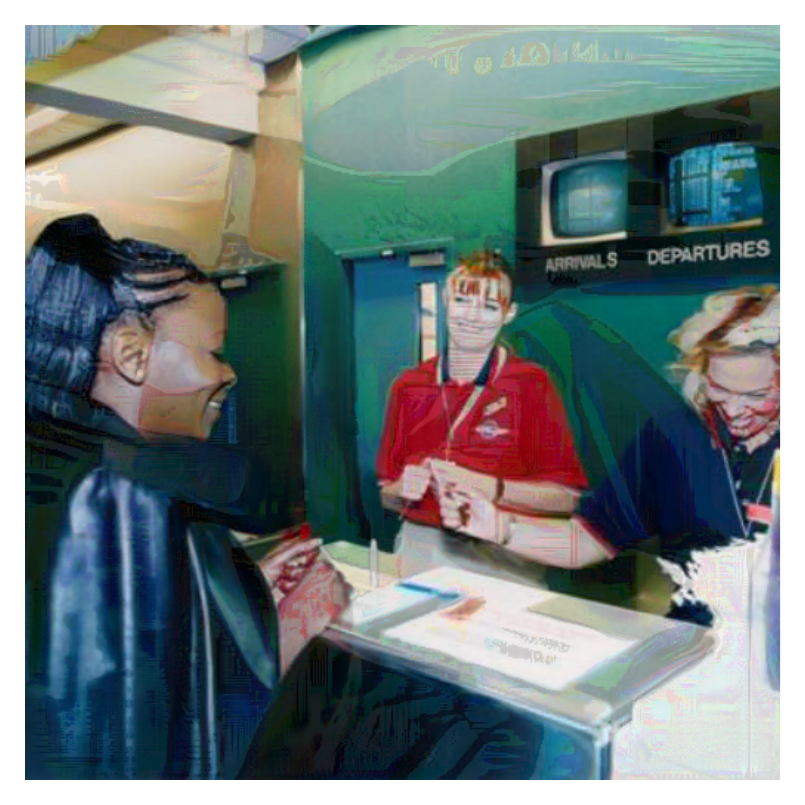} & \includegraphics[width=\newl, height=\newh]{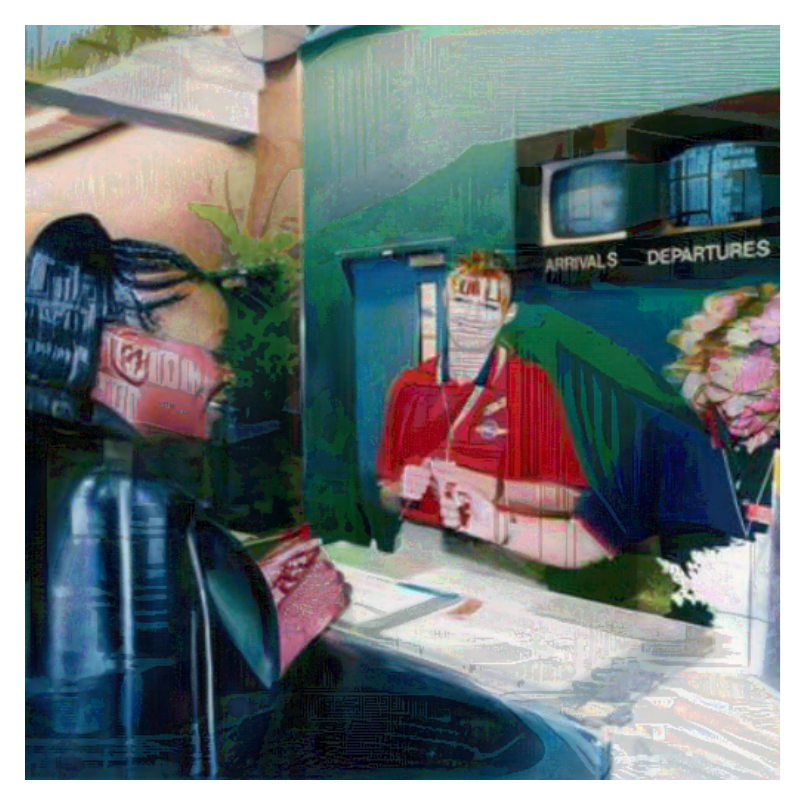} \\

\includegraphics[width=\newl, height=\newh]{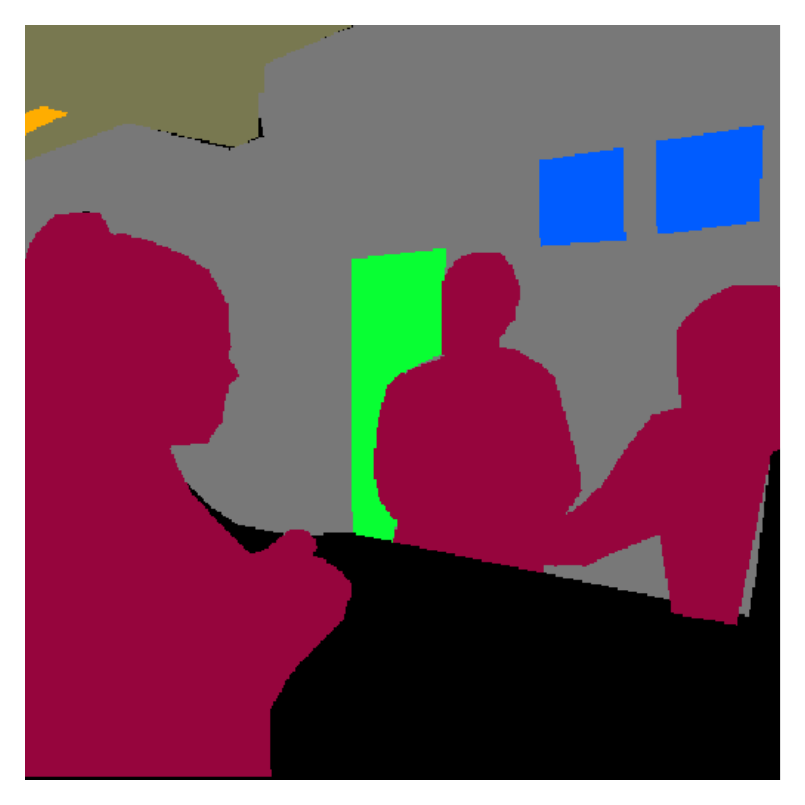} & \includegraphics[width=\newl, height=\newh]{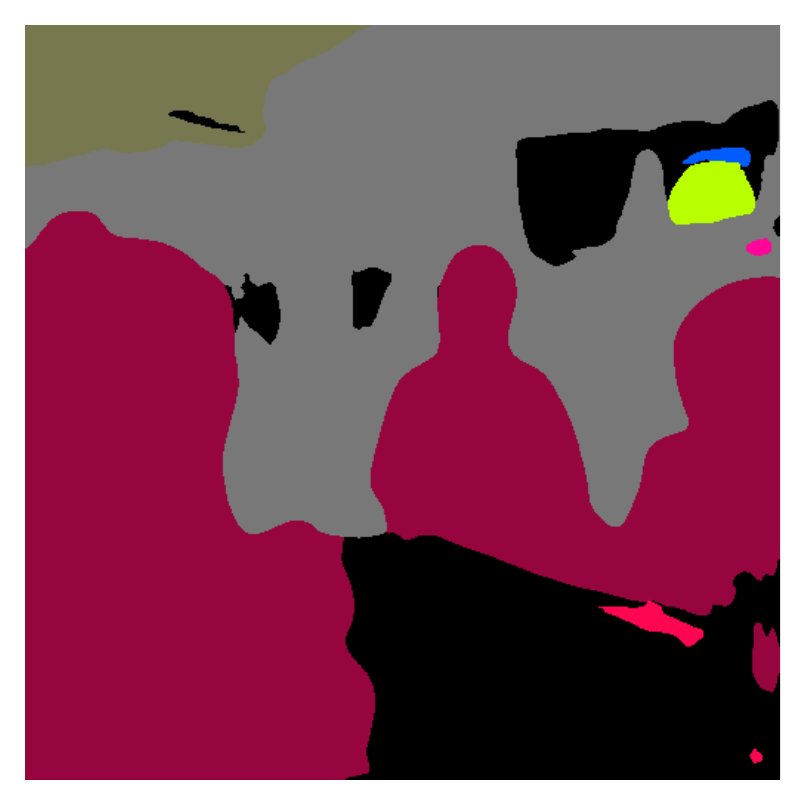} & \includegraphics[width=\newl, height=\newh]{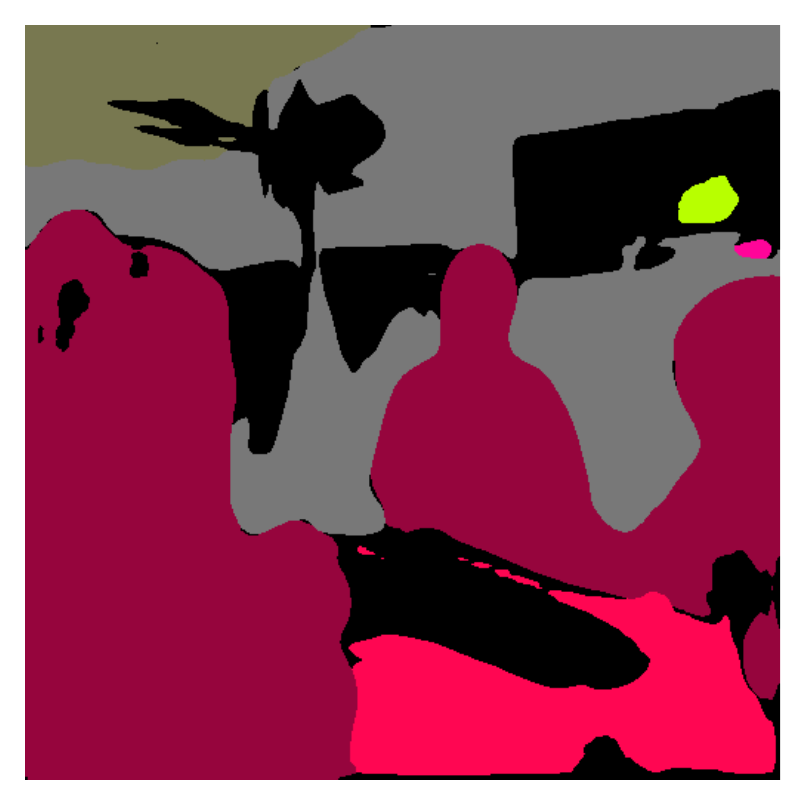} & \includegraphics[width=\newl, height=\newh]{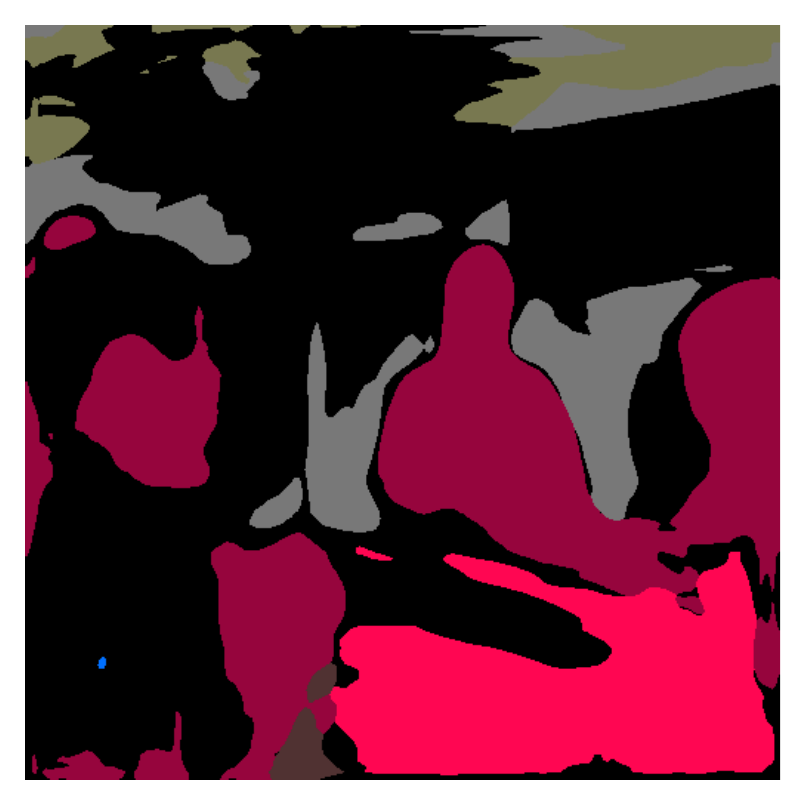} & \includegraphics[width=\newl, height=\newh]{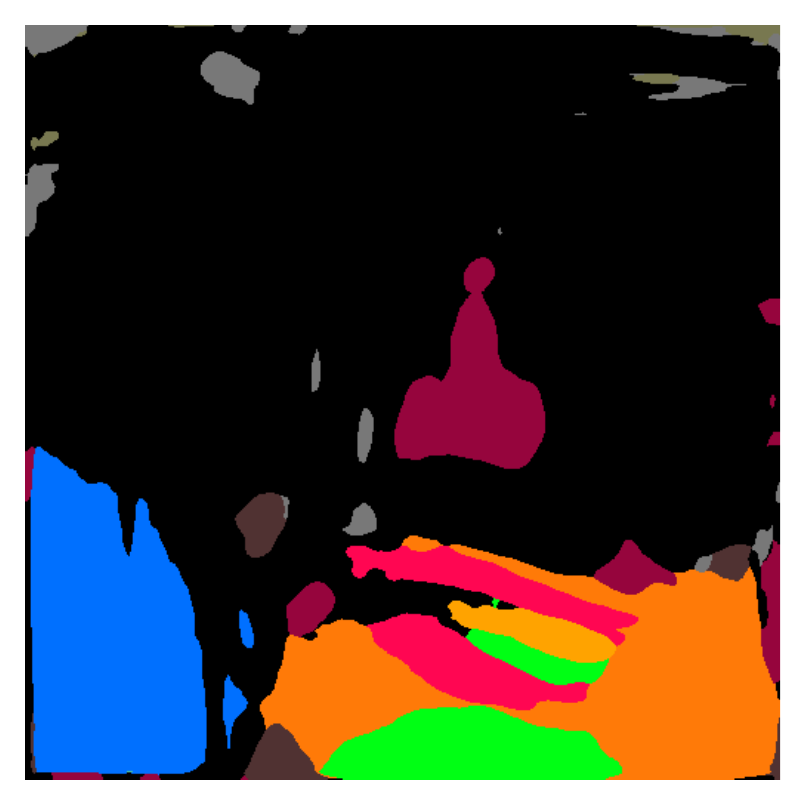} & \includegraphics[width=\newl, height=\newh]{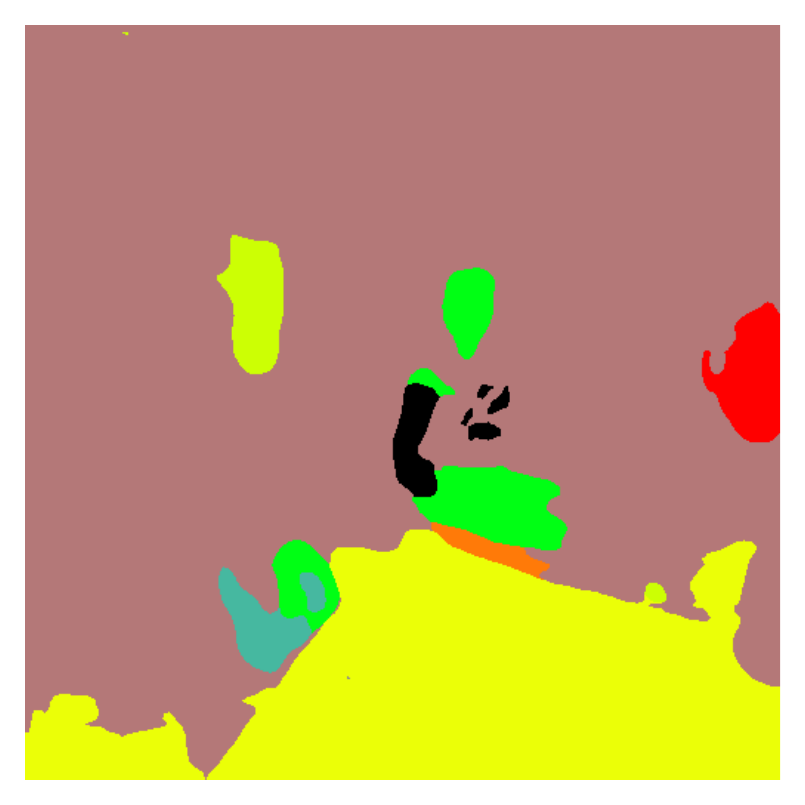}
\end{tabular}
\caption{
Same setting as in Fig.~\ref{fig:examples_ade_clean} for the 5 step \RobAT model. 
}
\label{fig:examples_ade_robust}

\end{figure}

\fi

\end{document}

%% file: commands.tex
\newcommand{\inner}[1]{\left\langle#1\right\rangle}
\def\R{\mathbb{R}}

\newcommand{\norm}[1]{\left\|#1\right\|}

\def\L{\mathcal{L}}

\def\argmax{\mathop{\rm arg\,max}\limits}

\def\maxop{\mathop{\rm max}\limits}

\newcolumntype{C}[1]{>{\centering\arraybackslash}p{#1}}
\newcolumntype{L}[1]{>{\raggedright\arraybackslash}p{#1}}
\newcolumntype{R}[1]{>{\raggedleft\arraybackslash}p{#1}}

\def\SP#1{\textsuperscript{\textcolor{black}{#1}}}

\definecolor{Gray}{gray}{0.85}
\definecolor{NewGray}{gray}{.45} 
\definecolor{BlueGray}{rgb}{0.92, 0.92, 1}
\definecolor{LightCyan}{rgb}{0.88,1,1}
\definecolor{LightGreen}{rgb}{0.8, 0.99, 0.95}
\definecolor{DarkGreen}{rgb}{.1, .75, .1}
\definecolor{DarkRed}{rgb}{.95, .0, .1}
\definecolor{GrayGreen}{rgb}{0.6,0.85,0.6}
\definecolor{GrayRed}{rgb}{0.85,0.6,0.6}
\definecolor{bluey}{rgb}{0.0,0.6,0.6}

\newcommand{\hlrev}[1]{\textcolor{black}{#1}}

\newcommand{\hlc}[2][yellow]{{%
    \colorlet{foo}{#1}%
    \sethlcolor{foo}\hl{#2}}%
}

\newcommand{\hide}[1]{}

\newlength\newl
\newlength\newh
\newlength\colwidth

\newcommand{\imagenet}{\textsc{ImageNet}\xspace}

\newcommand{\voc}{\textsc{Pascal-Voc}\xspace}
\newcommand{\ade}{\textsc{Ade20K}\xspace}

\newcommand{\acc}{\textsc{Acc}\xspace}
\newcommand{\miou}{m\textsc{IoU}\xspace}
\newcommand{\upernet}{{UPerNet}\xspace}
\newcommand{\convnext}{{ConvNeXt}\xspace}

\newcommand{\RobAT}{{PIR-AT}\xspace}

\newcommand{\multiloss}{
SEA\xspace}

\makeatletter
\def\blfootnote{\xdef\@thefnmark{}\@footnotetext}
\makeatother

%% file: math.tex
\usepackage{amsmath,amsfonts,bm}

\def\1{\bm{1}}

\def\eps{{\epsilon}}

\def\vdelta{{\bm{\delta}}}

\def\ve{{\bm{e}}}

\def\vm{{\bm{m}}}

\def\vp{{\bm{p}}}
\def\vq{{\bm{q}}}

\def\vu{{\bm{u}}}

\def\vx{{\bm{x}}}
\def\vy{{\bm{y}}}

\DeclareMathAlphabet{\mathsfit}{\encodingdefault}{\sfdefault}{m}{sl}
\SetMathAlphabet{\mathsfit}{bold}{\encodingdefault}{\sfdefault}{bx}{n}



%% file: tableLosses.tex
\begin{table}[ht!] \centering
\caption{\textbf{Comparison of attacks.} We 
compare the performance of all attacks for a budget of 300 iterations for clean and robust models trained on \voc and \ade. We report average pixel accuracy and \hlc[BlueGray]{\miou} (clean performance is next to the model name). Our JS and MCE attacks outperform SegPGD in almost all cases, whereas MCE-Bal targeted on minimizing mIoU achieves in most cases the best mIoU. Our ensemble attack \multiloss outperforms both CosPGD \cite{agnihotri2023cospgd} and SegPGD \cite{gu2022segpgd} by large margin. For each metric: best attack in \textbf{bold}, second best \underline{underlined}.
\label{tab:comp-losses}} 
\tabcolsep=0.5pt

\extrarowheight=1.0pt
\newl=7.95mm
\begin{tabular}{L{15mm}| *{2}{|C{\newl}%
>{\columncolor{BlueGray}}C{\newl}}| %
*{3}{|C{\newl}%
>{\columncolor{BlueGray}}C{\newl}} %
|C{\newl}
>{\columncolor{BlueGray}}C{\newl}
}
\multirow{2}{*}{$\epsilon_\infty$} & \multicolumn{4}{c||}{previous SOTA attacks %
} & \multicolumn{8}{c}{Our attacks}\\
& \multicolumn{2}{c}{CosPGD } 
& \multicolumn{2}{c||}{SegPGD } 
&\multicolumn{2}{c}{$\L_\textrm{JS}$} 
& \multicolumn{2}{c}{$\L_\textrm{MCE}$} 
& \multicolumn{2}{c|}{$\L_\textrm{MCE-Bal}$}  
& \multicolumn{2}{c}{SEA} \\
\toprule
\addlinespace[1mm]
\multicolumn{13}{l}{\textbf{\voc:} UPerNet+ConvNeXt-T \textbf{(clean training)} \quad (93.4\; \textcolor{NewGray}{\hlc[BlueGray]{77.2}})} \\ 
\midrule
0.25/255   & 76.6 & \textcolor{NewGray}{48.0} 
           & 74.0 & \textcolor{NewGray}{43.7} 
           & 72.4 & \textcolor{NewGray}{41.9} 
           & \underline{71.2} & \textcolor{NewGray}{\underline{39.6}} 
           & 76.6 & \textcolor{NewGray}{40.3}  
           & \textbf{70.0} & \textcolor{NewGray}{\textbf{36.9}} \\
0.5/255    & 46.9 & \textcolor{NewGray}{24.0}  
           & 43.0 & \textcolor{NewGray}{18.6} 
          & 36.7 & \textcolor{NewGray}{16.3}  
          & \underline{33.2} & \textcolor{NewGray}{13.0} 
          & 37.9 & \underline{\textcolor{NewGray}{10.1}} 
          & \textbf{31.1} & \textcolor{NewGray}{\textbf{8.6}} \\
1/255  & 17.2 & \textcolor{NewGray}{8.1} 
       & 12.9 & \textcolor{NewGray}{4.2} 
       & 8.2 &    \textcolor{NewGray}{3.4} 
       & \underline{5.9} & \textcolor{NewGray}{1.6} 
       & 6.9  &  \textcolor{NewGray}{\underline{0.8}}   
      & \textbf{4.9} & \textcolor{NewGray}{\textbf{0.6}}\\
\midrule
\multicolumn{13}{l}{\textbf{\voc:} PSPNet-50 \textbf{(DDC-AT~\cite{xu2021dynamic})}\quad (92.8\; \textcolor{NewGray}{\hlc[BlueGray]{76.0}})} \\ 
\midrule
2/255   & 7.1 & \textcolor{NewGray}{3.9} 
        & 3.9 & \textcolor{NewGray}{2.2} 
        & 1.2 & \textcolor{NewGray}{0.8} 
        & \underline{0.5} & \textcolor{NewGray}{\underline{0.3}}
        & 2.1 & \textcolor{NewGray}{1.0}
        & \textbf{0.2} & \textcolor{NewGray}{\textbf{0.1}} \\

4/255   & 6.5 &\textcolor{NewGray}{3.5} 
        & 2.6 &\textcolor{NewGray}{1.6} 
        & 0.3 & \textcolor{NewGray}{0.1} 
        & \underline{0.2} & \textcolor{NewGray}{0.1}
        & 0.5 & \textcolor{NewGray}{\underline{0.0}} 
        & \textbf{0.1} & \textcolor{NewGray}{\textbf{0.0}} \\
 
8/255   & 4.5 &\textcolor{NewGray}{3.2} 
        & 2.1 &\textcolor{NewGray}{1.3} 
        & \textbf{0.0} & \textcolor{NewGray}{\textbf{0.0}} 
        & \textbf{0.0} & \textcolor{NewGray}{\textbf{0.0}}
        & \textbf{0.0} & \textcolor{NewGray}{\textbf{0.0}}
        & \textbf{0.0} & \textcolor{NewGray}{\textbf{0.0}} \\

\midrule
\multicolumn{13}{l}{\textbf{\voc:} UPerNet+ConvNeXt-T \textbf{(\RobAT ours, \cref{tab:comp_robust_models_new})} \quad (92.7\; \textcolor{NewGray}{\hlc[BlueGray]{75.9}})}\\ 
\toprule
4/255   & 89.0 &\textcolor{NewGray}{65.6}    
        & \underline{88.7} &\textcolor{NewGray}{\textbf{64.8}} 
        &  \underline{88.7} &\textcolor{NewGray}{\underline{64.9}} 
         & 89.2 &\textcolor{NewGray}{65.9}  
        &   90.4 &\textcolor{NewGray}{67.4}  
        & \textbf{88.6} & \textcolor{NewGray}{\underline{64.9}}\\
8/255 & 77.8 & \textcolor{NewGray}{47.3} 
      & 74.2 & \textcolor{NewGray}{41.4} 
      & \underline{73.9} & \textcolor{NewGray}{41.3}
      &  74.0 &\textcolor{NewGray}{40.6}
      &  77.4 &\textcolor{NewGray}{\underline{38.4}} 
      &  \textbf{71.7} & \textcolor{NewGray}{\textbf{34.6}}\\ 
12/255  & 56.6 &\textcolor{NewGray}{26.4} 
        & 45.3 &\textcolor{NewGray}{15.8} 
        & 38.6 &\textcolor{NewGray}{15.1}  
         & \underline{31.5} &\textcolor{NewGray}{10.3} 
        & 36.9 &\textcolor{NewGray}{\underline{6.7}} 
        & \textbf{28.1} & \textcolor{NewGray}{\textbf{5.5}}\\
\midrule
\midrule
\addlinespace[1mm]
\multicolumn{13}{l}{\textbf{\ade:} UPerNet + \convnext-T \textbf{(AT 2 steps, \cref{tab:ablation})} \quad (73.4\; \textcolor{NewGray}{\hlc[BlueGray]{36.4}})}\\ 
\toprule
0.25/255 & 61.3 &\textcolor{NewGray}{24.6} 
      & 60.9 &\textcolor{NewGray}{24.4} 
      & \underline{58.8}&\textcolor{NewGray}{23.1} 
      & 59.4 &\textcolor{NewGray}{23.8}
      & 61.5 &\textcolor{NewGray}{\underline{21.6}}
      &\textbf{58.5} &\textcolor{NewGray}{\textbf{20.9}} \\ 
0.5/255 & 46.5 &\textcolor{NewGray}{14.6} 
      & 41.1 &\textcolor{NewGray}{12.2} 
      & 29.8&\textcolor{NewGray}{7.1} 
      & \underline{28.5}&\textcolor{NewGray}{6.3} 
      & 33.1 &\textcolor{NewGray}{\underline{5.9}}
      &  \textbf{27.5}&\textcolor{NewGray}{\textbf{5.1}} \\
1/255 & 18.3 &\textcolor{NewGray}{4.4} 
      & 9.9 &\textcolor{NewGray}{2.2} 
      & 1.8&\textcolor{NewGray}{0.3} 
      & \underline{1.1} &\textcolor{NewGray}{0.4}
      & 1.6 &\textcolor{NewGray}{\underline{0.1}}
      &\textbf{0.8} &\textcolor{NewGray}{\textbf{0.0}} \\ 
\midrule
\multicolumn{13}{l}{\textbf{\ade:} \upernet + \convnext -T \textbf{(\RobAT ours, \cref{tab:comp_robust_models_new})} \quad (70.5\; \textcolor{NewGray}{\hlc[BlueGray]{31.7}})} \\ 
\toprule
4/255   & 57.5 &\textcolor{NewGray}{19.9} 
        & \underline{55.9} &\textcolor{NewGray}{19.0} 
        & \underline{55.9} & \textcolor{NewGray}{18.9} 
        & 56.8 &\textcolor{NewGray}{20.0} 
        & 58.2 &\textcolor{NewGray}{\underline{17.9}} 
        & \textbf{55.5} & \textcolor{NewGray}{\textbf{17.2}} \\
8/255   & 37.6 &\textcolor{NewGray}{9.9} 
        & \underline{28.5} &\textcolor{NewGray}{7.4} 
        & \underline{28.5} & \textcolor{NewGray}{7.2}
        & \underline{28.5} &\textcolor{NewGray}{6.6} 
        & 31.1 &\textcolor{NewGray}{\underline{5.3}} 
        & \textbf{26.4} & \textcolor{NewGray}{\textbf{4.9}} \\
12/255  &  19.5 &\textcolor{NewGray}{4.1} 
        & 5.5 &\textcolor{NewGray}{1.2} 
        & 5.2 & \textcolor{NewGray}{1.1} 
        & \underline{3.7} &\textcolor{NewGray}{\underline{0.9}} 
        & 5.2 &\textcolor{NewGray}{\underline{0.9}} 
        & \textbf{3.1} & \textcolor{NewGray}{\textbf{0.4}} \\
\midrule

\multicolumn{13}{l}{\textbf{\ade:} Segmenter + ViT-S \textbf{(\RobAT ours, \cref{tab:comp_robust_models_new})} \quad (69.1\; \textcolor{NewGray}{\hlc[BlueGray]{28.7}})}\\ 
\toprule
4/255 & 57.4 &\textcolor{NewGray}{17.5} 
      & 57.3 &\textcolor{NewGray}{17.5} 
      & \underline{55.6}&\textcolor{NewGray}{16.6} 
      & 56.9 &\textcolor{NewGray}{17.8}
      & 57.6 &\textcolor{NewGray}{\underline{15.6}}
      &\textbf{55.3} &\textcolor{NewGray}{\textbf{14.9}} \\ 
8/255 & 41.7 &\textcolor{NewGray}{9.7} 
      & 38.5 &\textcolor{NewGray}{8.6} 
      & \underline{34.2}&\textcolor{NewGray}{7.7} 
      & 36.2&\textcolor{NewGray}{8.5} 
      & 37.8 &\textcolor{NewGray}{\underline{5.6}}
      &  \textbf{33.3}&\textcolor{NewGray}{\textbf{5.4}} \\
12/255 & 25.6 &\textcolor{NewGray}{4.9} 
       & 17.4 &\textcolor{NewGray}{2.9} 
       & 11.2&\textcolor{NewGray}{2.2}  
       & \underline{10.5}&\textcolor{NewGray}{2.2} 
        & 11.7&\textcolor{NewGray}{\underline{1.3}} 
       & \textbf{8.9}&\textcolor{NewGray}{\textbf{1.1}} \\ 

\bottomrule
\end{tabular} \end{table}

%% file: main_figure.tex
\begin{figure}[!t] \centering
\footnotesize
\tabcolsep=1.1pt
\newl=.16\columnwidth
\begin{tabular}{c | c c c c c} 
\toprule
\addlinespace[1mm]
\multicolumn{6}{l}{\textbf{\upernet with \convnext-T backbone - clean training}} \\
\addlinespace[1mm]

original & 0 & 0.25/255 & 0.5/255 & 1/255 & 2/255\\
\toprule 
& \acc: 95.9\%& \acc: 94.8\%& \acc: 75.9\%& \acc: 48.3\%& \acc: 0.0\%\\
\includegraphics[width=\newl]{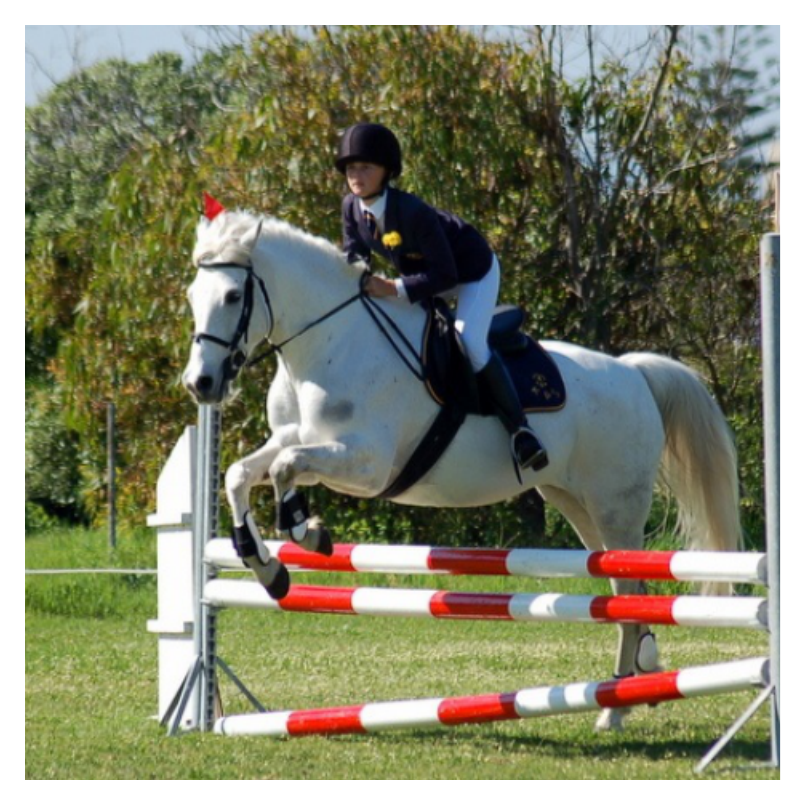} & \includegraphics[width=\newl]{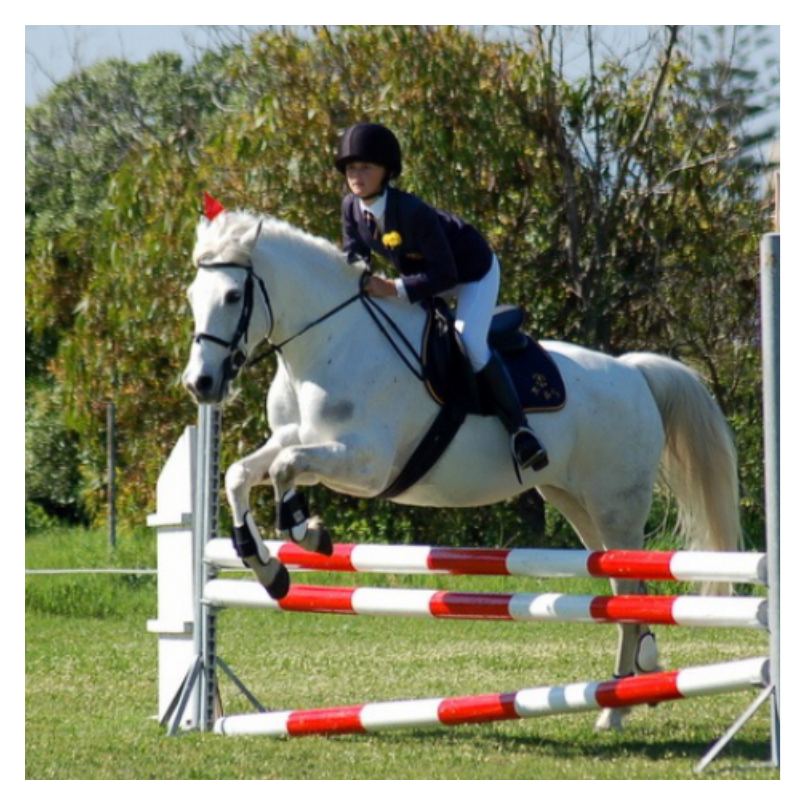} & \includegraphics[width=\newl]{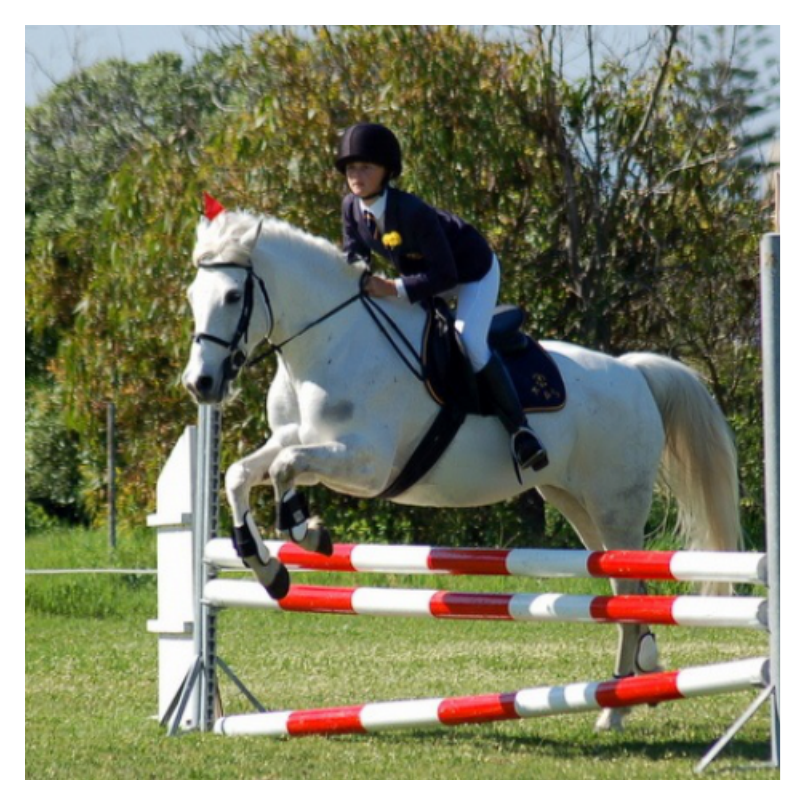} & \includegraphics[width=\newl]{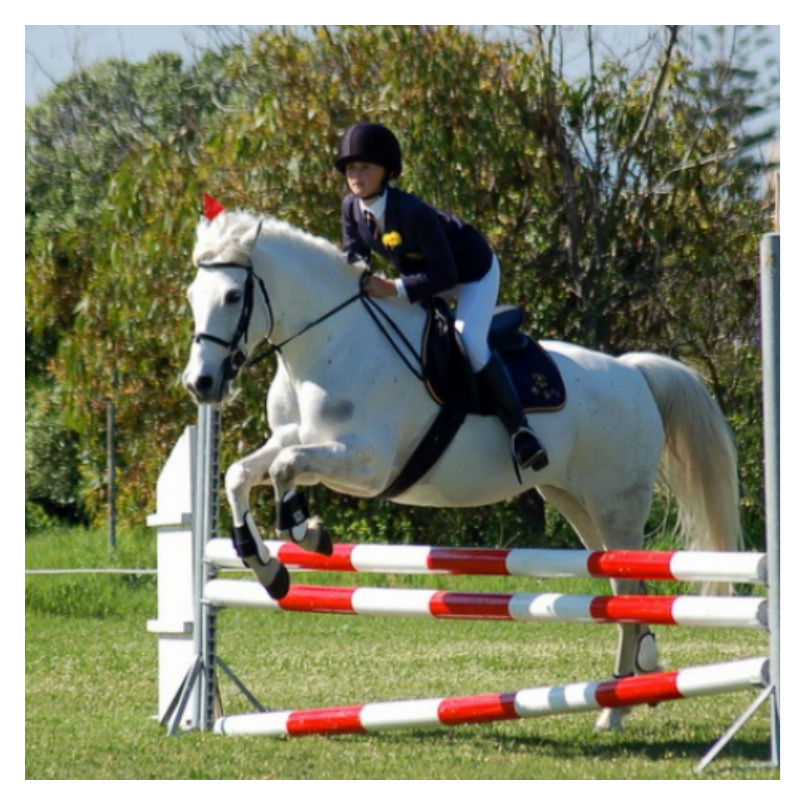} & \includegraphics[width=\newl]{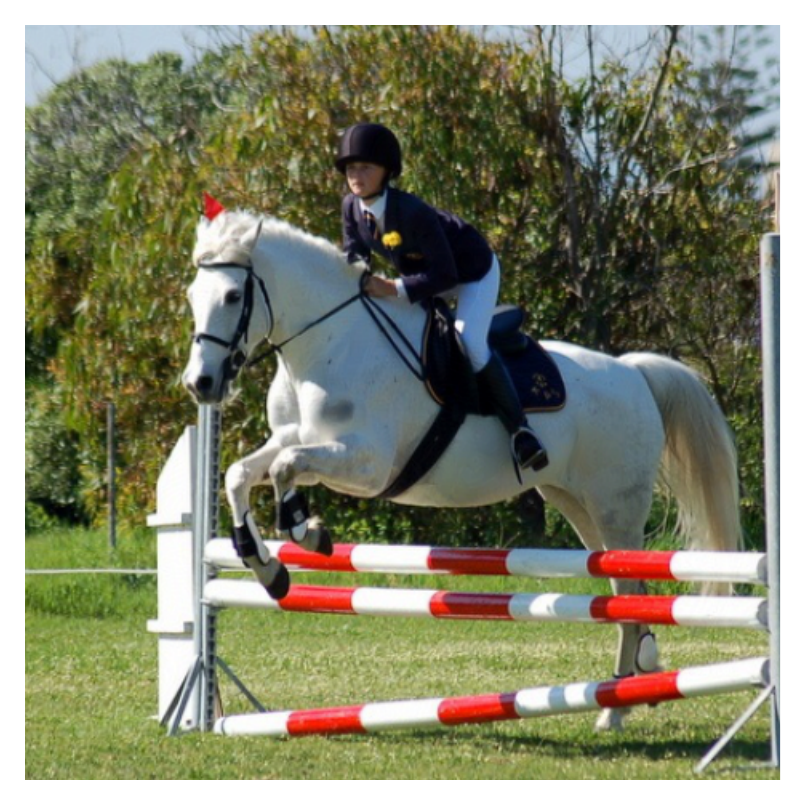} & \includegraphics[width=\newl]{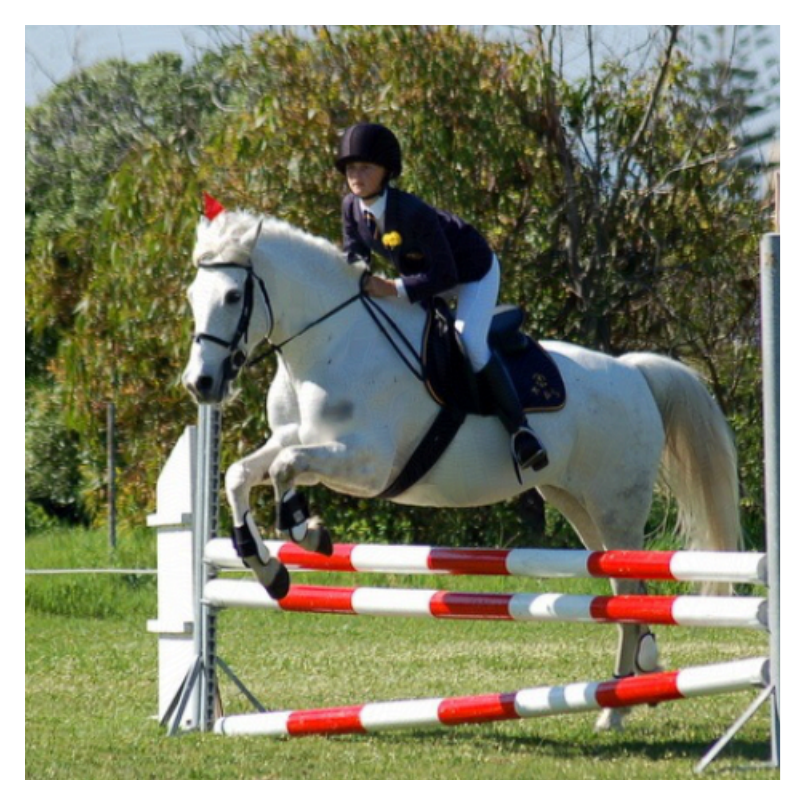} \\ \includegraphics[width=\newl]{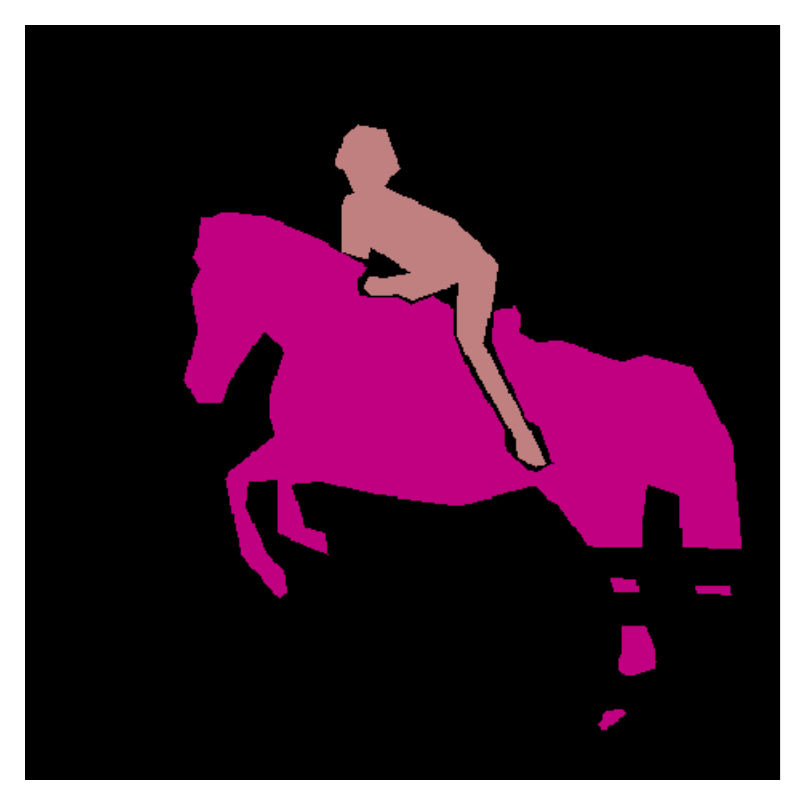} & \includegraphics[width=\newl]{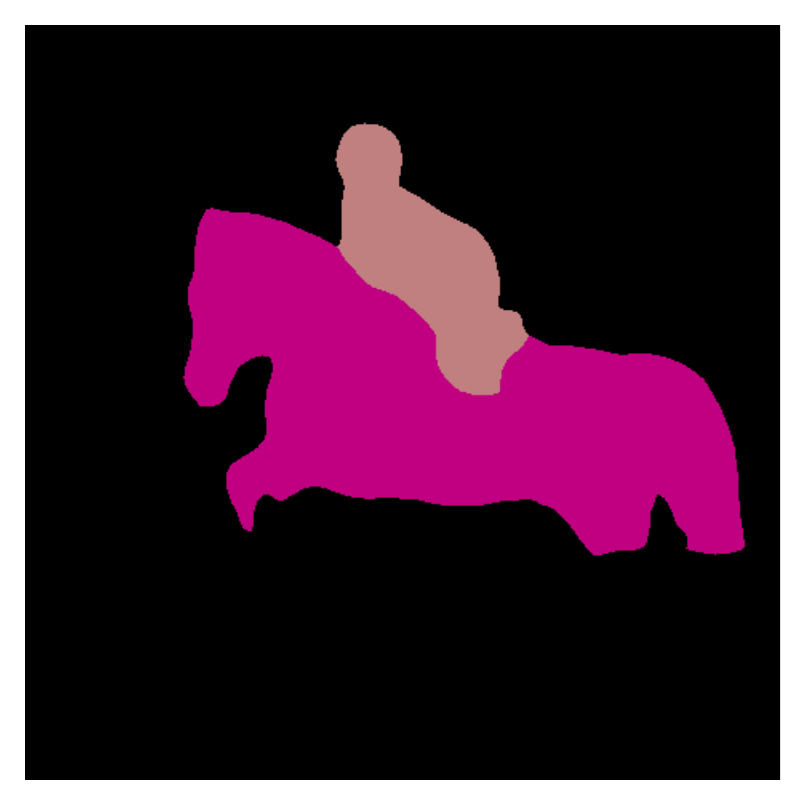} & \includegraphics[width=\newl]{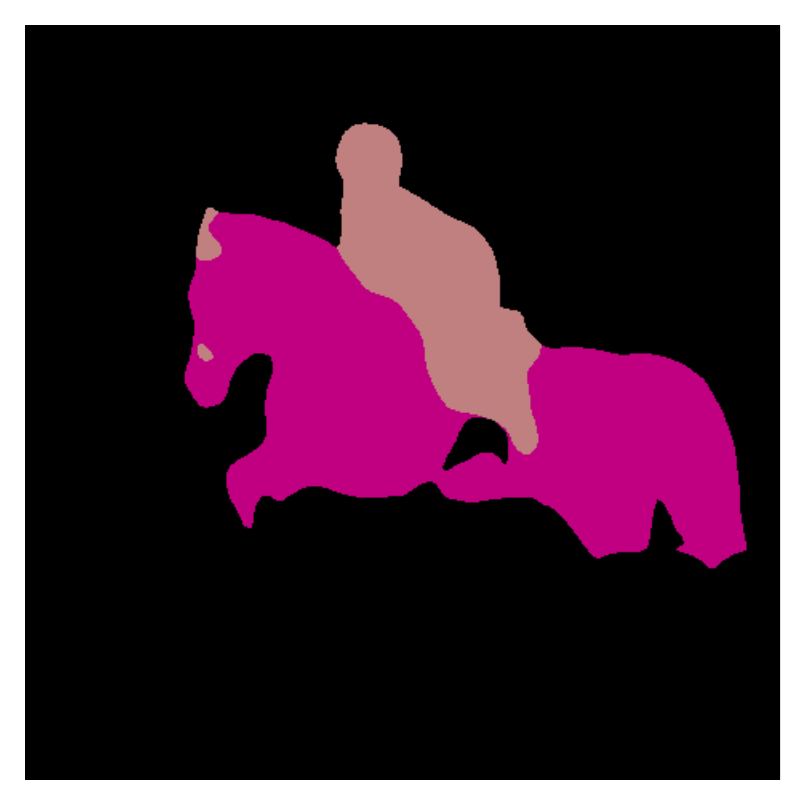} & \includegraphics[width=\newl]{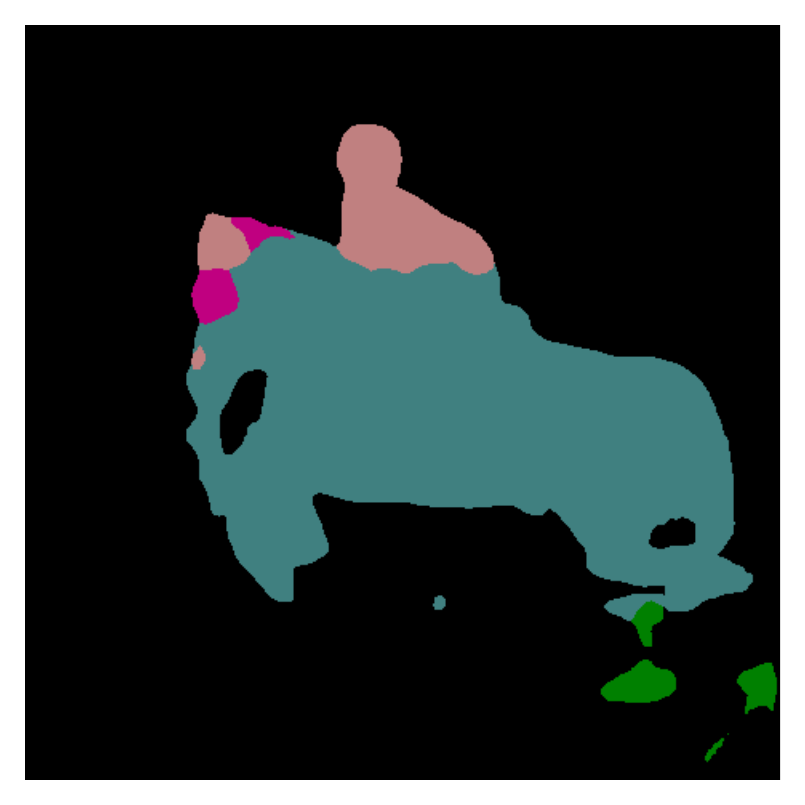} & \includegraphics[width=\newl]{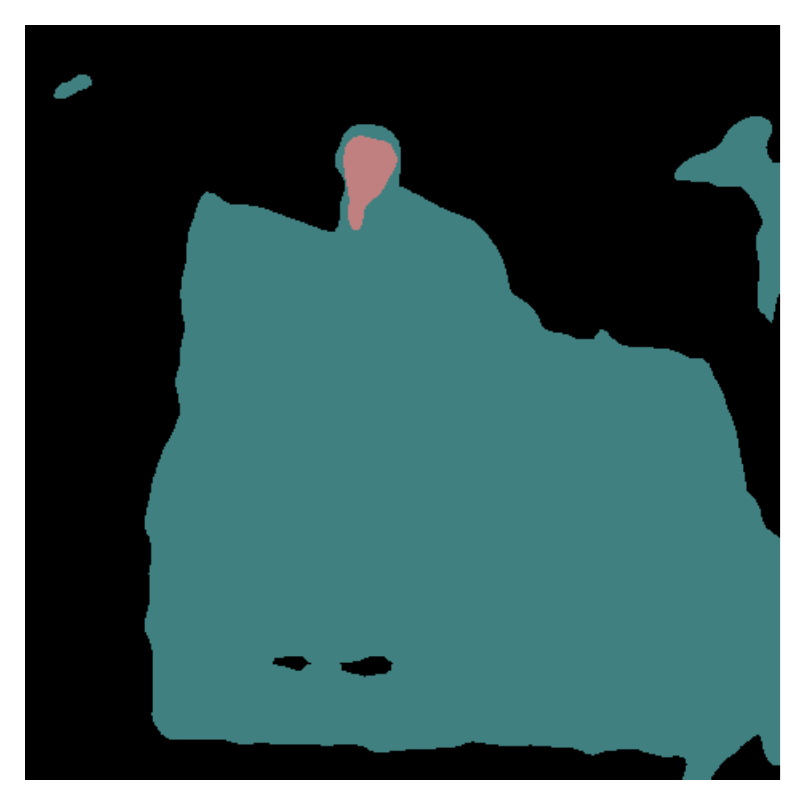} & \includegraphics[width=\newl]{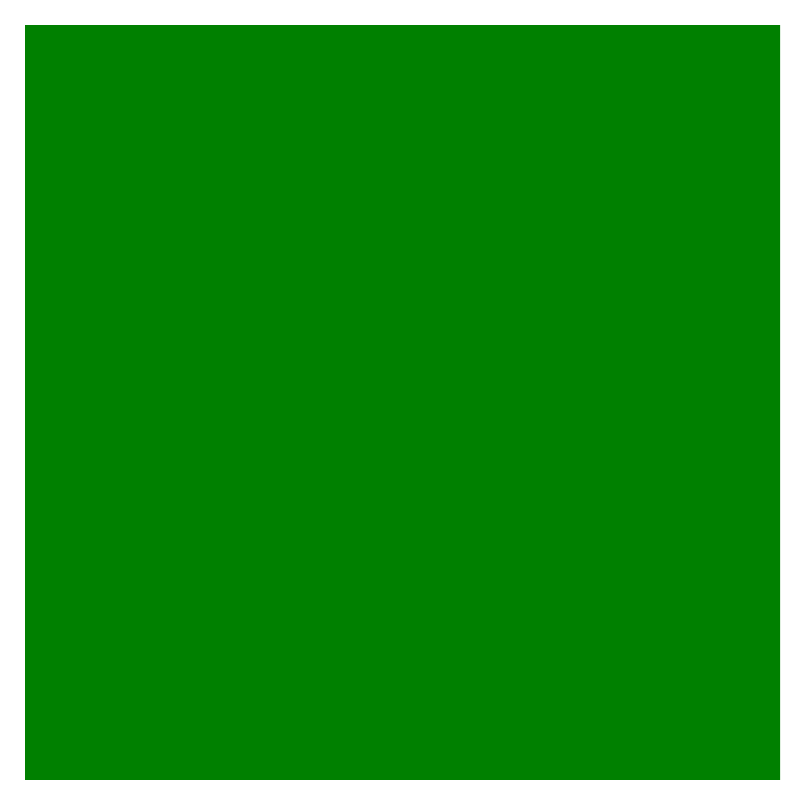}\\
\end{tabular}
\newl=.16\columnwidth
\begin{tabular}{c | c c c c c} 
\toprule
\addlinespace[1mm]
\multicolumn{6}{l}{\textbf{UperNet with \convnext-T backbone - our \RobAT}} \\
\addlinespace[1mm]

original & 0 & 4/255 & 8/255 & 12/255 & 16/255\\
\toprule
& \acc: 95.5\%& \acc: 94.6\%& \acc: 90.8\%& \acc: 49.2\%& \acc: 0.0\%\\
\includegraphics[width=\newl]{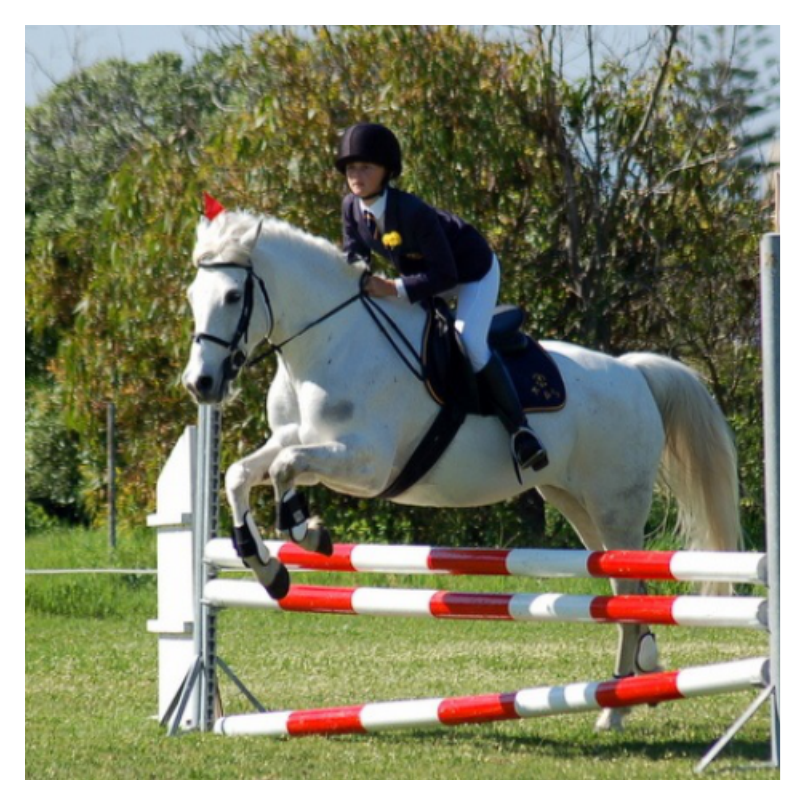} & \includegraphics[width=\newl]{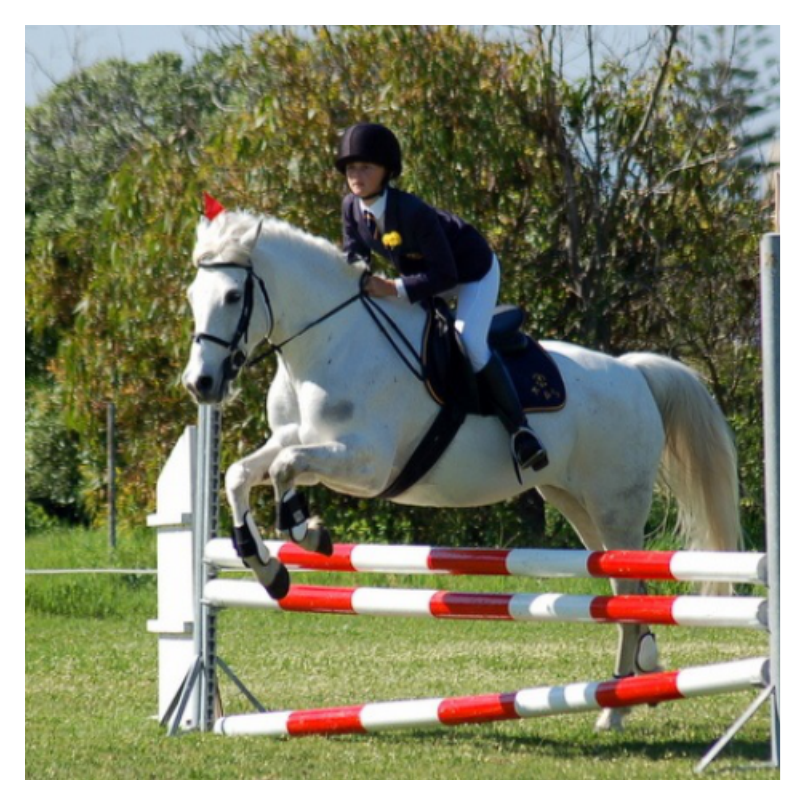} & \includegraphics[width=\newl]{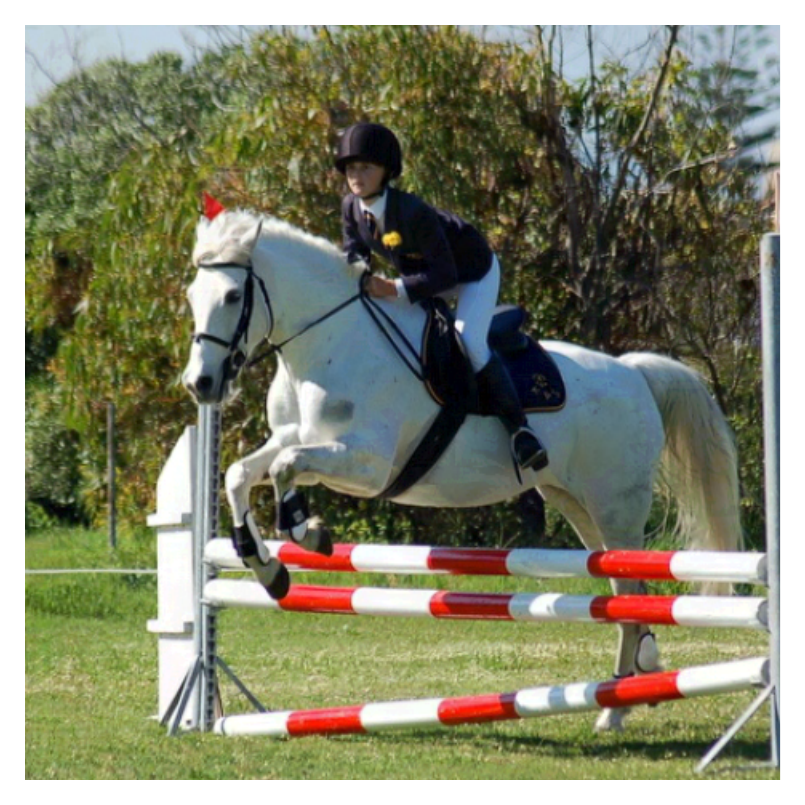} & \includegraphics[width=\newl]{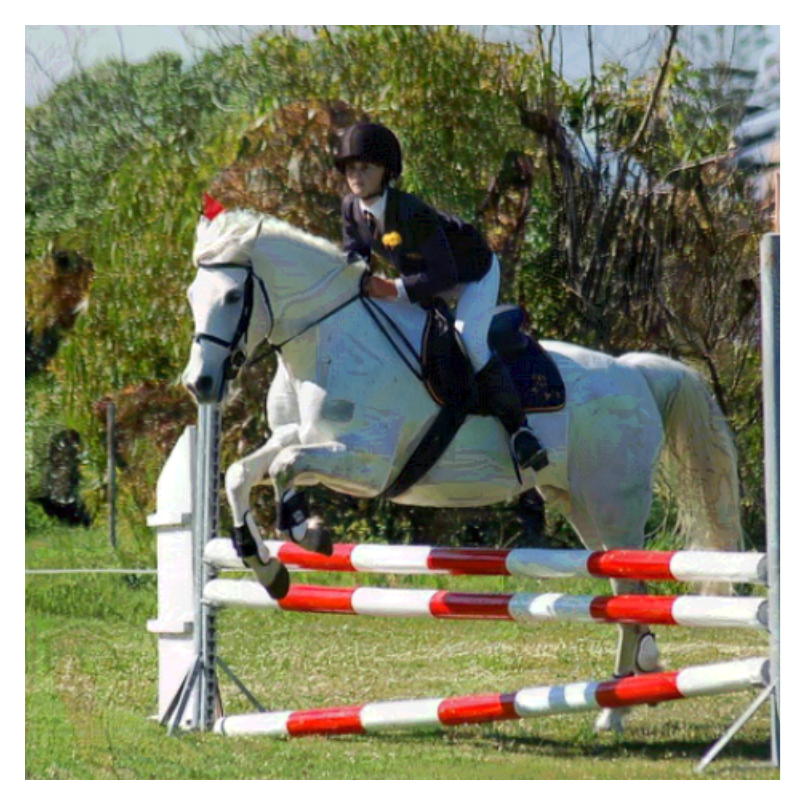} & \includegraphics[width=\newl]{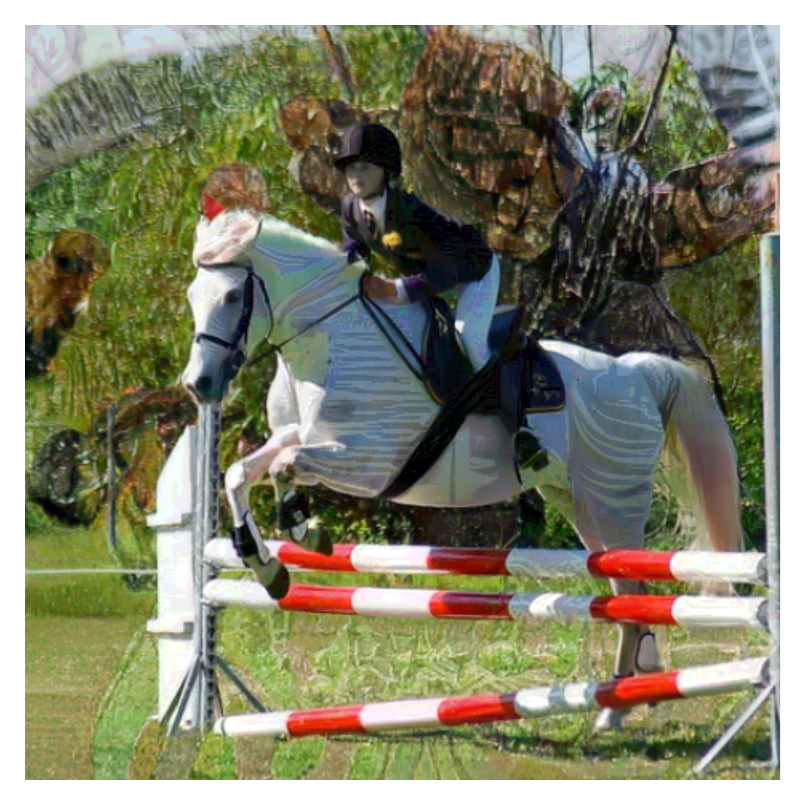} & \includegraphics[width=\newl]{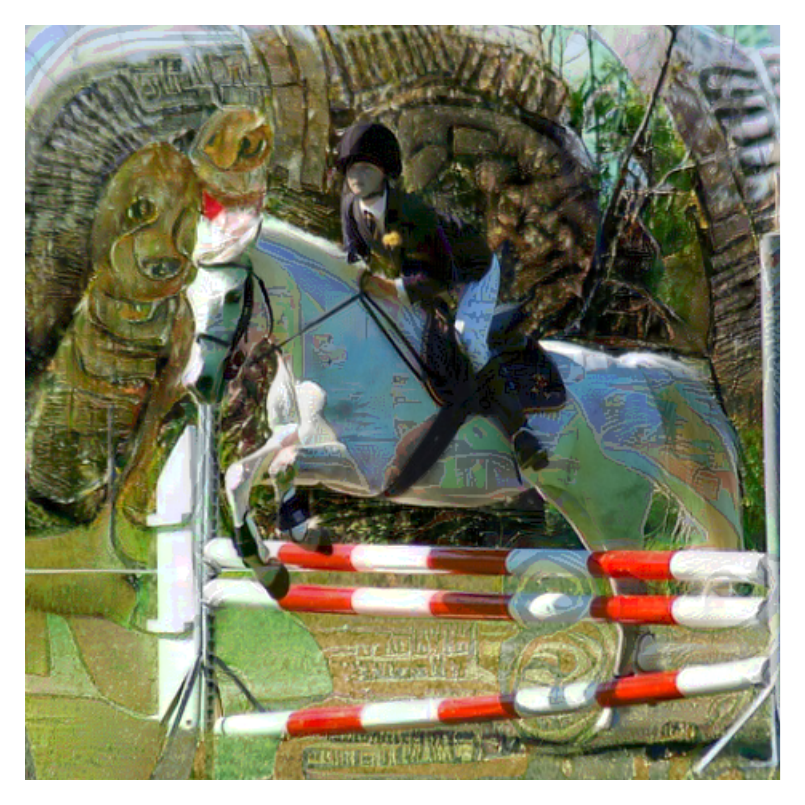} \\ \includegraphics[width=\newl]{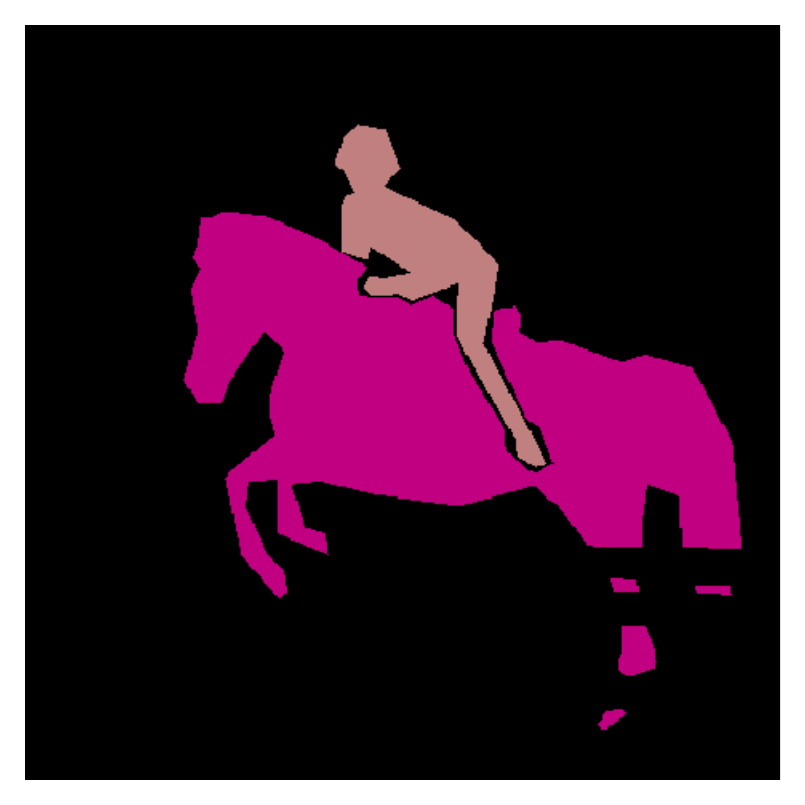} & \includegraphics[width=\newl]{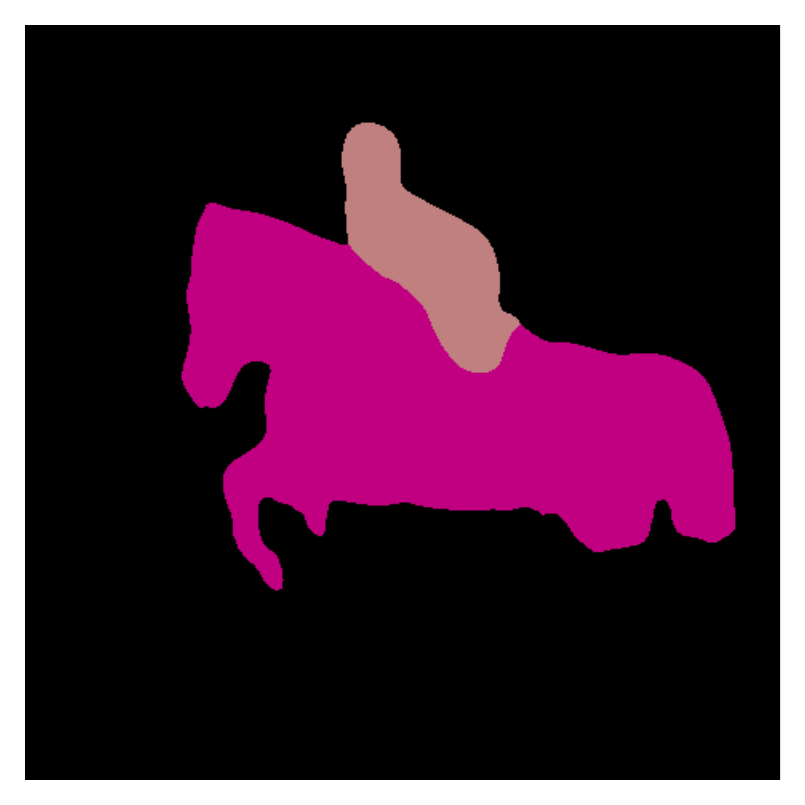} & \includegraphics[width=\newl]{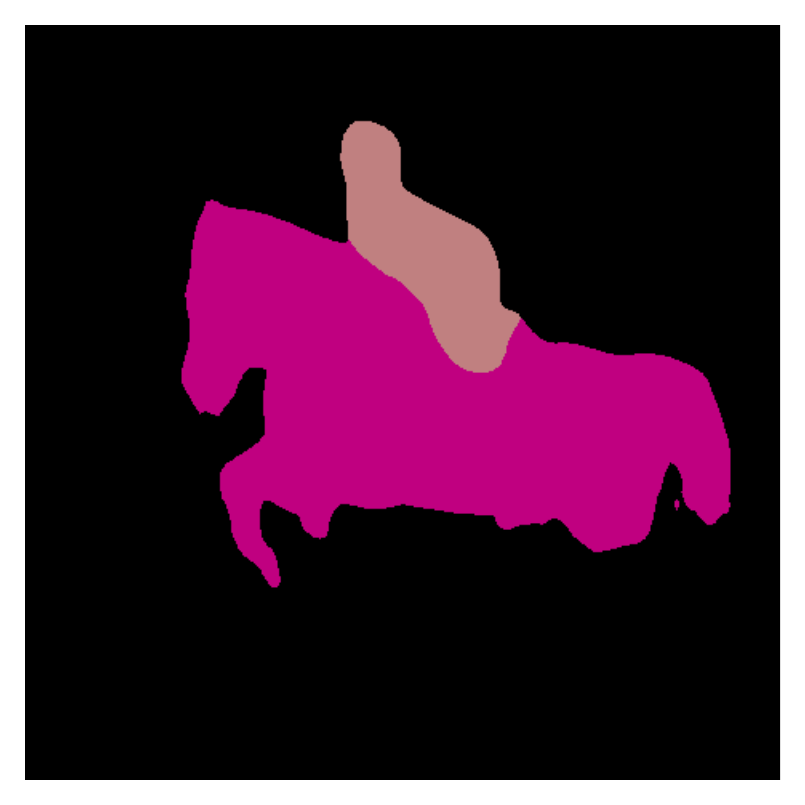} & \includegraphics[width=\newl]{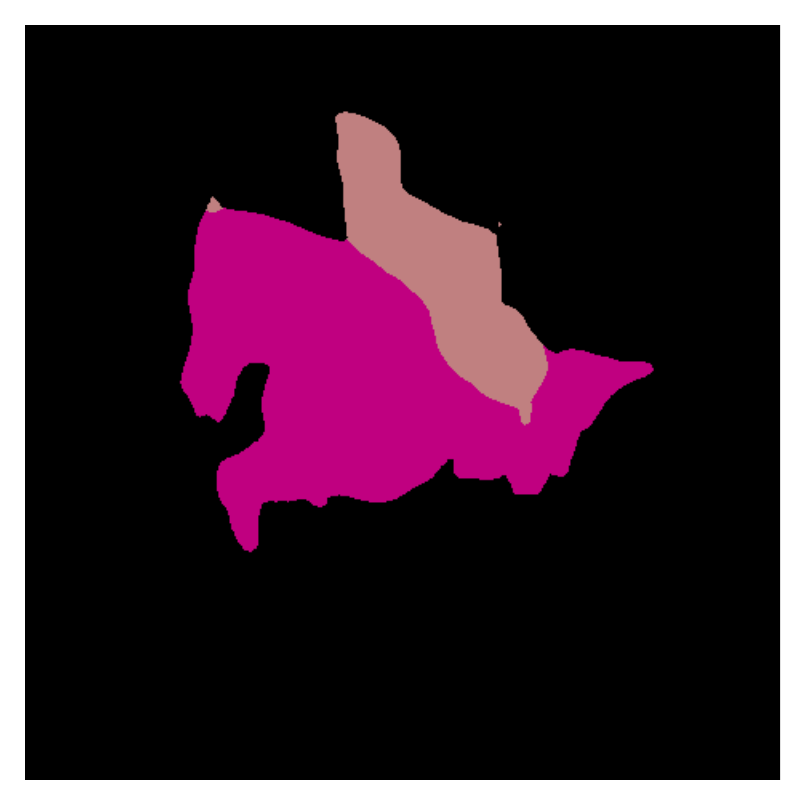} & \includegraphics[width=\newl]{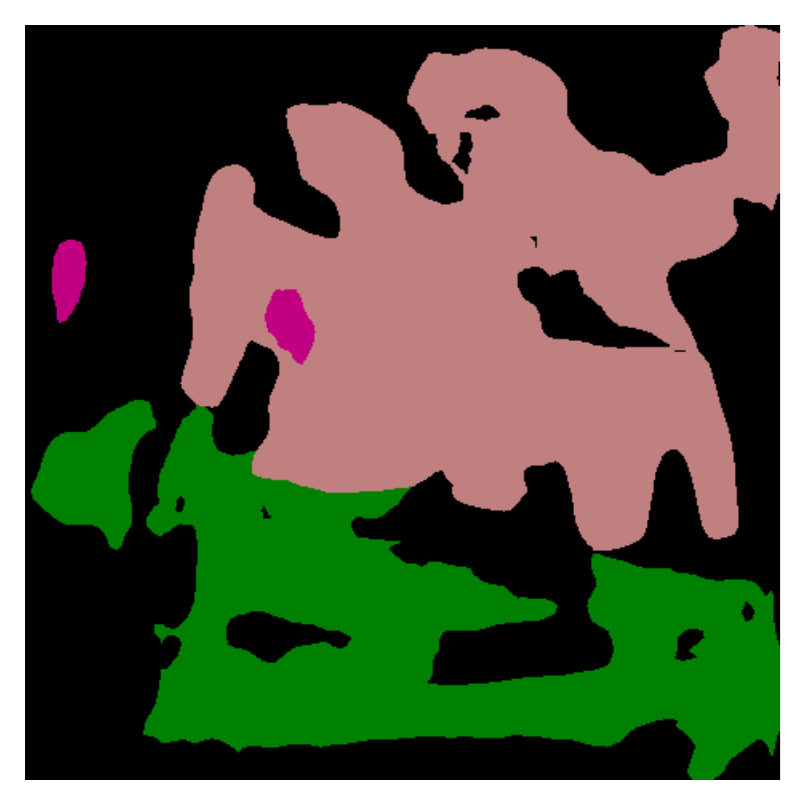} & \includegraphics[width=\newl]{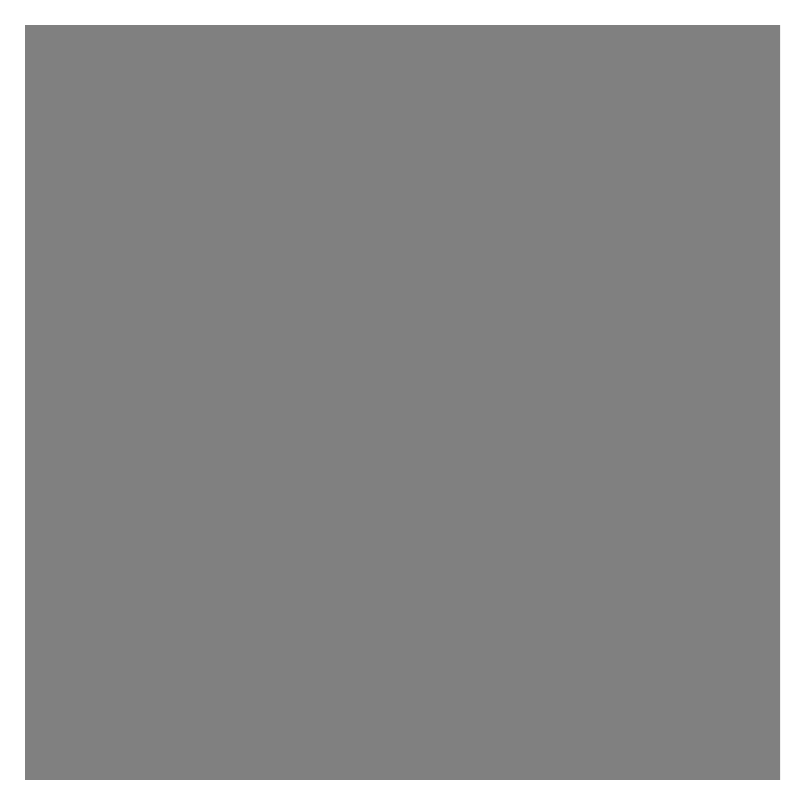}\\
\end{tabular}
\caption{%
\textbf{Visualizing adversarial images and their segmentation outputs.} We show the perturbed images, corresponding predicted segmentation masks and average accuracy for increasing radii for both the \textbf{clean} and our \textbf{\RobAT} models. The adversarial images are generated on \voc with APGD on $\L_\textrm{Mask-CE}$. For the clean model even at a smaller radii of \nicefrac{0.5}{255}, the predicted mask deviates from the ground truth significantly. Whereas for the \RobAT model the predicted mask is similar to the ground truth even at a high perturbation strength of \nicefrac{8}{255}. More visualizations can be found in \cref{app:additional-fig}.}
\label{fig:examples_voc}
\end{figure}